\documentclass{article}
\pdfoutput=1

\usepackage[preprint]{neurips_2021}
\usepackage[medium,compact]{titlesec}

\usepackage{setspace}
\usepackage{hyperref}
\usepackage{diagbox}
\usepackage[table,xcdraw]{xcolor}
\usepackage[utf8]{inputenc} % allow utf-8 input
\usepackage[T1]{fontenc}    % use 8-bit T1 fonts
\usepackage{hyperref}       % hyperlinks
\usepackage{url}            % simple URL typesetting
\usepackage{booktabs}       % professional-quality tables
\usepackage{amsfonts}       % blackboard math symbols
\usepackage{nicefrac}       % compact symbols for 1/2, etc.
\usepackage{microtype}      % microtypographyhttps://www.overleaf.com/project/62064cfa2a4921563b535a86
\usepackage{xcolor}         % colors
\usepackage{times}
\usepackage{epsfig}
\usepackage{graphicx}
\usepackage{amsmath}
\usepackage{amssymb}

\usepackage{amsmath}
\usepackage{amsfonts}
\usepackage{dsfont}
\usepackage{multirow}
\usepackage{adjustbox}
\usepackage{wrapfig}
\usepackage{threeparttable}    %这行要添加
\usepackage{appendix}
\usepackage[ruled,linesnumbered]{algorithm2e}
\usepackage{subfigure}
\makeatletter

\newcommand{\Rmnum}[1]{\expandafter\@slowromancap\romannumeral #1@}

\newcommand{\etal}{\emph{et~al.}\xspace}
\newcommand{\eg}{\emph{e.g.},\xspace}

\newcommand{\etc}{\emph{etc.}\xspace}

\DeclareMathAlphabet\mathbfcal{OMS}{cmsy}{b}{n}

\newcommand{\eat}[1]{}
\ifodd 1
\newcommand{\TODO}[1]{{\color{red}TODO:{#1}}}

\else

\newcommand{\TODO}[1]{}

\fi

\title{Bkd-FedGNN: A Benchmark for Classification Backdoor Attacks on Federated Graph Neural Network}

\author{%
  Fan~LIU$^1$, Siqi~LAI$^1$, Yansong~NING$^1$,  
    Hao~LIU$^{*, 1, 2}$ \\
   AI Thrust, The Hong Kong University of Science and Technology (Guangzhou)$^1$; \\ 
  CSE, The Hong Kong University of Science and Technology$^2$ \\
  \texttt{fliu236@connect.hkust-gz.edu.cn; siqilai@hkust-gz.edu.cn;} \\
    \texttt{ yansongning@ust.hk; liuh@ust.hk}\\
}

\begin{document}

\maketitle

\begin{abstract}
Federated Graph Neural Network (FedGNN) has recently  emerged as a rapidly growing research topic, as it integrates the strengths of graph neural networks and federated learning to enable advanced machine learning applications without direct access to sensitive data.
Despite its advantages, the distributed nature of FedGNN introduces additional vulnerabilities, particularly backdoor attacks stemming from malicious participants. 
Although graph backdoor attacks have been explored, the compounded complexity introduced by the combination of GNNs and federated learning has hindered a comprehensive understanding of these attacks, as existing research lacks extensive benchmark coverage and in-depth analysis of critical factors.
To address these limitations, we propose Bkd-FedGNN, a benchmark for backdoor attacks on FedGNN. 
Specifically, Bkd-FedGNN decomposes the graph backdoor attack into trigger generation and injection steps, and extending the attack to   the node-level federated setting, resulting in a unified framework that covers both node-level and graph-level classification tasks.
Moreover, we thoroughly investigate the impact of multiple critical factors in backdoor attacks on FedGNN.
These factors are categorized into global-level and local-level factors, including data distribution, the number of malicious attackers, attack time, overlapping rate, trigger size, trigger type, trigger position, and poisoning rate.
Finally, we conduct comprehensive evaluations on 13 benchmark datasets and 13 critical factors, comprising 1,725 experimental configurations for node-level and graph-level tasks from six domains. These experiments encompass over 8,000 individual tests, allowing us to provide a thorough evaluation and insightful observations that advance our understanding of backdoor attacks on FedGNN. The Bkd-FedGNN benchmark is publicly available at \url{https://github.com/usail-hkust/BkdFedGCN}.
\end{abstract}

\section{Introduction}

The Federated Graph Neural Network (FedGNN) has emerged as a fast-evolving research area that combines the capabilities of graph neural networks and federated learning. Such integration allows for advanced machine learning applications without requiring direct access to sensitive data~\cite{MaekawaN, LiuDZDHZDCPSSLC22, zhang2022multi, ST22, GuiLWJ22, QinZWZZ22, DuanLWZZCHW22, Huang0W0ZXCV22, he2021spreadgnn}. However, despite its numerous advantages, the distributed nature of FedGNN introduces additional vulnerabilities, particularly related to backdoor attacks originating from malicious participants. In particular, these adversaries have the ability to inject graph backdoor triggers into their training data, thereby undermining the overall trustworthiness of the system~\cite{NEURIPS2020_b8ffa41d, bagdasaryan2020backdoor, Xie0CL21, li2023learning, OzdayiKG21}.

Although considerable research efforts have explored graph backdoor attacks on FedGNN~\cite{chen2022graph, xu2022more, abs-2207-05521, XuBZAKL22}, a comprehensive understanding of these attacks is hindered by the compounded complexity introduced by the combination of Graph Neural Networks (GNNs) and Federated Learning (FL). Existing studies suffer from a lack of extensive benchmark coverage and in-depth analysis of critical factors. \textbf{(1) Lack of Extensive Benchmark Coverage.} Specifically, the lack of extensive benchmark coverage poses challenges in fairly and comprehensively comparing graph backdoor attacks on FedGNN across different settings. These settings can be categorized into two levels: the graph backdoor attack level and the FedGNN task level. At the graph backdoor attack level, trigger generation and injection steps are involved. Additionally, the classification tasks in FedGNN encompass both node and graph classification tasks. However, there is still a dearth of comprehensive exploration of graph backdoor attacks on FedGNN under these various settings. \textbf{(2) Insufficient Exploration of Multiple Factors.} Furthermore, there has been the insufficient exploration of multiple factors that impact FedGNN.  The combination of GNN with FL introduces various factors that affect backdoor attacks, such as trigger type, trigger size, and data distribution. The insufficient exploration and analysis of these multiple factors make it difficult to understand the influence of key factors on the behavior of FedGNN.

To address these limitations, we propose a benchmark for  graph backdoor attacks on FedGNN, called Bkd-FedGNN.  As far as we are aware, our work is the first comprehensive investigation of graph backdoor attacks on FedGNN. Our contributions can be summarized as follows.
\begin{itemize}
    \item \textbf{Unified Framework}: We propose a unified framework for classification backdoor attacks on FedGNN. Bkd-FedGNN decomposes the graph backdoor attack into trigger generation and injection steps and extends the attack to the node-level federated setting, resulting in a unified framework that covers both node-level and graph-level classification tasks.
    \item \textbf{Exploration of Multiple Critical Factors}: We thoroughly investigate the impact of multiple critical factors on graph backdoor attacks in FedGNN. We systematically categorize these factors into two levels: global level and local level. At the global level, factors such as data distribution, the number of malicious attackers, the start time of backdoor attacks, and the overlapping rate play significant roles. In addition, the local level factors involve factors such as trigger size, trigger type, trigger position, and poisoning rate.
    \item  \textbf{Comprehensive Experiments and Analysis}: We conduct comprehensive experiments on both benchmark experiments and critical factor analysis. For the benchmark experiments, we consider combinations of trigger types, trigger positions, datasets, and models, resulting in 315 configurations for the node level and 270 configurations for the graph-level tasks. Regarding the critical factors, we consider combinations of factors, datasets, and models, resulting in 672 configurations for the node-level tasks and 468 configurations for the graph-level tasks. Each configuration is tested five times, resulting in approximately 8,000 individual experiments in total. Based on these experiments, we thoroughly evaluate the presented comprehensive analysis and provide insightful observations that advance the field.
\end{itemize}

\section{Federated Graph Neural Network}
In this section, we provide an introduction to the preliminary aspects of FedGNN. Currently, FedGNN primarily focuses on exploring common classification tasks, which involve both node-level and graph-level classification. The  FedGNN consists of two levels: client-level local training and server-level federated optimization. We will begin by providing an overview of the notations used, followed by a detailed explanation of the client-level local training, which encompasses message passing and readout techniques. Lastly, we will introduce server-level federated optimization.

\subsection{Notations}
Assume that there exist $K$ clients denoted as $\mathcal{C}=\{\mathit{c}_{k}\}_{k=1}^{K}$. Each client, $\mathit{c}_{i}$, possesses a private dataset denoted as $\mathcal{D}^{i}=\{ (\mathcal{G}_{j}^{i},\mathbf{Y}_{j}^{i})\}_{j=1}^{N_{i}}$, wherein  
$\mathcal{G}_{j}^{i}=(\mathcal{V}_{j}^{i},\mathcal{E}_{j}^{i})$ is the graph, where $  \mathcal{V}^{i}=\{\mathit{v}_{t}\}_{t=1}^{n_{i}}$ ($n_i$ denotes the number of nodes) is the set of nodes, and $\mathcal{E}^{i}=\{\mathit{e}_{tk}\}_{t,k}$ is the set of edges (for simplicity, we exclude the subscript $j$ that indicates the index of the $j$-th dataset in the dataset $\mathcal{D}^{i}$). $N_{i} = \left | \mathcal{D}^{i}\right |$ denotes the total number of data samples in the private dataset of client $\mathit{c}_{i}$.
 We employ the notation $\mathbf{A}^{i}_{j}$ to denote the adjacency matrix of graph $\mathcal{G}^{i}_{j}$ belonging to client $\mathit{c}_{i}$ within the set of clients $\mathcal{C}$. $\mathbf{X}^{i}_{j}$ represents the node feature set , and $\mathbf{Y}^{i}_{j}$ corresponds to the label  sets. 

\subsection{Client-level Local Training}

To ensure versatility and inclusiveness, we employ the message passing neural network (MPNN) framework~\cite{GilmerSRVD17, RongXHHC0WDWM20}, which encompasses a diverse range of spectral-based  GNNs, such as GCN~\cite{GCN}, as well as spatial-based GNNs including GAT~\cite{GAT} and GraphSage~\cite{GraphSage}, \etc  Each client possesses a GNN model that collaboratively trains a global model. The local graph learning process can be divided into two stages: message passing and readout. 

\textbf{Message Passing.} For each client $\mathit{c}_{i}$, the $l$-th layer in MPNN can be formulated as follows,
\begin{equation}
    \mathbf{h}_{j}^{l,i} = \sigma (\mathit{w}^{l,i}\cdot(\mathbf{h}_{j}^{l-1,i},\textit{Agg}(\{\mathbf{h}_{k}^{l-1,i} |\mathit{v}_{k} \in \mathcal{N}(\mathit{v}_{j})\}))),
\end{equation}
where $ \mathbf{h}_{j}^{l,i}$ ($l=0,\cdots,L-1$)  represents the hidden feature of node $\mathit{v}_{j}$ in client $\mathit{c}_{i} $  and $ \mathbf{h}_{j}^{0,i} = \mathbf{x}_{j}$ denotes the node $\mathit{v}_{j}$'s raw feature.   The $\sigma$  represents the activation function (e.g., ReLU, sigmoid). The parameter $\mathit{w}^{l,i}$ corresponds to the $l$-th learnable parameter. The aggregation operation $\textit{Agg}$ (e.g., mean pooling) combines the hidden features $\mathbf{h}_{k}^{l-1,i}$  of neighboring nodes $\mathit{v}_{k} \in \mathcal{N}(\mathit{v}_{j})$  for node $\mathit{v}_{j} $, where $\mathcal{N}(\mathit{v}_{j})$ represents the set of neighbors of node $\mathit{v}_{j}$. Assume that the $\mathbf{w}^{i}=\{\mathit{w}^{l,i}\}_{l=0}^{L-1}$ is the set of learn able parameters for client $\mathit{c}_{i}$.

\textbf{Readout.} Following the propagation of information through $L$ layers of MPNN, the final hidden feature is computed using a readout function for subsequent tasks. 
\begin{equation}
    \mathit{\hat{y} }_{I}^{i} = R_{\theta^{i}}(\{ \mathbf{h}_{j}^{L,i} | \mathit{v}_{j} \in \mathcal{V}_{I}^{i} \}),
\end{equation}
where $\mathit{\hat{y}}_{I}^{i}$ represents the prediction for a node or graph. Specifically, $I$ serves as an indicator, where $I=\mathit{v}_{j}$ denotes the prediction for node $\mathit{v}_{j}$, and $I = \mathcal{G}^{i}$ denotes the prediction for the graph $\mathcal{G}^{i}$. The readout function $ R_{\theta^{i}}(\cdot)$ encompasses methods such as mean pooling or sum pooling \etc, where $\theta^{i}$ is the parameter for readout function.

\subsection{Server-level  Federated  Optimization}
Let us consider that $\mathbf{w}^{i}=\{\mathit{w}^{l,i}\}_{l=0}^{L-1}$ represents the set of trainable parameters within the MPNN framework associated with client $\mathit{c}_{i}$. Consequently, we define the overall model parameters as $\mathbf{W}^{i}=\{\mathbf{w}^{i}, \theta^{i}\}$ for each client $\mathit{c}_{i} \in \mathcal{C}$. The GNNs, which constitute a part of this framework, can be represented as $f_{i}(\mathbf{X}_{j}^{i},\mathbf{A}_{j}^{i};\mathbf{W}^{i})$.
The objective of  FL is to optimize the global objective function while preserving the privacy of local data on each individual local model. The overall  objective function can be formulated as follows,

\begin{gather}
	    \min_{\{\mathbf{W}^{i}\}} \sum_{i \in \mathcal{C}}\frac{N_{i}}{N}F_{i}(\mathbf{W}^{i}), \quad F_{i}(\mathbf{W}^{i}) = \frac{1}{N_{i}}\sum_{j \in \mathcal{D}^{i}}\mathcal{L}((f_{i}(\mathbf{X}_{j}^{i},\mathbf{A}_{j}^{i};\mathbf{W}^{i}),\mathbf{Y}_{j}^{i}),
\end{gather}

where   $F_{i}(\cdot)$  denotes the local objective function, and  $\mathcal{L}(\cdot)$ denote the loss function (\eg cross-entropy \etc), and $N = \sum_{i=1}^{K}N_{i}$ represent the total number of data samples encompassing all clients. 

We illustrate the process of federated optimization, aimed at achieving a generalized model while ensuring privacy preservation, by utilizing a representative federated algorithm, FedAvg~\cite{McMahanMRHA}.  Specifically, in each round denoted by $t$, the central server transmits the global model parameter $\mathbf{W}_{t}$ to a subset of clients that have been selected for local training. Subsequently, each chosen client $\mathit{c}_{i}$ refines the received parameter $\mathbf{W}_{t}$ using an optimizer operating on its private dataset $\mathcal{D}^{i}$. Following this, the selected clients upload the updated model parameter $\mathbf{W}_{t}^{i}$, and the central server aggregates the local model parameters to obtain the enhanced global model parameter $\mathbf{W}_{t+1}$.

In FedGNN setting, there exist diverse scenarios involving distributed graphs that are motivated by real-world applications. In these scenarios, classification tasks can be classified into two distinct settings based on how graphs are distributed across clients.
\textbf{Node-level FedGNN}. Each client is equipped with a subgraph, and the prevalent tasks involve node classification.  Real-world applications, such as social networks, demonstrate situations where relationships between nodes can span across different clients, and each node possesses a unique label. 
\textbf{Graph-level FedGNN}.  Each client possesses a set of graphs, and the primary focus lies on graph classification tasks. Real-world applications, such as protein discovery, exemplify instances where each institution holds a limited graph along with associated labels.

%review 0.75
%public 1.0
%\setlength{\belowcaptionskip}{-0.6cm}
\section{A Unified Framework for Classification Backdoor Attack on FedGNN}
\begin{figure}[t]
    \centering
\includegraphics[width=1.0\textwidth]{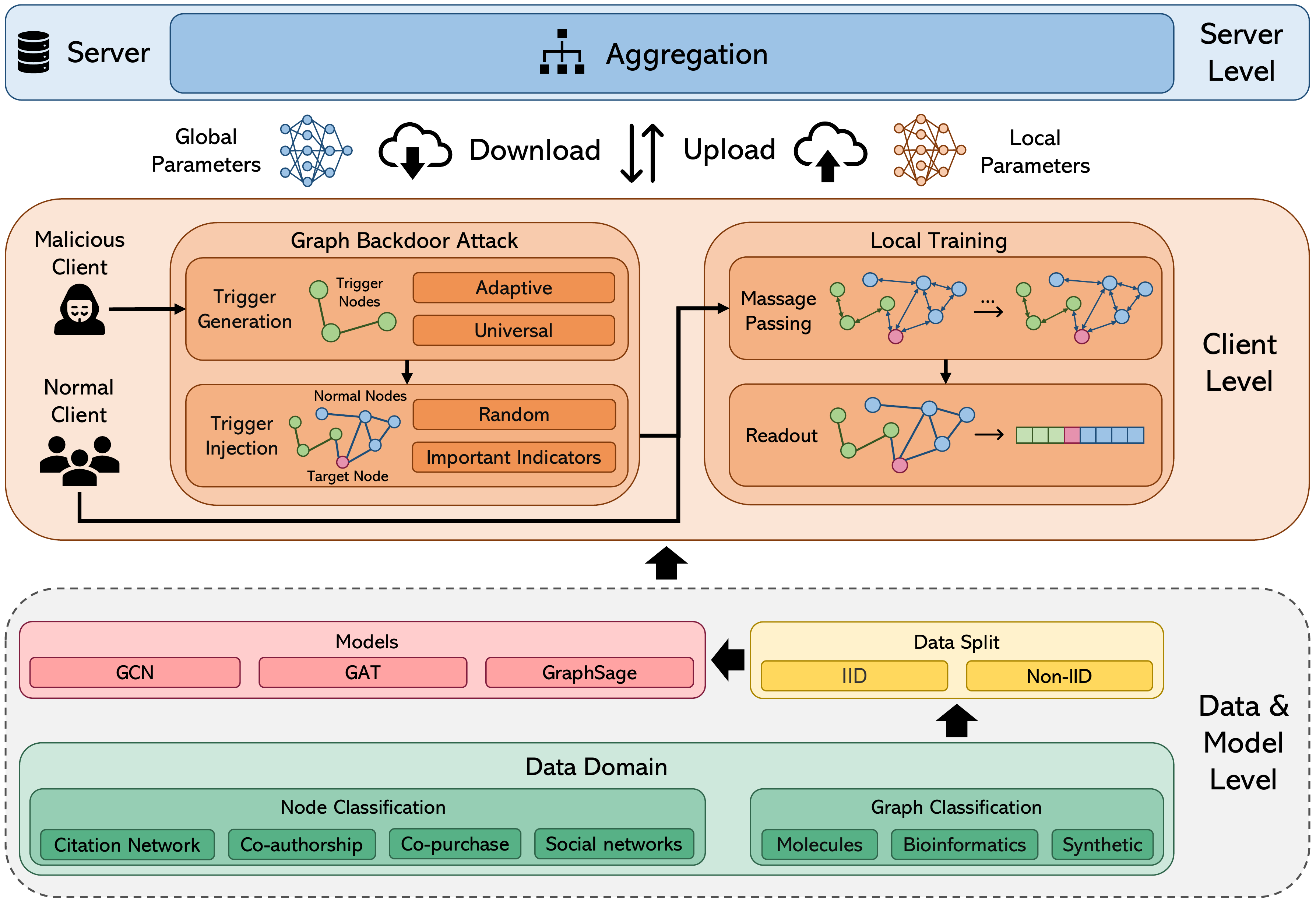}
\setlength{\belowcaptionskip}{-0.0cm}
\caption{ A unified framework for classification backdoor attack on FedGNN.}\label{fig:benchmark}
\end{figure}

This section presents a unified framework for classification backdoor attacks on federated  GNNs. Our primary focus is on graph-based backdoor attacks, where malicious entities strategically insert triggers into graphs or subgraphs to compromise the trustworthiness of FedGNN. A comprehensive illustration of our unified framework for classification backdoor attacks on FedGNN can be found in Figure~\ref{fig:benchmark}. In detail, we first introduce the dataset and models and then give the evaluation metric, then introduce the threat model. Next, we introduce the federated graph backdoor attack, which involves the formulation of the attack goal and a two-step attack process: trigger generation and trigger injection. Finally, we explore various critical factors at both global and local levels.

\subsection{Datasets and Models}

In this study, we have considered six distinct domains comprising a total of thirteen datasets, along with three widely used GNNs. \textit{Node-level Datasets:}
For node-level analysis, we have included three extensively studied citation graphs, such as Cora, CiteSeer, and PubMed. Additionally, we have incorporated the Co-authorship graphs (CS and Physics), along with the Amazon Co-purchase graphs (Photo and Computers). 
\textit{Graph-level Datasets:}
For graph-level analysis, we have utilized molecular graphs such as AIDS and NCI1. Furthermore, bioinformatics graphs, including PROTEINS-full, DD, and ENZYMES, have been incorporated. Lastly, a synthetic graph, COLORS-3, has also been employed.
\textit{Models:}
We have employed three widely adopted GNNs: GCN, GAT, and GraphSage, which have been demonstrated effective in various graph-based tasks. For detailed statistical information about the graphs used, please refer to Appendix~\ref{sec:app-data}.

\subsection{Evaluation Metrics}
To assess the effectiveness of the graph backdoor attack on FedGNN, three metrics are employed: the average clean accuracy (ACC) across all clients, the average attack success rate (ASR) on malicious clients, and the transferred attack success rate (TAST) on 
normal clients. The ACC metric evaluates the performance of federated  GNNs when exposed to clean examples from all clients. The ASR metric measures the performance of the graph backdoor attack specifically on the malicious clients. Lastly, the TAST metric gauges the vulnerability of 
normal clients to the graph backdoor attack. For the detailed equations corresponding to these metrics, please refer to Appendix~\ref{sec:app-eva}.

\subsection{Threat Model}

\textbf{Attack Objective.} Assuming there are a total of $K$ clients, with $M$ ($M \leq K$) of them being malicious, each malicious attacker independently conducts the backdoor attack on their own models.  The primary goal of a backdoor attack is to manipulate the model in such a way that it misclassifies specific pre-defined labels (known as target labels) only within the poisoned data samples. It is important to ensure that the model's accuracy remains unaffected when processing clean data. \textbf{Attack Knowledge.} In this setting, we assume that the malicious attacker has complete knowledge of their own training data. They have the capability to generate triggers. It should be noted that this scenario is quite practical since the clients have full control over their own data. \textbf{Attacker Capability.} The malicious client has the ability to inject triggers into the training datasets, but this capability is limited within predetermined constraints such as trigger size and poisoned data rate. The intention is to contaminate the training datasets. However, the malicious client lacks the ability to manipulate the server-side aggregation process or interfere with other clients' training processes and models.

%\subsection{Federated Graph Backdoor Attack}

\subsection{Federated Graph Backdoor Attack}

Mathematically, the formal attack objective for each malicious client $\mathit{c}_i$ during round $t$ can be defined as follows,

\begin{equation}
\begin{aligned}
 & \mathbf{W}_{t}^{i*}= \arg \min_{\mathbf{W_t}^{i}} \frac{1}{N_{i}} \left [ \sum_{j \in \mathcal{D}_{p}^{i}}\mathcal{L}((f_{i}(\mathbf{X}_{j}^{i}, g_{\tau}\circ \mathbf{A}_{j}^{i} ;\mathbf{W}_{t-1}^{i}),\tau)+\sum_{j \in \mathcal{D}_{c}^{i}}\mathcal{L}((f_{i}(\mathbf{X}_{j}^{i},\mathbf{A}_{j}^{i};\mathbf{W}_{t-1}^{i}),\mathbf{Y}_{j}^{i})\right ],\\ 
 &\forall j \in \mathcal{D}_{p}^{i}, N_{\tau} =|g_{\tau}| \leq  \triangle_{g} \quad  \text{and} \quad \rho = \frac{|\mathcal{D}_{p}^{i}|}{|\mathcal{D}^{i}|} \leq  \triangle_{p},
 \end{aligned}
\label{eq:attack_goal}
\end{equation}

where $\mathcal{D}_{p}^{i}$ refers to the set of poisoned data and $\mathcal{D}_{c}^{i}$ corresponds to the clean dataset. Noted that $\mathcal{D}_{p}^{i} \sqcup \mathcal{D}_{c}^{i} = \mathcal{D}^{i}$ and $\mathcal{D}_{p}^{i} \sqcap \mathcal{D}_{c}^{i} = \phi$, indicating the union and intersection of the poisoned and clean data sets, respectively. $g_{\tau}\circ \mathbf{A}_{j}^{i}$ represents the poisoned graph resulting from an attack. $g_{\tau}$ represents the trigger generated by the attacker, which is then embedded into the clean graph, thereby contaminating the datasets.  Additionally, $\tau$ denotes the target label.   $ N_{\tau}= |g_{\tau}|$ denotes the trigger size and  $ \triangle_{g}$  represents the  constrain to ensures\ that the trigger size remains within the specified limit.  $\rho = \frac{|\mathcal{D}{p}^{i}|}{|\mathcal{D}^{i}|}$ represents the poisoned rate, and $\triangle{p}$ denotes the budget allocated for poisoned data.

In the federated graph backdoor attack, to generate the trigger and poisoned data sets, the graph backdoor attack can be divided into two steps: trigger generation and trigger injection.  The term "trigger" (a specific pattern) has been formally defined as a subgraph in the work by Zhang \etal (2021), providing a clear and established framework for its characterization~\cite{zhang2021backdoor}.

\textbf{Trigger Generation.} The process of trigger generation can be defined as the function $\varphi(\mathbf{X}_{j}^{i},\mathbf{A}_{j}^{i})$, which yields the generated trigger $g_{\tau     }$ through $\varphi(\mathbf{X}_{j}^{i},\mathbf{A}_{j}^{i})=g_{\tau}$.

\textbf{Trigger Injection.} The process of trigger injection can be defined as the function $a(g_{\tau},\mathbf{A}_{j}^{i})$, which generates the final poisoned graph $g_{\tau}\circ \mathbf{A}_{j}^{i}$ by incorporating the trigger $g_{\tau}$ into the pristine graph $\mathbf{A}_{j}^{i}$.

\subsection{Factors in Federated Graph Backdoor}
The  graph backdoor attack framework in FedGNN encompasses various critical factors that warrant exploration. These factors can be categorized into two levels: the global level and the local level. At the global level, factors such as data distribution, the number of malicious attackers, the start time of backdoor attacks, and overlapping rate play significant roles. On the other hand, the local level involves parameters like trigger size, trigger type, trigger position, and poisoning rate. Notably, the overlapping  rate holds particular importance in node-level FedGNN, as it involves cross-nodes across multiple clients.

\textit{Global Level Factors:}
\textbf{Data Distribution.} The data distribution encompasses two distinct types: independent and identically distributed (IID) and non-independent and identically distributed (Non-IID).
In detail,  IID refers to  data distribution among clients remaining constant, while Non-IID (L-Non-IID~\cite{wangfed2022, zhang2021subgraph}, PD-Non-IID~\cite{FangCJG20}, N-Non-IID~\cite{li2022federated}) refers that the data distribution among clients exhibiting variations. \textbf{Number of Malicious
Attackers.}  The concept of the number of malicious attackers, denoted as $M$, can be defined in the following manner. Let us assume that the set of malicious clients is denoted as $\mathcal{C}_{m}$, and the set of normal clients is denoted as $\mathcal{C}_{n}$. It can be inferred that $\mathcal{C}_{m} \sqcup \mathcal{C}_{n} = \mathcal{C}$ and $\mathcal{C}_{m} \sqcap \mathcal{C}_{c} = \phi$. 
\textbf{Attack Time.} In the context of FL, the  attack time denotes the precise moment when a malicious attack is launched.  The attack time can be denoted by  $t^{*}$.
\textbf{Overlapping Rate (specific to Node-level FedGNN).}
The overlapping rate, represented by the variable $\alpha$, pertains to the proportion of additional samples of overlapping data that across clients. This phenomenon arises in node-level FedGNN, where cross-client nodes exist, resulting in the sharing of common data samples between different clients.

\textit{Local Level Factors:}
\textbf{Trigger Size.} The size of the trigger can be  quantified by counting the number of nodes within the corresponding graph. The trigger size is denoted by $N_{\tau}$.
\textbf{Trigger Type.} Based on the methods used to generate triggers(\eg Renyi~\cite{zhang2021backdoor}, WS~\cite{watts1998collective}, BA~\cite{barabasi1999emergence}, RR~\cite{steger1999generating}, GTA~\cite{xi2021graph}, and UGBA~\cite{daiunnoti2023} \etc), the categorization of trigger types can be refined into two  categories: universal triggers and adaptive triggers. Universal triggers are pre-generated through graph generation techniques, such as the Erdős-Rényi (ER) model~\cite{gilbert1959random}, which are agnostic to the underlying graph datasets. On the other hand, adaptive triggers are specifically designed for individual graphs using optimization methods.
\textbf{Trigger Position.} The trigger position refers to the specific location within a graph or sub-graph where the trigger is injected. Typically, the trigger position can be categorized into two types: random position and important indicator position. In the case of the random position, the trigger is injected into the graph in a random manner without any specific consideration. Conversely, the important indicator position entails injecting the trigger based on certain crucial centrality values, such as the degree or cluster-based scores, that indicate the significance of specific nodes within the graph.
\textbf{Poisoning Rate.}  The concept of poisoning rate, denoted as $\rho$, can be defined as the ratio of the cardinality of the set of poisoned data samples, $\mathcal{C}_{p}^{i}$, to the total number of data samples, denoted as $\mathcal{D}^{i}$. Mathematically, this can be expressed as $\rho = \frac{|\mathcal{D}_{p}^{i}|}{|\mathcal{D}^{i}|}$, where $\forall \mathit{c}_i \in \mathcal{C}$ signifies that the cardinality calculations are performed for every client $\mathit{c}_i$ belonging to the set $\mathcal{C}$.

\begin{table}[t]
\setlength{\abovecaptionskip}{-1.2cm}
\caption{Critical factors in federated graph backdoor.}
\scalebox{0.75}{\label{tab:fl_adversarial_params}
\begin{threeparttable}  
\begin{tabular}{clc|cc}
\hline
             & Factors                       & Symbol     & \multicolumn{1}{c|}{Node Level}                  & Graph Level                             \\ \hline
\multirow{4}{*}{Global Level} & Data Distribution             & -          & \multicolumn{1}{c|}{ $\{$IID$^{*}$, L-Non-IID$\}$}             & $\{$IID$^{*}$, PD-Non-IID,  N-Non-IID $\}$ \\ \cline{4-5} 
             & \#  of Malicious Attackers & $M$        & \multicolumn{2}{c}{$\{1^*,2,3,4,5\}$}                                                          \\ \cline{4-5} 
             &  Attack Time & $t^{*}$    & \multicolumn{2}{c}{$T * \{0.0^{*},0.1,0.2,0.3,0.4,0.5\}$}                                          \\ \cline{4-5} 
             & Overlapping Rate              & $\alpha$   & \multicolumn{1}{c|}{$\{0.1^{*},0.2,0.3,0.4,0.5\}$} & -                                       \\ \hline
\multirow{4}{*}{Local Level}  & Trigger Size                  & $N_{\tau}$ & \multicolumn{1}{c|}{$\{3^{*},4,5,6,7,8,9, 10\}$}           & $N_{d} *\{ 0.1^{*},0.2,0.3,0.4,0.5\}$       \\ \cline{4-5} 
             & Trigger Type                  & $g_{\tau}$ & \multicolumn{1}{c|}{ $\{$Renyi$^{*}$, WS, BA, GTA,  UGBA $\}$}  & $\{$ Renyi$^{*}$, WS, BA, RR, GTA   $\}$         \\ \cline{4-5} 
             & Trigger Position              & -          & \multicolumn{2}{c}{ $\{$Random$^{*}$, Degree,  Cluster $\}$}                                            \\ \cline{4-5} 
             & Poisoning Rate                & $\rho$     & \multicolumn{2}{c}{$\{0.1^{*}, 0.2, 0.3, 0.4, 0.5\}$}                                            \\ \hline
\end{tabular}
       \begin{tablenotes}    %这行要添加， 从这开始
        \footnotesize               %这行要添加
        \item    $^*$ marks the default value and  \# represents the Number.
        \item $T$ represents the total training round time and $N_{d}$ represents the average number of graph nodes.
      \end{tablenotes}            %这行要添加
\end{threeparttable} 
}
\centering
\end{table}

%\tnote{2}

\section{Experimental Studies}

In this section, we present the experimental studies conducted to investigate classification backdoor attacks on FedGNN. Our main objective is to evaluate the impact of graph backdoor attacks on FedGNN covering both the node and graph level tasks. Additionally, we aim to explore the critical factors that influence the effectiveness of graph backdoor attacks on FedGNN, considering aspects from both the global and local levels. \eat{The structure of this section is organized as follows. Firstly, we provide an overview of the evaluation process, encompassing the datasets and models, the evaluation metrics, and the configurations of the influencing factors. Subsequently, we present the experimental results and detailed analyses pertaining to the graph backdoor attack on FedGNN. Finally, we conduct experimental studies to examine the impact of various factors on the efficacy of such attacks. }

\subsection{Experimental Settings}

\textbf{Factors Settings.} We present the detailed factors setup considered in our study. It is important to note that the first value presented represents the default setting. To assess the individual impact of each factor, we keep the remaining factors fixed while systematically varying the corresponding values in our experiments.  The factors range is shown in Table~\ref{tab:fl_adversarial_params}. For the detailed setting for factor, please refer to Appendix~\ref{sec:app-factors}.

\textbf{Federated Graph Backdoor Attack.} The federated graph backdoor attack can be characterized by the combination of trigger generation techniques (Renyi~\cite{zhang2021backdoor}, WS~\cite{watts1998collective}, BA~\cite{barabasi1999emergence}, RR~\cite{steger1999generating}, GTA~\cite{xi2021graph}, and UGBA~\cite{daiunnoti2023}) and trigger position strategies (Random, Degree, and Cluster). For instance, the attack method Renyi-Random refers to the utilization of the ER model to generate the trigger, which is then randomly injected into the graph. 

\textbf{Implementation Details.}
Our implementation of the backdoor attack on FedGNN is based on the PyTorch framework. The experiments were carried out on two server configurations: three Linux Centos Servers, each with 4 RTX 3090 GPUs, and two Linux Ubuntu Servers, each with 2 V100 GPUs. In both node-level and graph-level tasks, we adopt the inductive learning settings as outlined in~\cite{xu2022more, daiunnoti2023}.
For each dataset, we ensure consistent experimental conditions by employing the same training and attack settings. We set the total number of clients to $5$, and all clients participate in the training process at each round. Each experiment is repeated five times. For a detailed description of the training and attack settings, please refer to Appendix~\ref{sec:implement}.

% \begin{figure}[t]
%     \centering
%     \includegraphics[width=0.9\linewidth]{figs/Main-Comparison-GCN.pdf}
%     \setlength{\belowcaptionskip}{-0.5cm}
%   \caption{Graph backdoor attack on both node and graph level tasks for GCN. (Color intensity corresponds to value magnitude)}\label{figs:Main-Citeseer-AIDS-GCN}
% \end{figure}

\begin{figure}[t]
    \centering
    \includegraphics[width=1.0\linewidth]{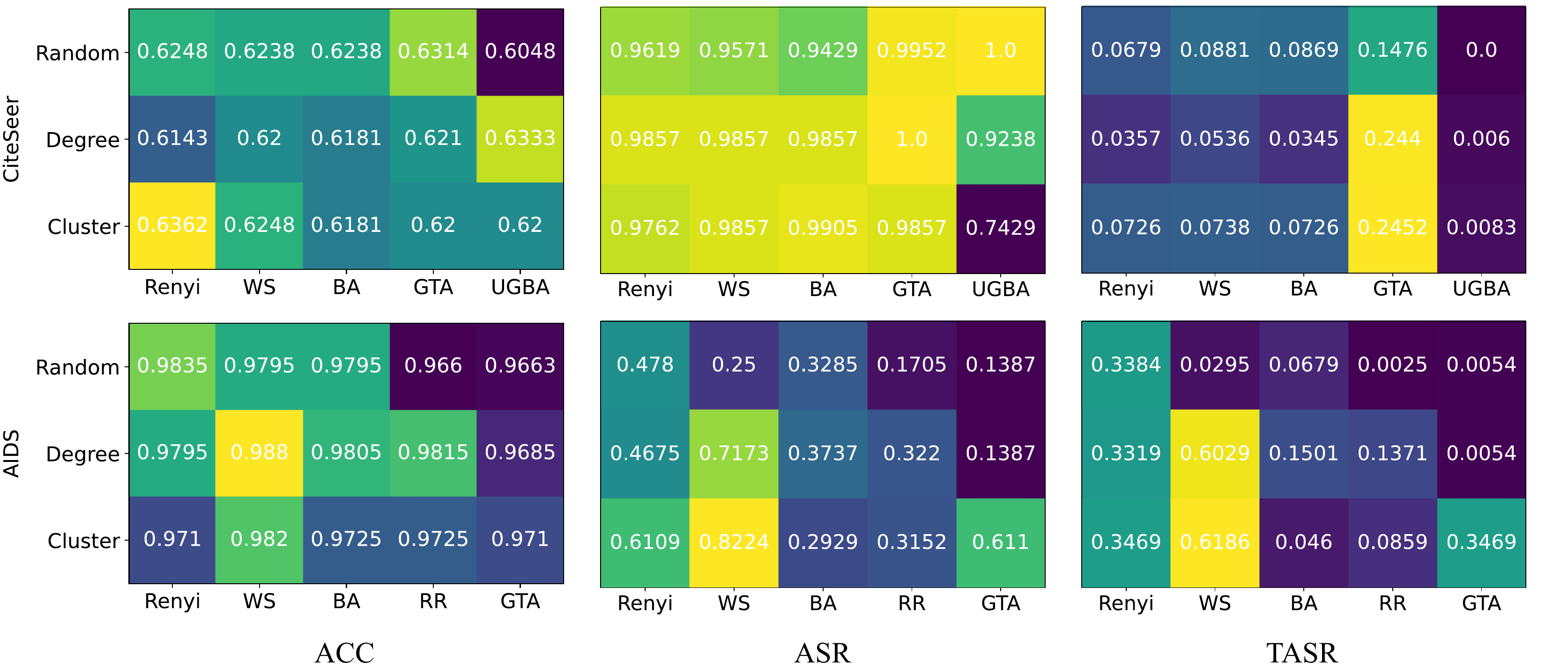}
    \setlength{\belowcaptionskip}{-0.0cm}
  \caption{Graph backdoor attack on both node and graph level tasks for GCN. (Color intensity corresponds to value magnitude)}\label{figs:Main-Citeseer-AIDS-GCN}
\end{figure}

\subsection{Benchmark Results of Graph Backdoor Attack on FedGNN}

The results of the benchmark for the graph backdoor attack on FedGNN are presented in Figure~\ref{figs:Main-Citeseer-AIDS-GCN}.  The observations are summarized as follows. (1) The node-level task exhibits higher vulnerability to attacks compared to the graph-level task at a relatively small trigger size. Specifically, a significant majority of graph backdoor attacks achieve an ASR (Attack Success Rate) exceeding $90\%$, while the highest ASR recorded at the graph level is $82.24\%$.
(2) Despite not being intentionally poisoned by malicious attackers, the
normal clients are still susceptible to graph backdoor attacks. For instance, in the node-level task, there is a TASR (Transfered Attack Success Rate) of $24.52\%$, while the graph-level task exhibits even higher vulnerability with a TASR of $61.86\%$. This observation suggests that the weights uploaded by the malicious clients can inadvertently influence the 
normal clients when they download the global model's weights. 3). The combination of trigger size and trigger position significantly influences the attack performance on the graph-level task compared to the node-level task. For instance, the attack WS-Cluster achieves an ASR of approximately $82.24\%$, while the GTA-Random achieves only about $13.87\%$. Due to the page limit, the  benchmark results on other datasets and models please refer to Appendix~\ref{sec:app-benchmark}.

% \vspace{-5pt}
% \subsection{Factors in Federated GNN}
% \vspace{-5pt}

\subsection{Factors in Federated GNN}

\begin{figure}[t]
    \centering
    \includegraphics[width=0.95\linewidth]{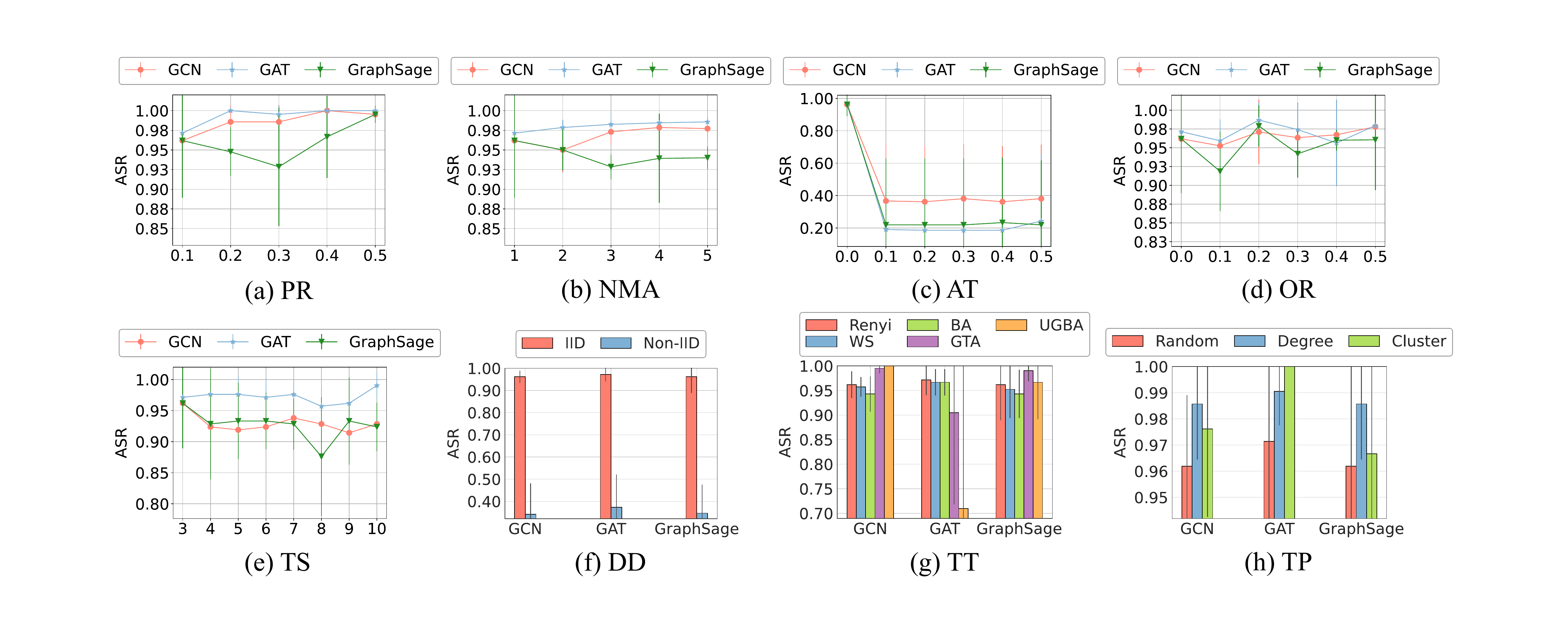}
    \setlength{\abovecaptionskip}{0.0cm}
    \setlength{\belowcaptionskip}{-0.4cm}
  \caption{Node-level task factors.}\label{figs:node-Citeseer-AIDS-factors}
\end{figure}

\begin{figure}[t]
    \centering
    \includegraphics[width=0.95\linewidth]{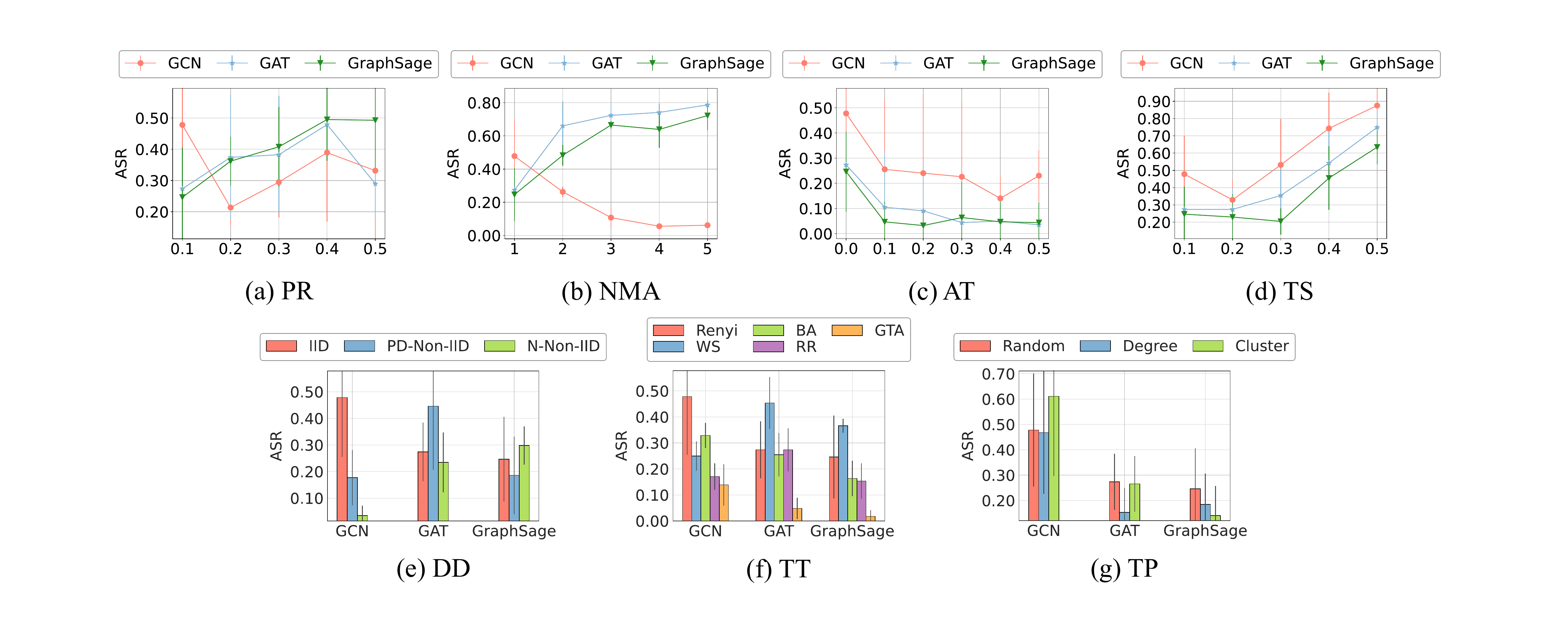}
    \setlength{\abovecaptionskip}{0.0cm}
    \setlength{\belowcaptionskip}{-0.4cm}
  \caption{Graph-level task factors.}\label{figs:graph-Citeseer-AIDS-factors}
\end{figure}

The overall results of factors can be shown in Figures~\ref{figs:node-Citeseer-AIDS-factors}-\ref{figs:graph-Citeseer-AIDS-factors}. 
\textit{Global Level Factors:}
\textbf{Data Distribution (DD).} For node-level tasks, there models trained on IID data are more vulnerable than models trained on Non-IID data. For graph-level tasks,  the GCN trained on IID data are more vulnerable than models trained on Non-IID data (PD-Non-IID and N-Non-IID), while GAT and GraphSagr trained on Non-IID data are more vulnerable than models trained on IID data. \textbf{Number of Malicious Attackers (NMA).}
For node-level tasks, an increase in NMA leads to an increase in ASR for both GCN and GAT models. Conversely, an increase in NMA results in a decrease in ASR for both GraphSage. Concerning  graph-level tasks, the ASR demonstrates an increase with the increase of NMA in the case of GAT and GraphSage. However, in the scenario of GCN, the ASR shows a decrease with the increase of NMA. \textbf{Attack Time (AT).} For both node-level and graph-level tasks, an increase in AT results in a decrease in ASR for three models. \textbf{Overlapping Rate (OR).} 
The ASR demonstrates an upward trend as the overlapping rate increases. This correlation can be attributed to the possibility that overlapping nodes facilitate the backdooring of normal clients, primarily through the presence of cross-edges.

\textit{Local Level Factors:}
\textbf{Trigger Size (TS).} For node-level tasks, an increase in TS leads to an increase in ASR for GCN. However, in the case of GAT and GraphSage, the ASR demonstrates a decrease with the increase of TS. Concerning the graph-level task, the ASR shows an increase with the increase of TS across all three  GNNs.
\textbf{Trigger Types (TT).} In the node-level task, the adaptive trigger demonstrates a higher ASR on most models. Conversely, in the graph-level task, the universal trigger exhibits higher ASR. \eat{For instance, the GTA achieves a maximum ASR of 100\% on the CiteSeer dataset, along with a TASR of 24.52\%. On the AIDS graph dataset, the WS method achieves the highest ASR of 98.80\% and TASR of 61.86\%.}
\textbf{Trigger Position (TP).} In node-level tasks, we observed a significantly large ASR when using importance-based positions (Degree and Cluster) compared to random positions. However, for the graph-level task, while importance-based positions showed higher ASR for GCN, random positions yielded higher ASR for GAT and GraphSage.  \textbf{Poisoning Rate (PR).} On node classification, an increase in PR results in a slight decrease in ASR. However, graph classification exhibits an upward trend in ASR.
Due to the page limit, the results on other datasets and metrics, please refer to  Appendix~\ref{sec:app-impact-factors}.
\section{Related Works}
\eat{This approach exhibits two distinct settings based on the underlying structural information within the graph~\cite{fu2022federated}. The initial setting involves each party accessing  the graph data with the objective of developing a model capable of making accurate predictions. The second setting revolves around leveraging structural information among clients, leading to the formation of a client-level graph. Such structural information proves valuable in training more effective federated learning models.}

\textbf{FedGNN.} FedGNN present a distributed machine learning paradigm that facilitates collaborative training of GNNs among multiple parties, ensuring the privacy of their sensitive data. In recent years, extensive research has been conducted on FedGNN, with a particular focus on addressing security concerns~\cite{chen2022graph, xu2022more, abs-2207-05521, XuBZAKL22}. Among these concerns, poisoning attacks have garnered significant attention, encompassing both data poisoning attacks and model poisoning attacks. Data poisoning attacks occur when an adversary employs tainted data to train the local model, while model poisoning attacks involve manipulation of either the training process or the local model itself. Currently, the majority of attacks on FedGNN primarily concentrate on data poisoning attacks. Chen \etal~\cite{chen2022graph} proposed adversarial attacks on vertical federated learning, utilizing adversarial perturbations on global node embeddings based on gradient leakage from pairwise nodes. Additionally, Xu \etal~\cite{xu2022more} investigated centralized and distributed backdoor attacks on FedGNN.

\textbf{Graph Backdoor Attacks.}
Backdoor attacks on GNNs have received significant attention in recent years~\cite{zhang2021backdoor, xu2023rethinking, yang2022transferable, xu2021explainability, xi2021graph, zheng2023motif, daiunnoti2023}. Regarding graph backdoor attacks, they can be classified into two types based on the employed trigger: universal graph backdoor attacks and adaptive backdoor attacks. In universal graph backdoor attacks, Zhang \etal ~\cite{zhang2021backdoor} generated sub-graphs using the Erdős-Rényi (ER) model as triggers and injected them into the training data. Additionally, Xu \etal~\cite{xi2021graph} observed that the position of the trigger injection into the graph can also affect the attack's performance.
As for adaptive trigger backdoor attacks, Xi \etal~\cite{xi2021graph} developed an adaptive trigger generator that optimizes the attack's effectiveness for both transductive and inductive tasks. \eat{Dai \etal (2023) proposed an adaptive trigger generation method using optimization to create more effective and inconspicuous triggers, with a primary focus on node-level tasks.} \eat{While the adaptive method allows tailoring the trigger to specific graphs or nodes, it requires additional information, such as graph structure and computation resources, to optimize the trigger based on feedback from the surrogate model.} In our benchmark, we focus primarily on data poisoning attacks. While model poisoning attacks can be effective, data poisoning attacks may be more convenient because they do not require tampering with the model learning process, and they allow non-expert actors to participate~\cite{tolpegin2020data}.

\section{Conclusions and Open Problems}
\textbf{Conclusions.} In this paper, we proposed a unified framework for classification backdoor attacks on FedGNN. We then  introduced the critical factors involved in graph backdoor attacks on FedGNN, including both global and local level factors. Along this line, we performed approximately 8,000 experiments on the graph backdoor attacks benchmark and conduct critical factor experiments to provide a comprehensive analysis.

\textbf{Open Problems.} (1) Enhancing the success rate of transferred attacks: Our findings reveal that malicious attackers can also backdoor 
normal clients through the FL mechanism. However, there is a need to explore methods that can identify and exploit the worst vulnerabilities under these circumstances.
(2) Evaluating the defense method under backdoor  attack: We demonstrate that FedGNN can be compromised by malicious attackers. However, assessing the effectiveness of defense mechanisms against such attacks still requires further exploration.
(3) Cooperative malicious attackers: Currently, the majority of malicious attackers operate independently during the attack process, neglecting the potential benefits of collaboration. An intriguing research direction lies in investigating the utilization of collaboration to enhance attack performance.

\begin{ack}
This research was supported in part by the National Natural Science Foundation of China under Grant No.62102110, Guangzhou Science and Technology Plan Guangzhou-HKUST(GZ) Joint Project No. 2023A03J0144, and Foshan HKUST Projects (FSUST21-FYTRI01A, FSUST21-FYTRI02A).
\end{ack}

\bibliographystyle{plainnat}

\begin{thebibliography}{10}

\bibitem{MaekawaN}
Seiji Maekawa, Koki Noda, Yuya Sasaki, and Makoto Onizuka.
\newblock Beyond real-world benchmark datasets: An empirical study of node
  classification with gnns.
\newblock In {\em NeurIPS}, 2022.

\bibitem{LiuDZDHZDCPSSLC22}
Kay Liu, Yingtong Dou, Yue Zhao, Xueying Ding, Xiyang Hu, Ruitong Zhang, Kaize
  Ding, Canyu Chen, Hao Peng, Kai Shu, Lichao Sun, Jundong Li, George~H. Chen,
  Zhihao Jia, and Philip~S. Yu.
\newblock {BOND:} benchmarking unsupervised outlier node detection on static
  attributed graphs.
\newblock In {\em NeurIPS}, 2022.

\bibitem{zhang2022multi}
Weijia Zhang, Hao Liu, Jindong Han, Yong Ge, and Hui Xiong.
\newblock Multi-agent graph convolutional reinforcement learning for dynamic
  electric vehicle charging pricing.
\newblock In {\em Proceedings of the 28th ACM SIGKDD conference on knowledge
  discovery and data mining}, pages 2471--2481, 2022.

\bibitem{ST22}
Xikun Zhang, Dongjin Song, and Dacheng Tao.
\newblock {CGLB:} benchmark tasks for continual graph learning.
\newblock In {\em NeurIPS}, 2022.

\bibitem{GuiLWJ22}
Shurui Gui, Xiner Li, Limei Wang, and Shuiwang Ji.
\newblock {GOOD:} {A} graph out-of-distribution benchmark.
\newblock In {\em NeurIPS}, 2022.

\bibitem{QinZWZZ22}
Yijian Qin, Ziwei Zhang, Xin Wang, Zeyang Zhang, and Wenwu Zhu.
\newblock Nas-bench-graph: Benchmarking graph neural architecture search.
\newblock In {\em NeurIPS}, 2022.

\bibitem{DuanLWZZCHW22}
Keyu Duan, Zirui Liu, Peihao Wang, Wenqing Zheng, Kaixiong Zhou, Tianlong Chen,
  Xia Hu, and Zhangyang Wang.
\newblock A comprehensive study on large-scale graph training: Benchmarking and
  rethinking.
\newblock In {\em NeurIPS}, 2022.

\bibitem{Huang0W0ZXCV22}
Xuanwen Huang, Yang Yang, Yang Wang, Chunping Wang, Zhisheng Zhang, Jiarong Xu,
  Lei Chen, and Michalis Vazirgiannis.
\newblock Dgraph: {A} large-scale financial dataset for graph anomaly
  detection.
\newblock In {\em NeurIPS}, 2022.

\bibitem{he2021spreadgnn}
Chaoyang He, Emir Ceyani, Keshav Balasubramanian, Murali Annavaram, and Salman
  Avestimehr.
\newblock Spreadgnn: Decentralized multi-task federated learning for graph
  neural networks on molecular data, 2021.

\bibitem{NEURIPS2020_b8ffa41d}
Hongyi Wang, Kartik Sreenivasan, Shashank Rajput, Harit Vishwakarma, Saurabh
  Agarwal, Jy-yong Sohn, Kangwook Lee, and Dimitris Papailiopoulos.
\newblock Attack of the tails: Yes, you really can backdoor federated learning.
\newblock In {\em In NeurIPS}, volume~33, pages 16070--16084, 2020.

\bibitem{bagdasaryan2020backdoor}
Eugene Bagdasaryan, Andreas Veit, Yiqing Hua, Deborah Estrin, and Vitaly
  Shmatikov.
\newblock How to backdoor federated learning.
\newblock In {\em International Conference on Artificial Intelligence and
  Statistics}, pages 2938--2948. PMLR, 2020.

\bibitem{Xie0CL21}
Chulin Xie, Minghao Chen, Pin{-}Yu Chen, and Bo~Li.
\newblock {CRFL:} certifiably robust federated learning against backdoor
  attacks.
\newblock In {\em Proceedings of the 38th International Conference on Machine
  Learning, {ICML} 2021, 18-24 July 2021, Virtual Event}, volume 139 of {\em
  Proceedings of Machine Learning Research}, pages 11372--11382. {PMLR}, 2021.

\bibitem{li2023learning}
Henger Li, Chen Wu, Senchun Zhu, and Zizhan Zheng.
\newblock Learning to backdoor federated learning.
\newblock {\em arXiv preprint arXiv:2303.03320}, 2023.

\bibitem{OzdayiKG21}
Mustafa~Safa {\"{O}}zdayi, Murat Kantarcioglu, and Yulia~R. Gel.
\newblock Defending against backdoors in federated learning with robust
  learning rate.
\newblock In {\em Thirty-Fifth {AAAI} Conference on Artificial Intelligence,
  {AAAI} 2021}, pages 9268--9276. {AAAI} Press, 2021.

\bibitem{chen2022graph}
Jinyin Chen, Guohan Huang, Haibin Zheng, Shanqing Yu, Wenrong Jiang, and Chen
  Cui.
\newblock Graph-fraudster: Adversarial attacks on graph neural network-based
  vertical federated learning.
\newblock {\em IEEE Transactions on Computational Social Systems}, 2022.

\bibitem{xu2022more}
Jing Xu, Rui Wang, Stefanos Koffas, Kaitai Liang, and Stjepan Picek.
\newblock More is better (mostly): On the backdoor attacks in federated graph
  neural networks.
\newblock In {\em Proceedings of the 38th Annual Computer Security Applications
  Conference}, pages 684--698, 2022.

\bibitem{abs-2207-05521}
Anisa Halimi, Swanand Kadhe, Ambrish Rawat, and Nathalie Baracaldo.
\newblock Federated unlearning: How to efficiently erase a client in {FL}?
\newblock {\em CoRR}, abs/2207.05521, 2022.

\bibitem{XuBZAKL22}
Runhua Xu, Nathalie Baracaldo, Yi~Zhou, Ali Anwar, Swanand Kadhe, and Heiko
  Ludwig.
\newblock Detrust-{FL}: Privacy-preserving federated learning in decentralized
  trust setting.
\newblock In {\em {IEEE} 15th International Conference on Cloud Computing,
  {CLOUD} 2022, Barcelona, Spain, July 10-16, 2022}, pages 417--426. {IEEE},
  2022.

\bibitem{GilmerSRVD17}
Justin Gilmer, Samuel~S. Schoenholz, Patrick~F. Riley, Oriol Vinyals, and
  George~E. Dahl.
\newblock Neural message passing for quantum chemistry.
\newblock In {\em Proceedings of the 34th International Conference on Machine
  Learning, {ICML} 2017, Sydney, NSW, Australia, 6-11 August 2017}, volume~70
  of {\em Proceedings of Machine Learning Research}, pages 1263--1272. {PMLR},
  2017.

\bibitem{RongXHHC0WDWM20}
Yu~Rong, Tingyang Xu, Junzhou Huang, Wenbing Huang, Hong Cheng, Yao Ma, Yiqi
  Wang, Tyler Derr, Lingfei Wu, and Tengfei Ma.
\newblock Deep graph learning: Foundations, advances and applications.
\newblock In {\em {KDD} '20: The 26th {ACM} {SIGKDD} Conference on Knowledge
  Discovery and Data Mining, Virtual Event, CA, USA, August 23-27, 2020}, pages
  3555--3556. {ACM}, 2020.

\bibitem{GCN}
Thomas~N. Kipf and Max Welling.
\newblock Semi-supervised classification with graph convolutional networks.
\newblock In {\em 5th International Conference on Learning Representations,
  {ICLR} 2017, Toulon, France, April 24-26, 2017, Conference Track
  Proceedings}. OpenReview.net, 2017.

\bibitem{GAT}
Petar Velickovic, Guillem Cucurull, Arantxa Casanova, Adriana Romero, Pietro
  Li{\`{o}}, and Yoshua Bengio.
\newblock Graph attention networks.
\newblock In {\em 6th International Conference on Learning Representations,
  {ICLR} 2018, Vancouver, BC, Canada, April 30 - May 3, 2018, Conference Track
  Proceedings}, 2018.

\bibitem{GraphSage}
William~L. Hamilton, Zhitao Ying, and Jure Leskovec.
\newblock Inductive representation learning on large graphs.
\newblock In {\em Advances in Neural Information Processing Systems 30: Annual
  Conference on Neural Information Processing Systems 2017, December 4-9, 2017,
  Long Beach, CA, {USA}}, pages 1024--1034, 2017.

\bibitem{McMahanMRHA}
Brendan McMahan, Eider Moore, Daniel Ramage, Seth Hampson, and
  Blaise~Ag{\"{u}}era y~Arcas.
\newblock Communication-efficient learning of deep networks from decentralized
  data.
\newblock In {\em Proceedings of the 20th International Conference on
  Artificial Intelligence and Statistics, {AISTATS} 2017, 20-22 April 2017,
  Fort Lauderdale, FL, {USA}}, volume~54 of {\em Proceedings of Machine
  Learning Research}, pages 1273--1282. {PMLR}, 2017.

\bibitem{zhang2021backdoor}
Zaixi Zhang, Jinyuan Jia, Binghui Wang, and Neil~Zhenqiang Gong.
\newblock Backdoor attacks to graph neural networks.
\newblock In {\em Proceedings of the 26th ACM Symposium on Access Control
  Models and Technologies}, pages 15--26, 2021.

\bibitem{wangfed2022}
Zhen Wang, Weirui Kuang, Yuexiang Xie, Liuyi Yao, Yaliang Li, Bolin Ding, and
  Jingren Zhou.
\newblock Federatedscope-gnn: Towards a unified, comprehensive and efficient
  package for federated graph learning.
\newblock In {\em Proceedings of the 28th ACM SIGKDD Conference on Knowledge
  Discovery and Data Mining}, KDD '22, page 4110–4120, New York, NY, USA,
  2022.

\bibitem{zhang2021subgraph}
Ke~Zhang, Carl Yang, Xiaoxiao Li, Lichao Sun, and Siu~Ming Yiu.
\newblock Subgraph federated learning with missing neighbor generation.
\newblock {\em Advances in Neural Information Processing Systems},
  34:6671--6682, 2021.

\bibitem{FangCJG20}
Minghong Fang, Xiaoyu Cao, Jinyuan Jia, and Neil~Zhenqiang Gong.
\newblock Local model poisoning attacks to byzantine-robust federated learning.
\newblock In {\em 29th {USENIX} Security Symposium, {USENIX} Security 2020,
  August 12-14, 2020}, pages 1605--1622. {USENIX} Association, 2020.

\bibitem{li2022federated}
Qinbin Li, Yiqun Diao, Quan Chen, and Bingsheng He.
\newblock Federated learning on non-iid data silos: An experimental study.
\newblock In {\em 2022 IEEE 38th International Conference on Data Engineering
  (ICDE)}, pages 965--978. IEEE, 2022.

\bibitem{watts1998collective}
Duncan~J Watts and Steven~H Strogatz.
\newblock Collective dynamics of ‘small-world’networks.
\newblock {\em nature}, 393(6684):440--442, 1998.

\bibitem{barabasi1999emergence}
Albert-L{\'a}szl{\'o} Barab{\'a}si and R{\'e}ka Albert.
\newblock Emergence of scaling in random networks.
\newblock {\em science}, 286(5439):509--512, 1999.

\bibitem{steger1999generating}
Angelika Steger and Nicholas~C Wormald.
\newblock Generating random regular graphs quickly.
\newblock {\em Combinatorics, Probability and Computing}, 8(4):377--396, 1999.

\bibitem{xi2021graph}
Zhaohan Xi, Ren Pang, Shouling Ji, and Ting Wang.
\newblock Graph backdoor.
\newblock In {\em USENIX Security Symposium}, pages 1523--1540, 2021.

\bibitem{daiunnoti2023}
Enyan Dai, Minhua Lin, Xiang Zhang, and Suhang Wang.
\newblock Unnoticeable backdoor attacks on graph neural networks.
\newblock In {\em Proceedings of the ACM Web Conference 2023}, WWW '23, page
  2263–2273, New York, NY, USA, 2023.

\bibitem{gilbert1959random}
Edgar~N Gilbert.
\newblock Random graphs.
\newblock {\em The Annals of Mathematical Statistics}, 30(4):1141--1144, 1959.

\bibitem{xu2023rethinking}
Jing Xu, Gorka Abad, and Stjepan Picek.
\newblock Rethinking the trigger-injecting position in graph backdoor attack.
\newblock {\em arXiv preprint arXiv:2304.02277}, 2023.

\bibitem{yang2022transferable}
Shuiqiao Yang, Bao~Gia Doan, Paul Montague, Olivier De~Vel, Tamas Abraham,
  Seyit Camtepe, Damith~C Ranasinghe, and Salil~S Kanhere.
\newblock Transferable graph backdoor attack.
\newblock In {\em Proceedings of the 25th International Symposium on Research
  in Attacks, Intrusions and Defenses}, pages 321--332, 2022.

\bibitem{xu2021explainability}
Jing Xu, Minhui Xue, and Stjepan Picek.
\newblock Explainability-based backdoor attacks against graph neural networks.
\newblock In {\em Proceedings of the 3rd ACM Workshop on Wireless Security and
  Machine Learning}, pages 31--36, 2021.

\bibitem{zheng2023motif}
Haibin Zheng, Haiyang Xiong, Jinyin Chen, Haonan Ma, and Guohan Huang.
\newblock Motif-backdoor: Rethinking the backdoor attack on graph neural
  networks via motifs.
\newblock {\em IEEE Transactions on Computational Social Systems}, 2023.

\bibitem{tolpegin2020data}
Vale Tolpegin, Stacey Truex, Mehmet~Emre Gursoy, and Ling Liu.
\newblock Data poisoning attacks against federated learning systems.
\newblock In {\em Computer Security--ESORICS 2020: 25th European Symposium on
  Research in Computer Security, ESORICS 2020, Guildford, UK, September 14--18,
  2020, Proceedings, Part I 25}, pages 480--501. Springer, 2020.

\bibitem{YangCS16}
Zhilin Yang, William~W. Cohen, and Ruslan Salakhutdinov.
\newblock Revisiting semi-supervised learning with graph embeddings.
\newblock In {\em Proceedings of the 33nd International Conference on Machine
  Learning, {ICML} 2016, New York City, NY, USA, June 19-24, 2016}, volume~48
  of {\em {JMLR} Workshop and Conference Proceedings}, pages 40--48. JMLR.org,
  2016.

\bibitem{shchur2018pitfalls}
Oleksandr Shchur, Maximilian Mumme, Aleksandar Bojchevski, and Stephan
  G{\"u}nnemann.
\newblock Pitfalls of graph neural network evaluation.
\newblock {\em Relational Representation Learning Workshop, NeurIPS 2018},
  2018.

\bibitem{McAuley2015}
Julian McAuley, Christopher Targett, Qinfeng Shi, and Anton van~den Hengel.
\newblock Image-based recommendations on styles and substitutes.
\newblock In {\em Proceedings of the 38th International ACM SIGIR Conference on
  Research and Development in Information Retrieval}, SIGIR '15, page 43–52.
  Association for Computing Machinery, 2015.

\bibitem{RiesenB08}
Kaspar Riesen and Horst Bunke.
\newblock {IAM} graph database repository for graph based pattern recognition
  and machine learning.
\newblock In {\em Structural, Syntactic, and Statistical Pattern Recognition,
  Joint {IAPR} International Workshop, {SSPR} {\&} {SPR} 2008, Orlando, USA,
  December 4-6, 2008. Proceedings}, volume 5342 of {\em Lecture Notes in
  Computer Science}, pages 287--297. Springer, 2008.

\bibitem{waleNCI1}
Nikil Wale and George Karypis.
\newblock Comparison of descriptor spaces for chemical compound retrieval and
  classification.
\newblock In {\em Sixth International Conference on Data Mining (ICDM'06)},
  pages 678--689, 2006.

\bibitem{Rossi_Ahmed_2015}
Ryan Rossi and Nesreen Ahmed.
\newblock The network data repository with interactive graph analytics and
  visualization.
\newblock {\em Proceedings of the AAAI Conference on Artificial Intelligence},
  29(1), Mar. 2015.

\bibitem{BorgwardtOSVSK05}
Karsten~M. Borgwardt, Cheng~Soon Ong, Stefan Sch{\"{o}}nauer, S.~V.~N.
  Vishwanathan, Alexander~J. Smola, and Hans{-}Peter Kriegel.
\newblock Protein function prediction via graph kernels.
\newblock In {\em Proceedings Thirteenth International Conference on
  Intelligent Systems for Molecular Biology 2005, Detroit, MI, USA, 25-29 June
  2005}, pages 47--56, 2005.

\bibitem{cheibub2010democracy}
Jos{\'e}~Antonio Cheibub, Jennifer Gandhi, and James~Raymond Vreeland.
\newblock Democracy and dictatorship revisited.
\newblock {\em Public choice}, pages 67--101, 2010.

\bibitem{knyazev2019understanding}
Boris Knyazev, Graham~W Taylor, and Mohamed Amer.
\newblock Understanding attention and generalization in graph neural networks.
\newblock {\em Advances in neural information processing systems}, 32, 2019.

\end{thebibliography}

\clearpage
\appendix

\section{Appendix}

\subsection{Datasets}\label{sec:app-data}
In this study, we consider six distinct domains comprising a total of thirteen datasets. The statistics of them are summarized in Table~\ref{tab:dataStats}.
\subsubsection{Datasets for Node Classification}
\begin{itemize}
    \item \textbf{Citation Network} (Cora, CiteSeer, PubMed): The citation network datasets~\cite{YangCS16} consist of interconnected research papers, where each node represents a study and the edges represent citation relationships between papers. The objective of these datasets is to predict the research field of each study.

    \item \textbf{Co-authorship} (CS, Physics): The co-authorship datasets~\cite{shchur2018pitfalls} are constructed using the Microsoft Academic Graph from the KDD Cup 2016 challenge. In these datasets, nodes represent authors, and the edges indicate co-author connections between authors. The goal is to predict the research fields of the authors.

    \item \textbf{Co-purchase} (Photo, Computers): The co-purchase datasets~\cite{shchur2018pitfalls} are derived from Amazon's co-purchase relations~\cite{McAuley2015}. In these datasets, nodes represent products, and the edges denote co-purchase relationships between pairs of products. The objective is to predict the category of each product.
\end{itemize}

\subsubsection{Datasets for Graph Classification}
\begin{itemize}
    \item \textbf{Molecules} (AIDS, NCI1): The molecules dataset consists of graphs representing chemical compounds. Each graph represents a compound, with nodes representing atoms and edges representing bonds between atoms. The AIDS dataset~\cite{RiesenB08} is used for binary classification, while the NCI1 dataset~\cite{waleNCI1} is for multi-class classification.

    \item \textbf{Bioinformatics} (PROTEINS-full, ENZYMES, DD): The bioinformatics datasets are constructed based on the relationships between organic substances. The PROTEINS-full~\cite{Rossi_Ahmed_2015} and ENZYMES~\cite{BorgwardtOSVSK05} datasets are used to classify proteins as enzymes or non-enzymes. The DD dataset~\cite{cheibub2010democracy}focuses on the classification of pharmacological substances.

    \item \textbf{Synthetic} (COLORS-3): The COLORS-3 dataset~\cite{knyazev2019understanding} involves a color counting task. Random graphs are generated, where each node is assigned one of three colors: red, green, or blue. The goal is to count the number of green nodes in the graph.
\end{itemize}

\begin{table}[h]
 \centering
 \caption{Statistics of datasets.}
  \begin{tabular}{cccccc}
   \toprule
   \textbf{Data Domain} & \textbf{Datasets} & \textbf{\# of Graphs} & \textbf{\# of Nodes} & \textbf{\# of Edges} & \textbf{\# of Classes}\\
   \midrule
   \multirow{6}{*}{Graph Level} 
          & AIDS & 2,000 & 15.69 & 16.20 & 2 \\
   \space & NCI1 & 4,110 & 29.87 & 32.30 & 2 \\
   \space & PROTEINS-full & 1,113 & 39.06 & 72.82 & 2 \\
   \space & ENZYMES & 600 & 32.63 & 64.14 & 2 \\
   \space & DD & 1,178 & 284.32 & 715.66 & 2 \\
   \space & COLORS-3 & 10,500 & 61.31 & 91.03 & 11 \\
   \midrule
   \multirow{7}{*}{Node Level} 
          & Cora & - & 2,708 & 5,278 & 7 \\
   \space & Citeseer & - & 3,327 & 4,552 & 6 \\
   \space & Pubmed & - & 19,717 & 44,324 & 3 \\
   \space & CS & - & 18,333 & 163,788 & 15 \\
   \space & Physics & - & 34,493 & 495,924 & 5 \\
   \space & Photo & - & 7,484  & 126,530 & 8 \\
   \space & Computers & - & 13,381 & 259,159 & 10 \\
   \bottomrule
 \end{tabular}
 \label{tab:dataStats}
\end{table}

\subsection{Evaluation Metric}\label{sec:app-eva}
In this section, we will introdecue three metrics used to evaluate the graph backdoor attacks on FedGNN. 

ACC is the   average clean accuracy (ACC) across all clients,  the equation is defined as follows,
\begin{equation}
  \text{ACC} =  \frac{1}{K} \sum_{\mathit{c}_{j}\in \mathcal{C}} \frac{1}{N_{i}} \sum_{j\in\mathcal{D}^{i}} \mathbb{I}_{( \mathit{\hat{y}}_{j}^{i}=\mathbf{Y}^{i}_{j})},
\end{equation}
where  $\hat{y}_{j}^{i}$ is the predicted label outputted by $f_{i}(\mathbf{X}_{j}^{i},\mathbf{A}_{j}^{i};\mathbf{W}^{i})$.

The average attack success rate (ASR) on malicious clients, the ASR is defined as follows,
\begin{equation}
  \text{ASR} =   \frac{1}{M} \sum_{\mathit{c}_{j}\in \mathcal{C}_{p}} \frac{1}{|\mathcal{D}^{i}_{p}|} \sum_{j\in\mathcal{D}^{i}_{p}} \mathbb{I}_{( \mathit{\hat{y}}_{j}^{i}=\mathbf{Y}_{\tau})},
\end{equation}
where $\mathbf{Y}_{\tau}$ is the target label.

TASR  is the   metric gauges the vulnerability of 
normal clients to the graph backdoor attack, the TASR is defined as follows,
\begin{equation}
  \text{TASR} =   \frac{1}{K-M} \sum_{\mathit{c}_{j}\in \mathcal{C}_{m}} \frac{1}{|\mathcal{D}^{i}_{p}|} \sum_{j\in\mathcal{D}^{i}_{p}} \mathbb{I}_{( \mathit{\hat{y}}_{j}^{i}=\mathbf{Y}_{\tau})},
\end{equation}
where $ 1 \le K < M$. It is noted that the condition $1 \le K < M$ is necessary because TASR is designed to evaluate the transferred attack ability of graph backdoor attacks on normal clients. When $M = K$, meaning that all clients are malicious, there are no normal clients to evaluate the transferred attack ability. 

\subsection{Factors Setting}\label{sec:app-factors}
 we present the detailed factors setup considered in our study. It is important to note that the first value presented represents the default setting. To assess the individual impact of each factor, we keep the remaining factors fixed while systematically varying the corresponding values in our experiments. We first present the global factors, and then introduce the local factors. 

\subsubsection{Global Level}

\textit{Data Distribution}: The data distribution in our experiments consists of two categories, namely IID (independent and identically distributed) and Non-IID (non-independent and non-identically distributed). For the IID setting, both node-level and graph-level tasks involve random sampling of data within each client. However, the Non-IID setting differs for node-level and graph-level tasks. Specifically, for node-level Non-IID, we adopt the approach proposed in~\cite{wangfed2022, zhang2021subgraph}, which utilizes the Louvain community splitter to partition the graph (referred to as Louvain-Non-IID). For graph-level Non-IID, we employ two different approaches. The first approach, known as PD-Non-IID~\cite{FangCJG20}, involves assigning training instances with label $i$ to client $\mathit{c}_{i}$ with a probability $p$. The second approach allows different clients to possess significantly different amounts of data ( Num-Non-IID,).

\eat{For further details on the specific data distribution settings, }
\textit{Number of Malicious Attackers}: The number of malicious attackers 
$M$ is varied within the range of $[1,2,3,4,5]$.

\textit{Attack Time}: Assuming a total of $T$ training rounds, the start time $t^*$ of the backdoor attack is selected from the set $[0.0, 0.1, 0.2, 0.3, 0.4, 0.5]$.

\textit{Overlapping Rate}: The overlapping rate, represented as $\rho$, is experimented with different values from the set $T * [0.1, 0.2, 0.3, 0.4, 0.5]$.

\subsubsection{Local Level}
\textit{Trigger Size:} In the case of node-level datasets, we set the trigger size to $[3,4,5,6,7,8,9]$. For graph-level datasets, the trigger size is determined based on the relative average number of graph nodes. Let $N_{d}$ represent the average number of graph nodes, and the trigger size for graph-level datasets is set as $N_{d} *[ 0.1,0.2,0.3,0.4,0.5]$. (For graph-level tasks, it should be noted that the default trigger size is set to 0.3 for the DD and COLORS-3 datasets.)

\textit{Trigger Type:} The universal trigger type comprises erdos renyi graph (Renyi), watts strogatz graph (WS), barabasi albert graph (BA), and random regular graph (RR). On the other hand, the adaptive trigger type consists of GTA and UGBA, where UGBA is specifically designed for node-level tasks. For node-level tasks, we adopt trigger types including Renyi, WS, BA, RR, and GTA. For graph-level tasks, the trigger types include Renyi, WS, BA, GTA, and UGBA. 

\textit{Trigger Position:} We consider two types of trigger positions: random position (Random) and position based on important indicators such as degree and cluster-based scores (Degree and Cluster). 

\textit{Poisoning Rate.} The $\rho$ takes values from the set $[0.1, 0.2, 0.3, 0.4, 0.5]$.

\subsection{Implementation Details}\label{sec:implement}
Our implementation of the backdoor attack on FedGNN is based on the PyTorch framework. The experiments were carried out on two server configurations: three Linux Centos Servers, each with 4 RTX 3090 GPUs, and two Linux Ubuntu Servers, each with 2 V100 GPUs. In both node-level and graph-level tasks, we adopt the inductive learning settings as outlined in~\cite{xu2022more, daiunnoti2023}. 
For each dataset, we ensure consistent experimental conditions by employing the same training and attack settings. We set the total number of clients to $5$, and all clients participate in the training process at each round. Each experiment is repeated five times. The default training rounds for node-level tasks and graph-level tasks are set at 200 and 1000, respectively. However, the Computers and Photo datasets require 2000 training rounds, while the CS and Computer datasets necessitate 1000 rounds.

\subsection{Experiment Results}

\subsubsection{Benchmark Results of Graph Backdoor Attack on FedGNN on Other Datasets}\label{sec:app-benchmark}
\begin{figure}[!]
    \centering
    \includegraphics[width=1.0\linewidth]{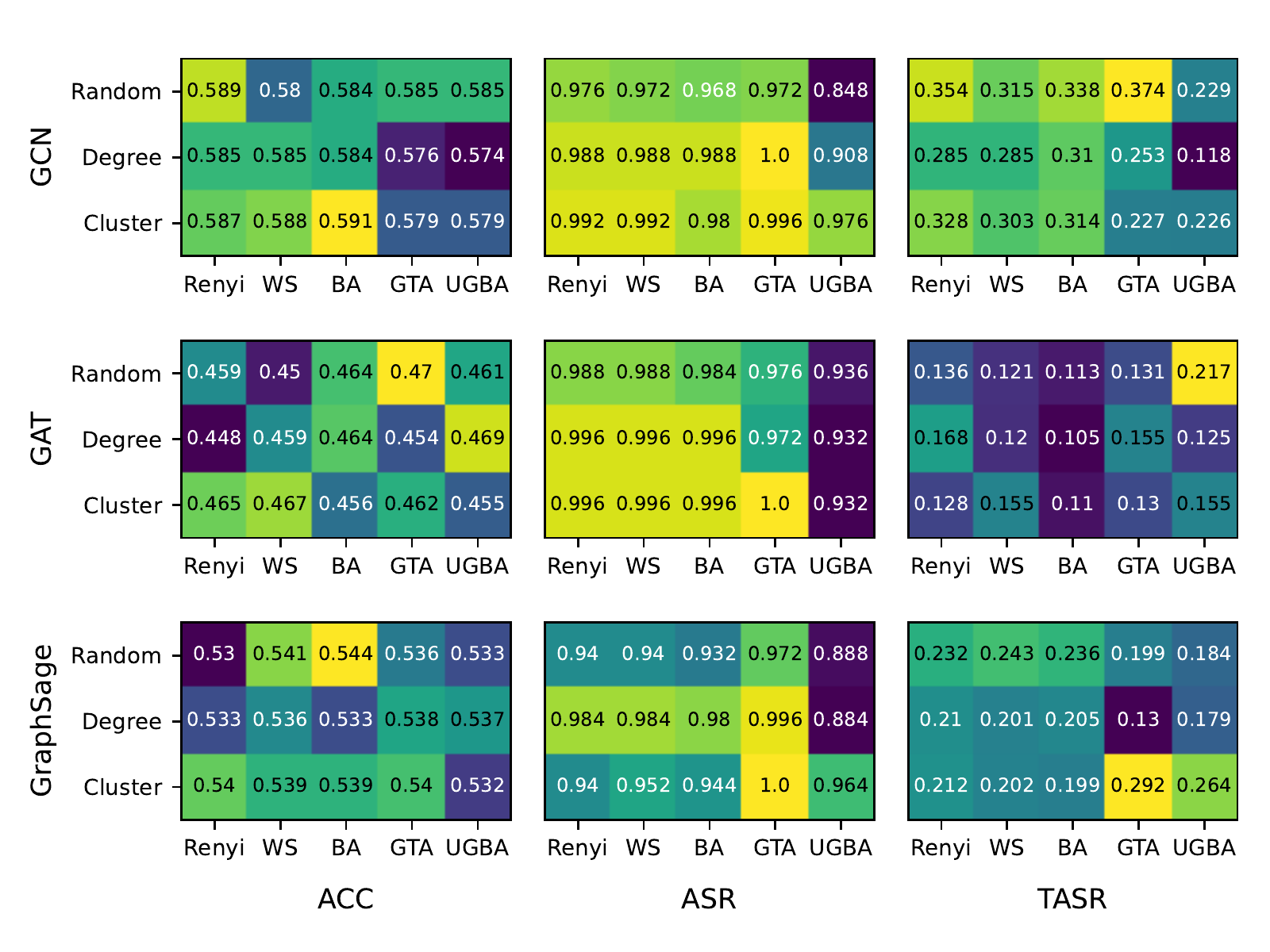}
    \setlength{\belowcaptionskip}{-0.5cm}
  \caption{Graph backdoor attack on Cora. }\label{figs:Appendix-Main-Cora}
\end{figure}

\begin{figure}[!]
    \centering
    \includegraphics[width=1.0\linewidth]{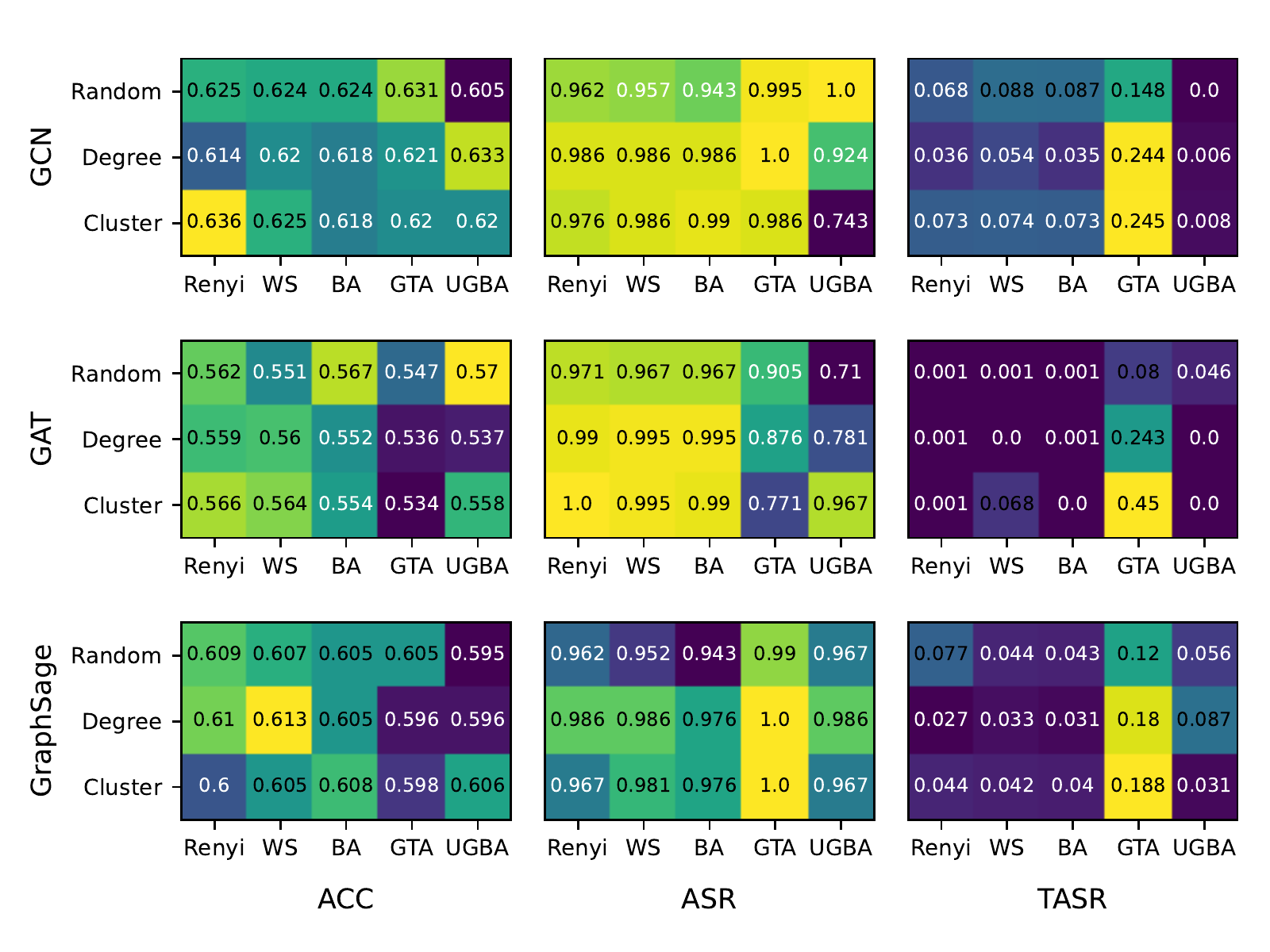}
    \setlength{\belowcaptionskip}{-0.5cm}
  \caption{Graph backdoor attack on CiteSeer. }\label{figs:Appendix-Main-CiteSeer}
\end{figure}

\begin{figure}[!]
    \centering
    \includegraphics[width=1.0\linewidth]{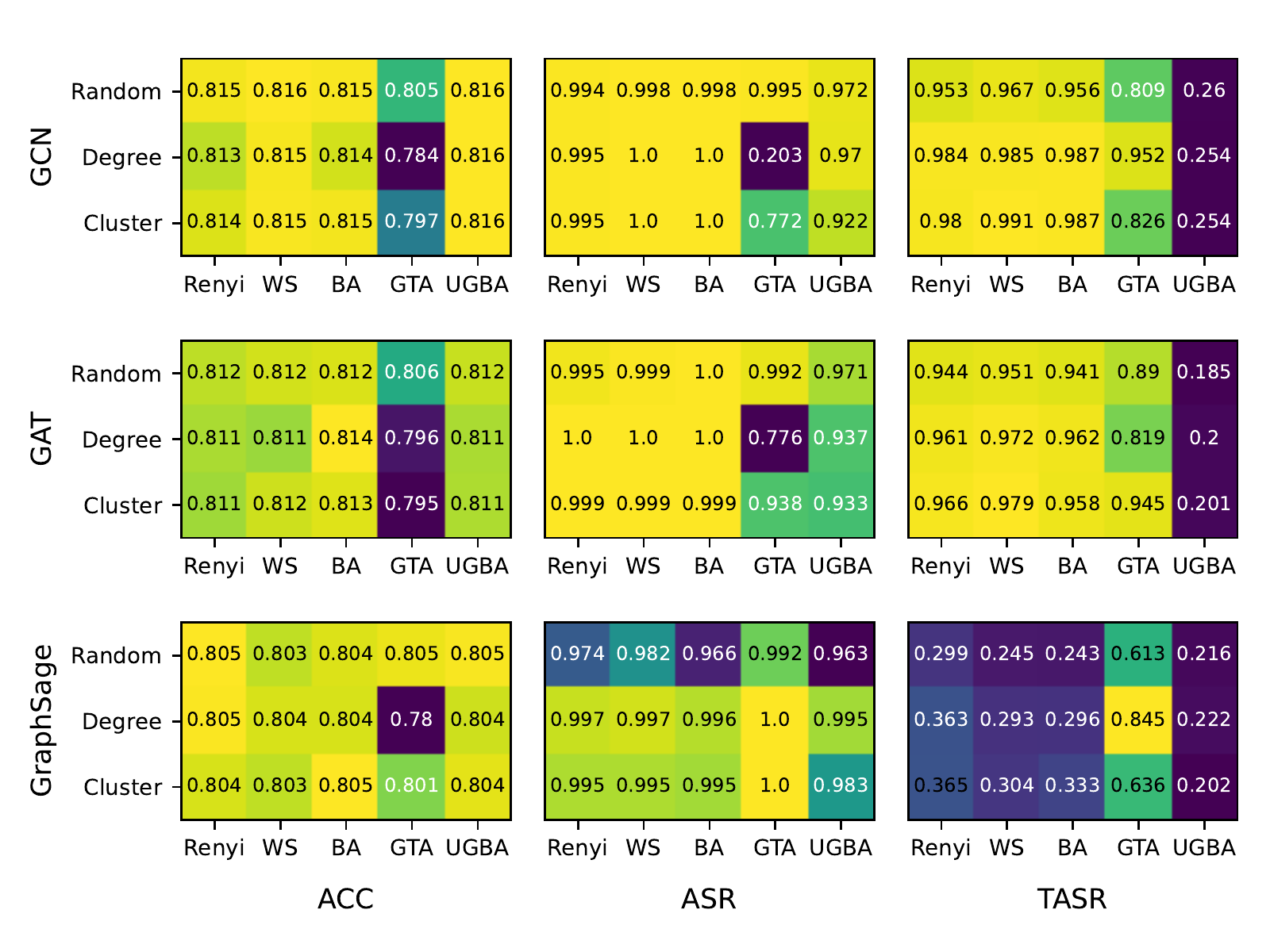}
    \setlength{\belowcaptionskip}{-0.5cm}
  \caption{Graph backdoor attack on Pubmed. }\label{figs:Appendix-Main-Pubmed}
\end{figure}

\begin{figure}[!]
    \centering
    \includegraphics[width=1.0\linewidth]{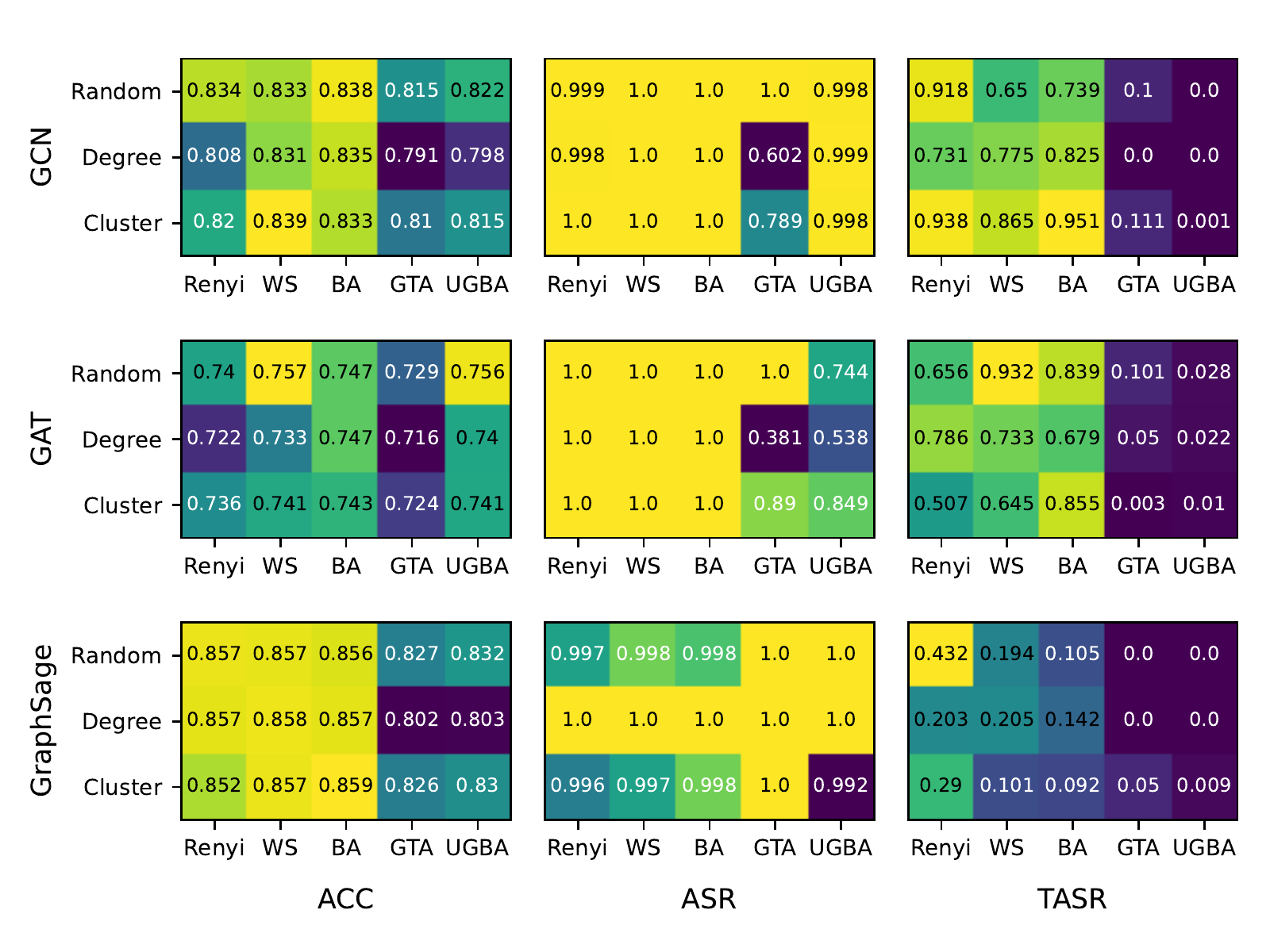}
    \setlength{\belowcaptionskip}{-0.5cm}
  \caption{Graph backdoor attack on CS. }\label{figs:Appendix-Main-CS}
\end{figure}

\begin{figure}[!]
    \centering
    \includegraphics[width=1.0\linewidth]{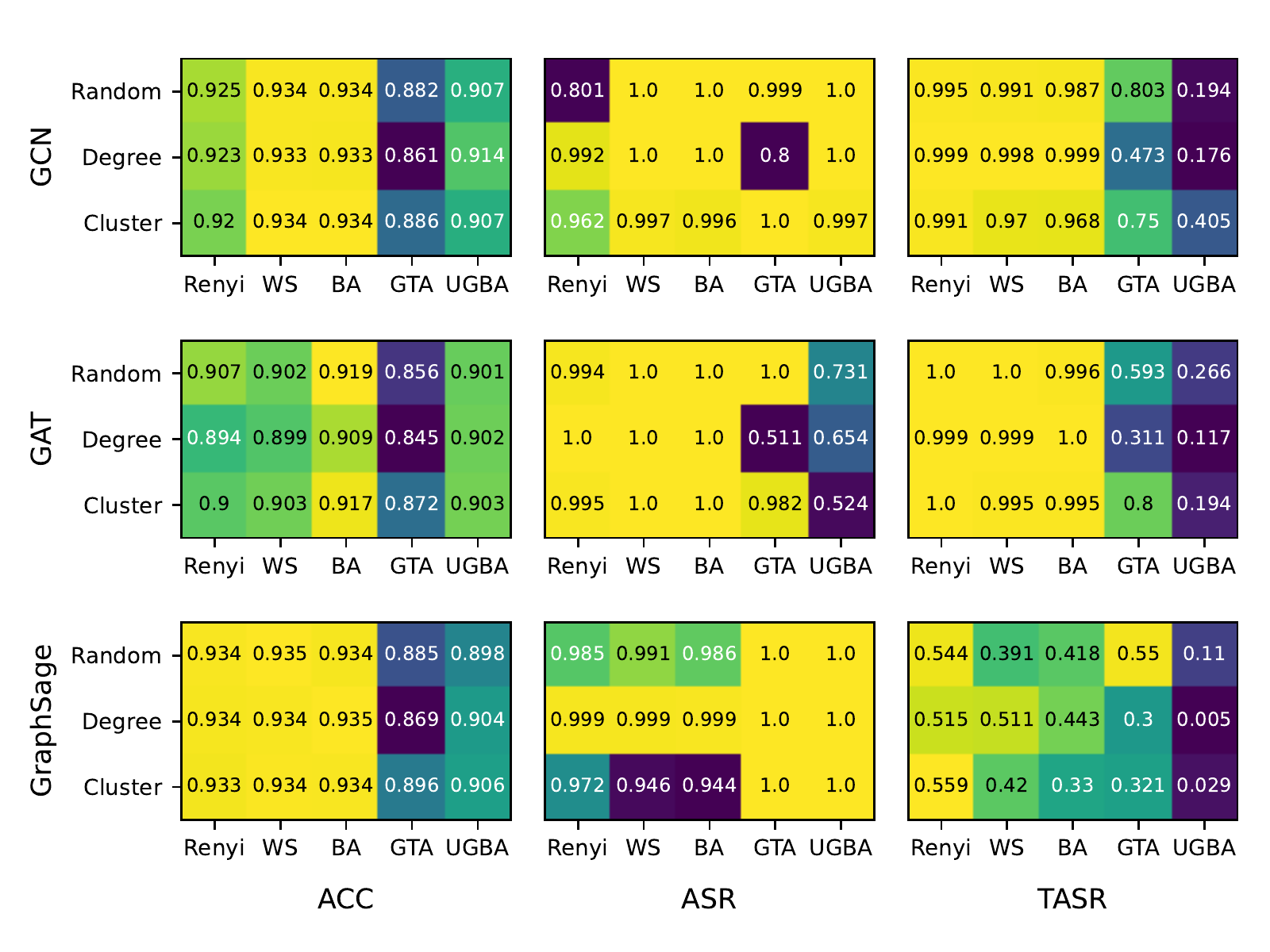}
    \setlength{\belowcaptionskip}{-0.5cm}
  \caption{Graph backdoor attack on Physics. }\label{figs:Appendix-Main-Physics}
\end{figure}

\begin{figure}[!]
    \centering
    \includegraphics[width=1.0\linewidth]{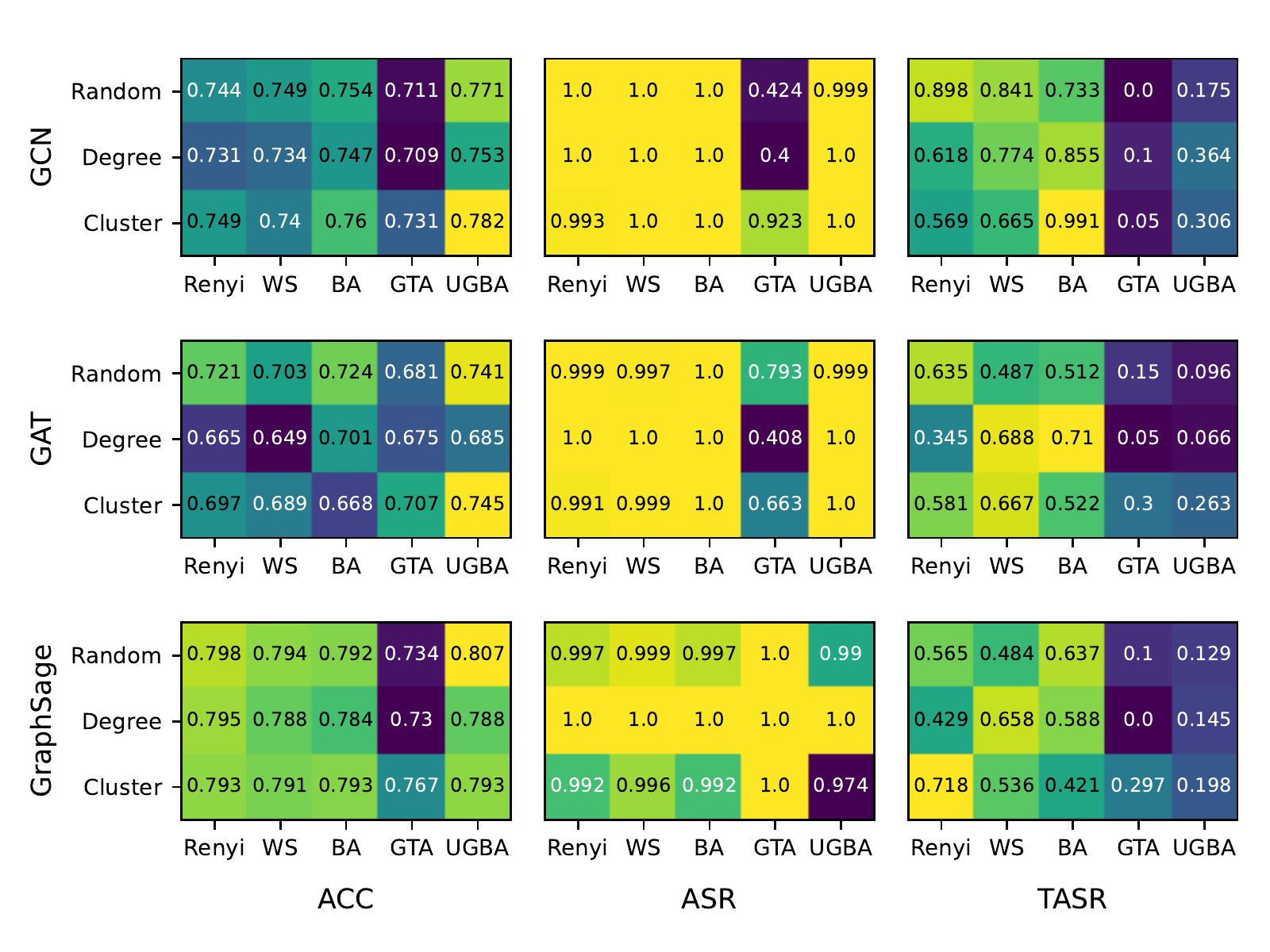}
    \setlength{\belowcaptionskip}{-0.5cm}
  \caption{Graph backdoor attack on Photo. }\label{figs:Appendix-Main-Photo}
\end{figure}

\begin{figure}[!]
    \centering
    \includegraphics[width=1.0\linewidth]{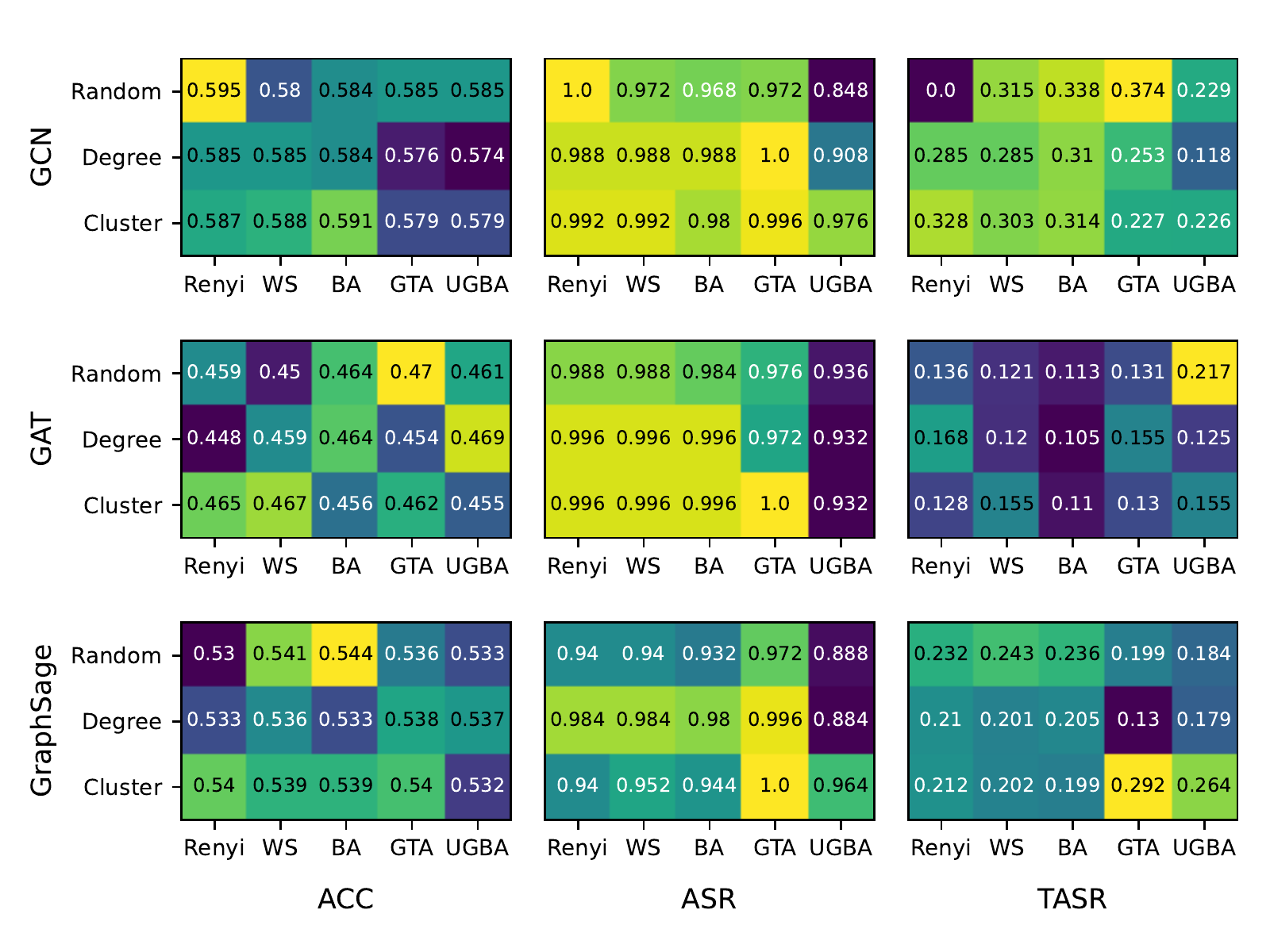}
    \setlength{\belowcaptionskip}{-0.5cm}
  \caption{Graph backdoor attack on Computers. }\label{figs:Appendix-Main-Computers}
\end{figure}

\begin{figure}[!]
    \centering
    \includegraphics[width=1.0\linewidth]{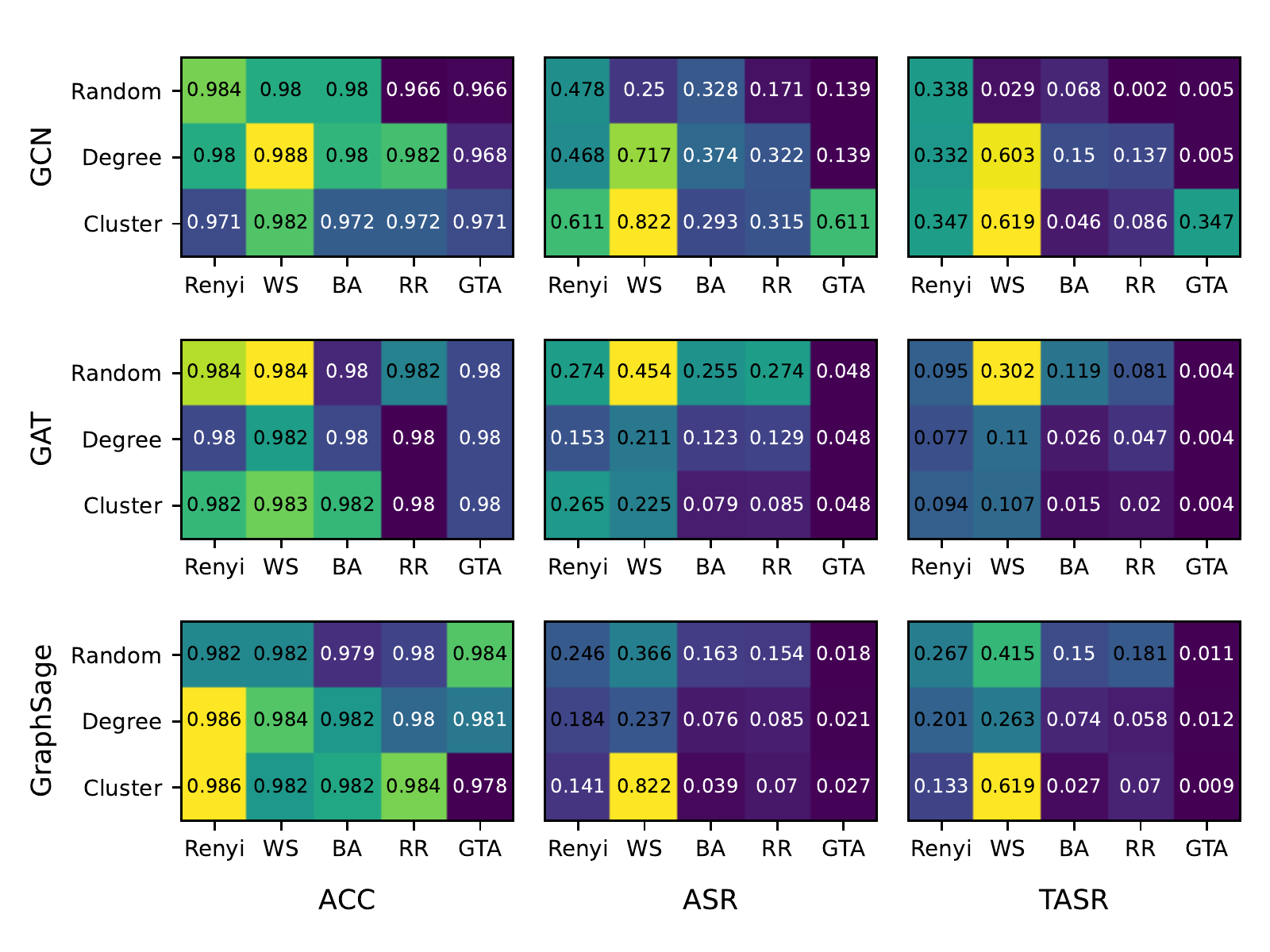}
    \setlength{\belowcaptionskip}{-0.5cm}
  \caption{Graph backdoor attack on AIDS. }\label{figs:Appendix-Main-AIDS}
\end{figure}

\begin{figure}[!]
    \centering
    \includegraphics[width=1.0\linewidth]{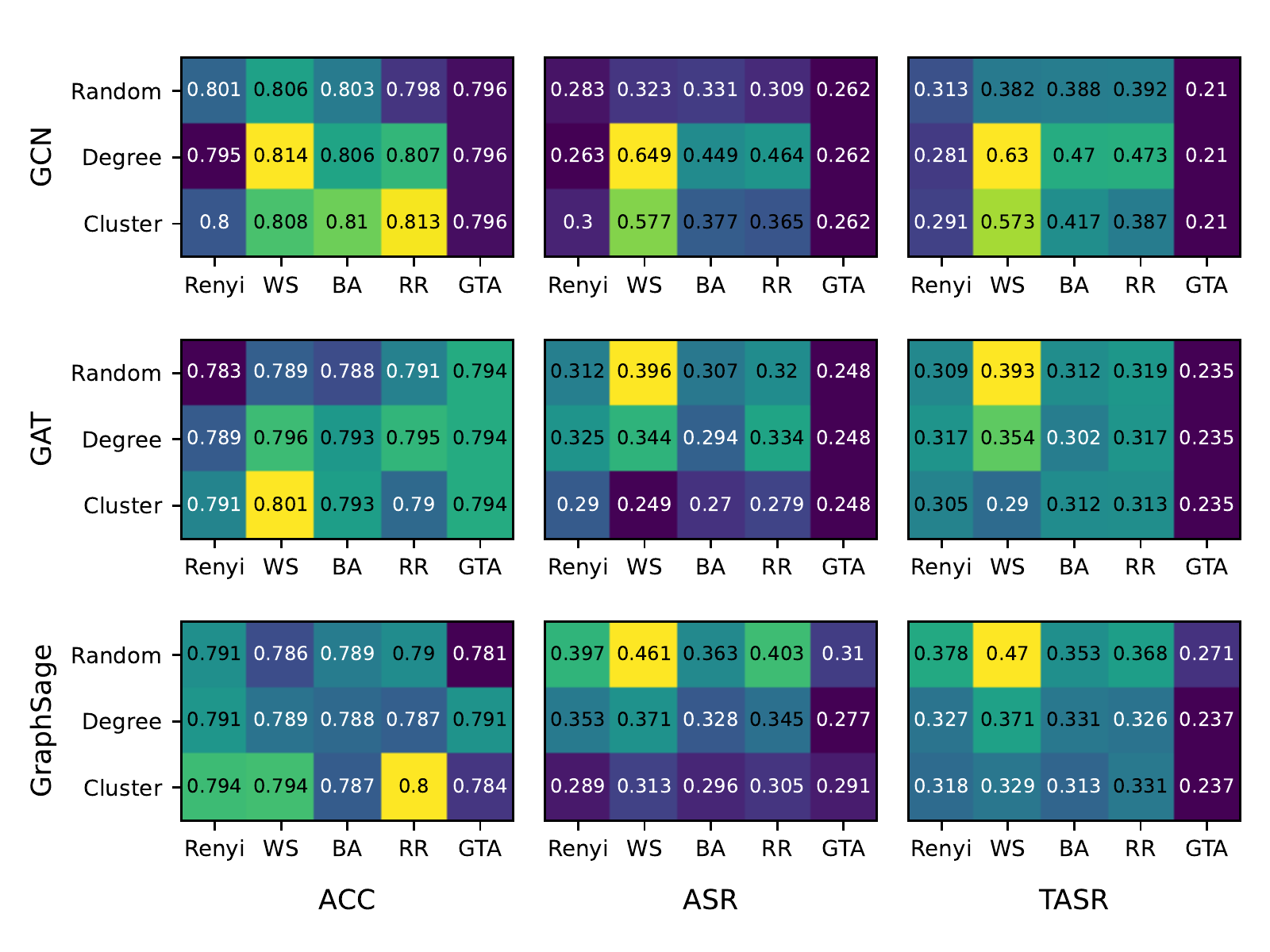}
    \setlength{\belowcaptionskip}{-0.5cm}
  \caption{Graph backdoor attack on NCI1. }\label{figs:Appendix-Main-NCI1}
\end{figure}

\begin{figure}[!]
    \centering
    \includegraphics[width=1.0\linewidth]{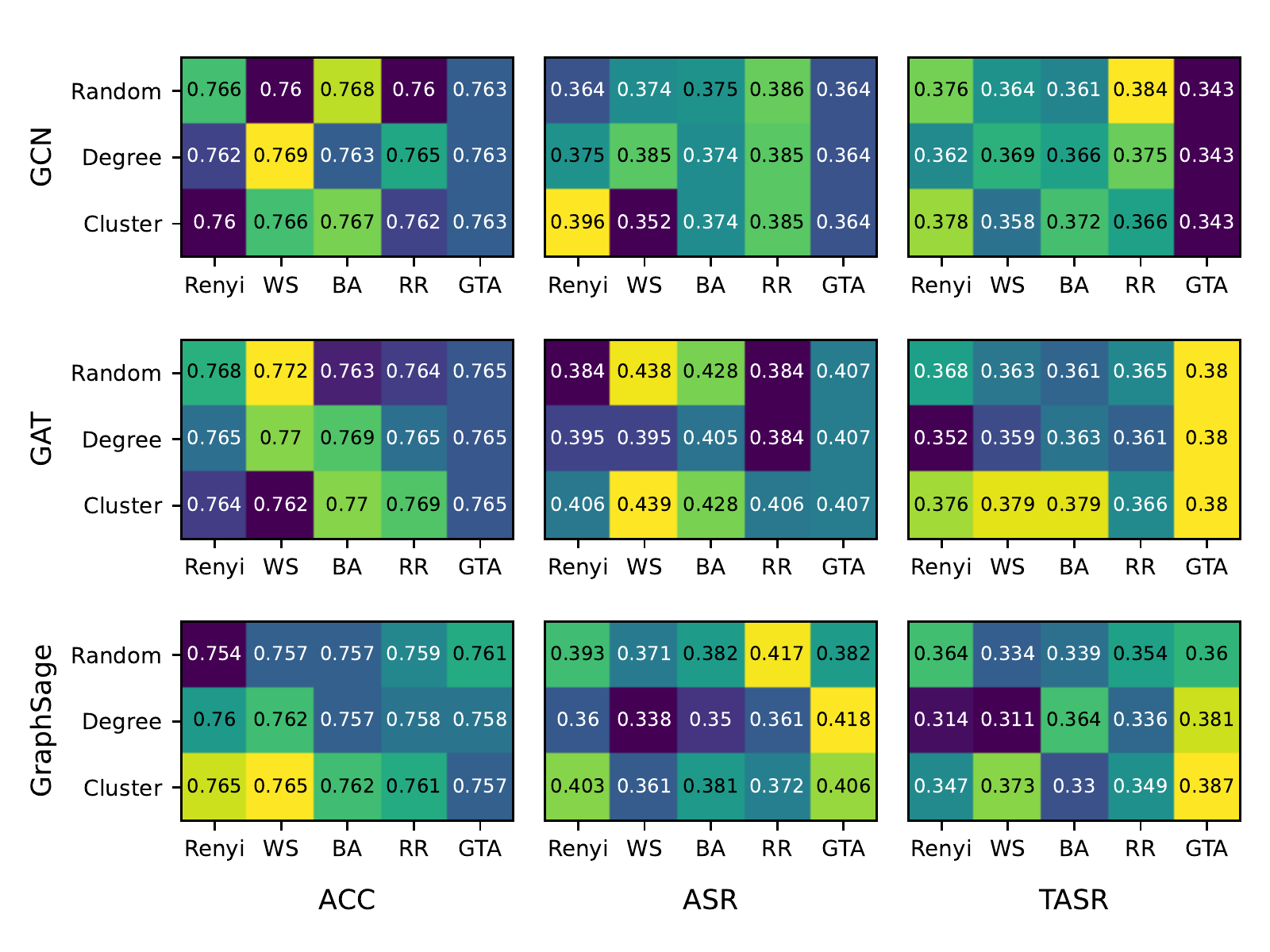}
    \setlength{\belowcaptionskip}{-0.5cm}
  \caption{Graph backdoor attack on PROTEINS-full. }\label{figs:Appendix-Main-PROTEINS_full}
\end{figure}

\begin{figure}[!]
    \centering
    \includegraphics[width=1.0\linewidth]{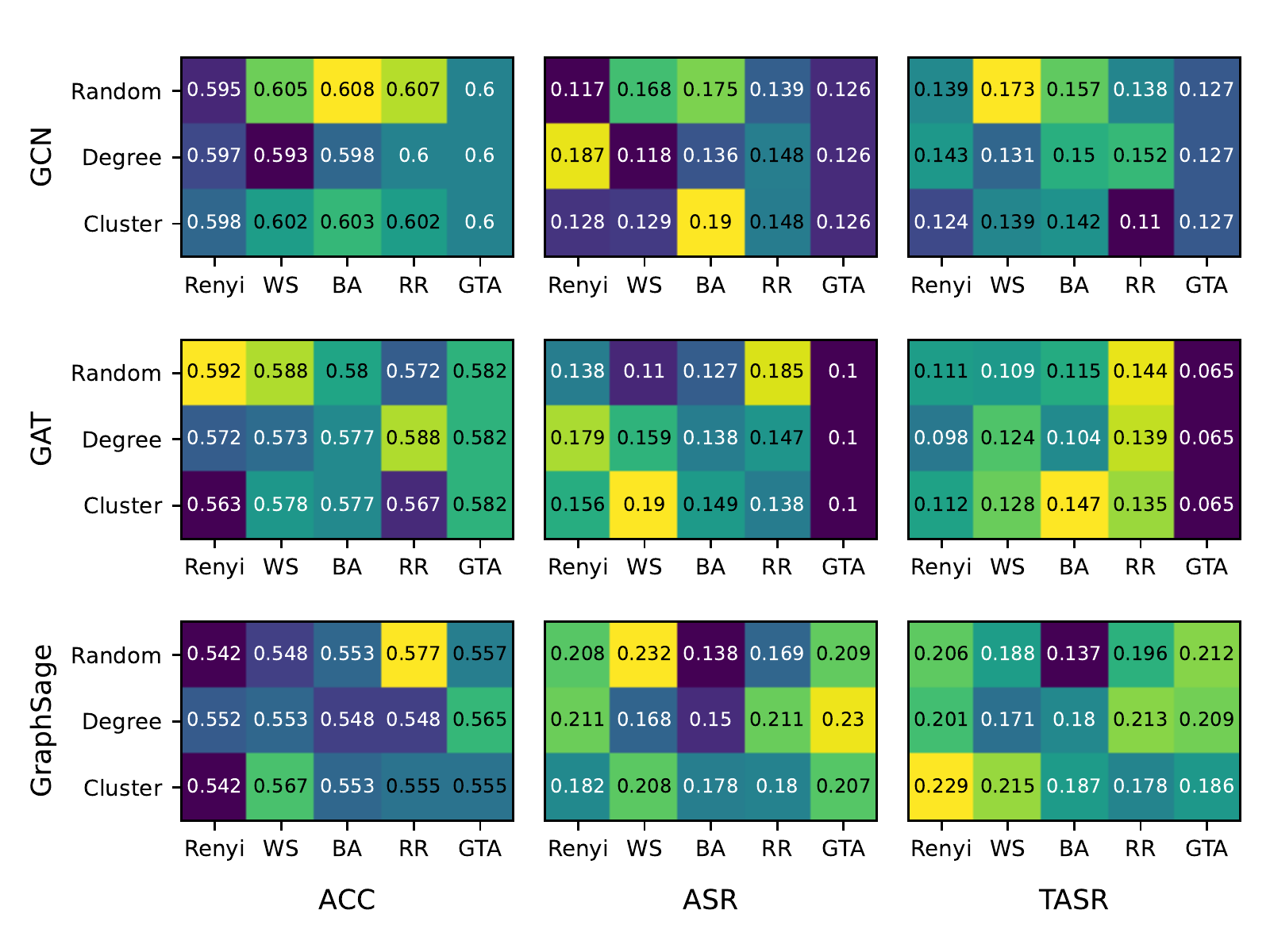}
    \setlength{\belowcaptionskip}{-0.5cm}
  \caption{Graph backdoor attack on ENZYMES. }\label{figs:Appendix-Main-ENZYMES}
\end{figure}

\begin{figure}[!]
    \centering
    \includegraphics[width=1.0\linewidth]{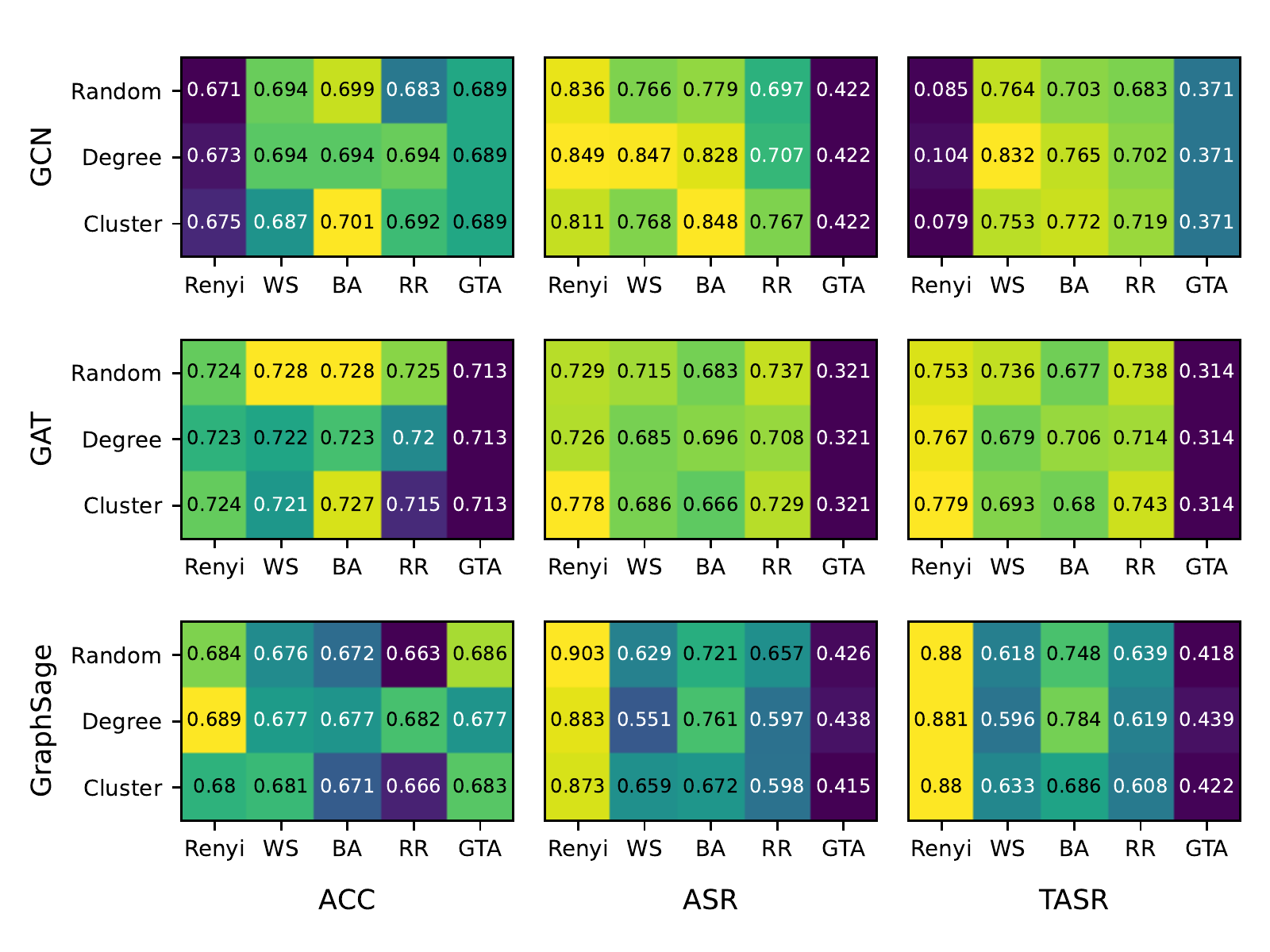}
    \setlength{\belowcaptionskip}{-0.5cm}
  \caption{Graph backdoor attack on DD. }\label{figs:Appendix-Main-DD}
\end{figure}

\begin{figure}[!]
    \centering
    \includegraphics[width=1.0\linewidth]{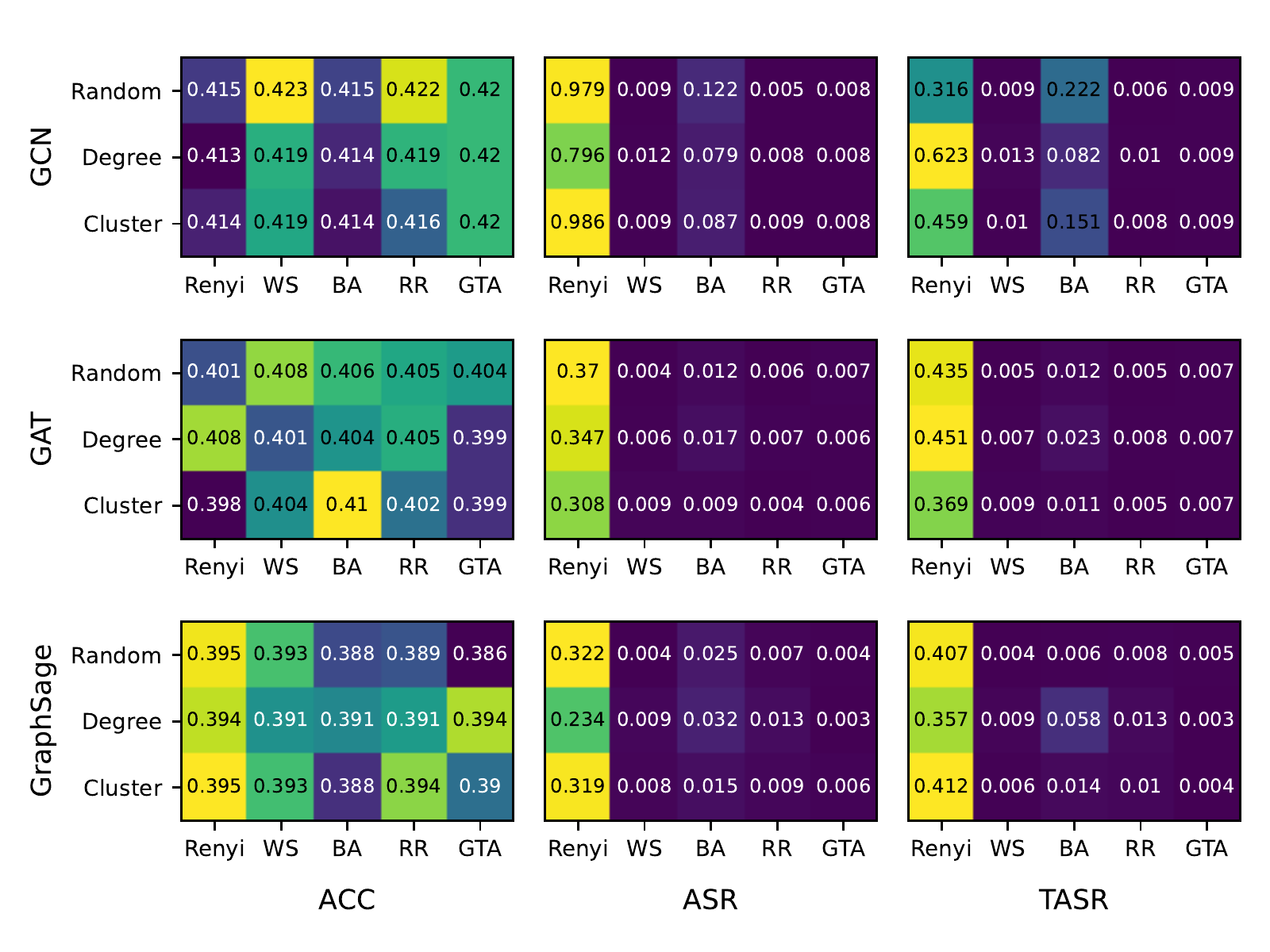}
    \setlength{\belowcaptionskip}{-0.5cm}
  \caption{Graph backdoor attack on COLORS-3. }\label{figs:Appendix-Main-COLORS-3}
\end{figure}

The primary experiments of graph backdoor attacks on FedGNN are presented in Figures~\ref{figs:Appendix-Main-Cora}-\ref{figs:Appendix-Main-COLORS-3}, revealing several key findings. First, the adaptive graph backdoor attack on node-level classification demonstrates a higher ASR and TASR across the majority of datasets, while the universal graph backdoor attack graph-level datasets exhibits increased ASR and TASR. Second, the combination of trigger type and trigger position is particularly significant in graph-level tasks, as it substantially influences the attack performance. In contrast, the interaction between trigger type and position has a minimal impact on node classification tasks, given the consistently high attack success rates. Third, in node-level tasks, the GCN exhibits a higher TASR compared to GAT and GraphSage. Conversely, in graph-level tasks, GraphSage and GAT demonstrate a significantly increased TASR. The details results of each datasets are presented as follows.

% \textbf{Cora.} The  experimental results of graph backdoor attacks on Cora are shown in Figure~\ref{figs:Appendix-Main-Cora}.

% \textbf{CiteSeer.} The  experimental results of graph backdoor attacks on CiteSeer are shown in Figure~\ref{figs:Appendix-Main-CiteSeer}.

% \textbf{Pubmed.} The  experimental results of graph backdoor attacks on Pubmed are shown in Figure~\ref{figs:Appendix-Main-Pubmed}.

% \textbf{CS.} The  experimental results of graph backdoor attacks on CS are shown in Figure~\ref{figs:Appendix-Main-CS}.

% \textbf{Physics.} The  experimental results of graph backdoor attacks on Physics are shown in Figure~\ref{figs:Appendix-Main-Physics}.

% \textbf{Photo.} The  experimental results of graph backdoor attacks on Photo are shown in Figure~\ref{figs:Appendix-Main-Photo}.

% \textbf{Computers.} The  experimental results of graph backdoor attacks on Computers are shown in Figure~\ref{figs:Appendix-Main-Computers}.

%%%%%%%%%%%%%%%%%%%%%%%%%%%%%%%%%%%%%%%%%%%%%%%%%%%
% \textbf{AIDS.} The  experimental results of graph backdoor attacks on AIDS are shown in Figure~\ref{figs:Appendix-Main-AIDS}.

% \textbf{NCI1.} The  experimental results of graph backdoor attacks on NCI1 are shown in Figure~\ref{figs:Appendix-Main-NCI1}.

% \textbf{PROTEINS-full.} The  experimental results of graph backdoor attacks on PROTEINS-full are shown in Figure~\ref{figs:Appendix-Main-PROTEINS_full}.

% \textbf{ENZYMES.} The  experimental results of graph backdoor attacks on ENZYMES are shown in Figure~\ref{figs:Appendix-Main-ENZYMES}.

% \textbf{DD.} The  experimental results of graph backdoor attacks on DD are shown in Figure~\ref{figs:Appendix-Main-DD}.

% \textbf{COLORS-3.} The  experimental results of graph backdoor attacks on COLORS-3 are shown in Figure~\ref{figs:Appendix-Main-COLORS-3}.

\subsubsection{Impact of Critical Factors on Other Datasets}~\label{sec:app-impact-factors}
\vspace{-20pt}

\begin{figure}[!]
\centering
\subfigure[ACC]{
\begin{minipage}[t]{1.0\linewidth}
\centering
\includegraphics[width=5.0in]{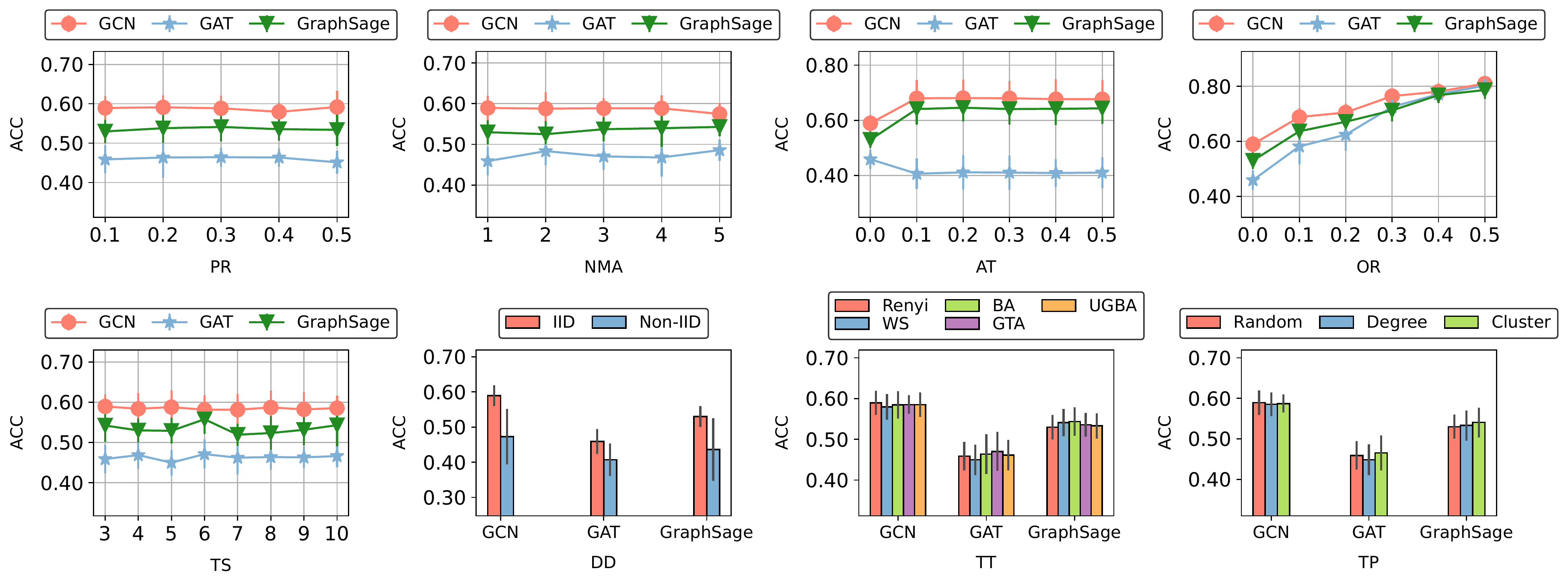}
%\caption{fig1}
\end{minipage}%
}%
\\
\subfigure[ASR]{
\begin{minipage}[t]{1.0\linewidth}
\centering
\includegraphics[width=5.0in]{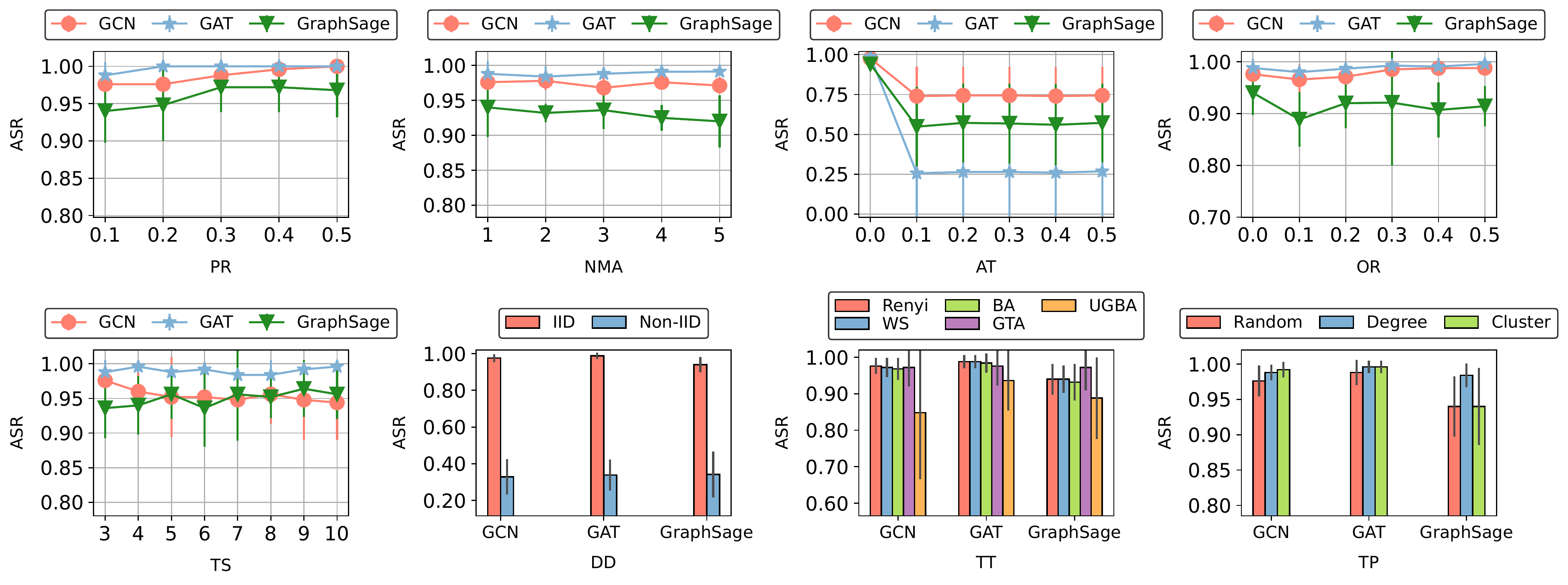}
%\caption{fig2}
\end{minipage}%
}%
\\
\subfigure[TASR]{
\begin{minipage}[t]{1.0\linewidth}
\centering
\includegraphics[width=5.0in]{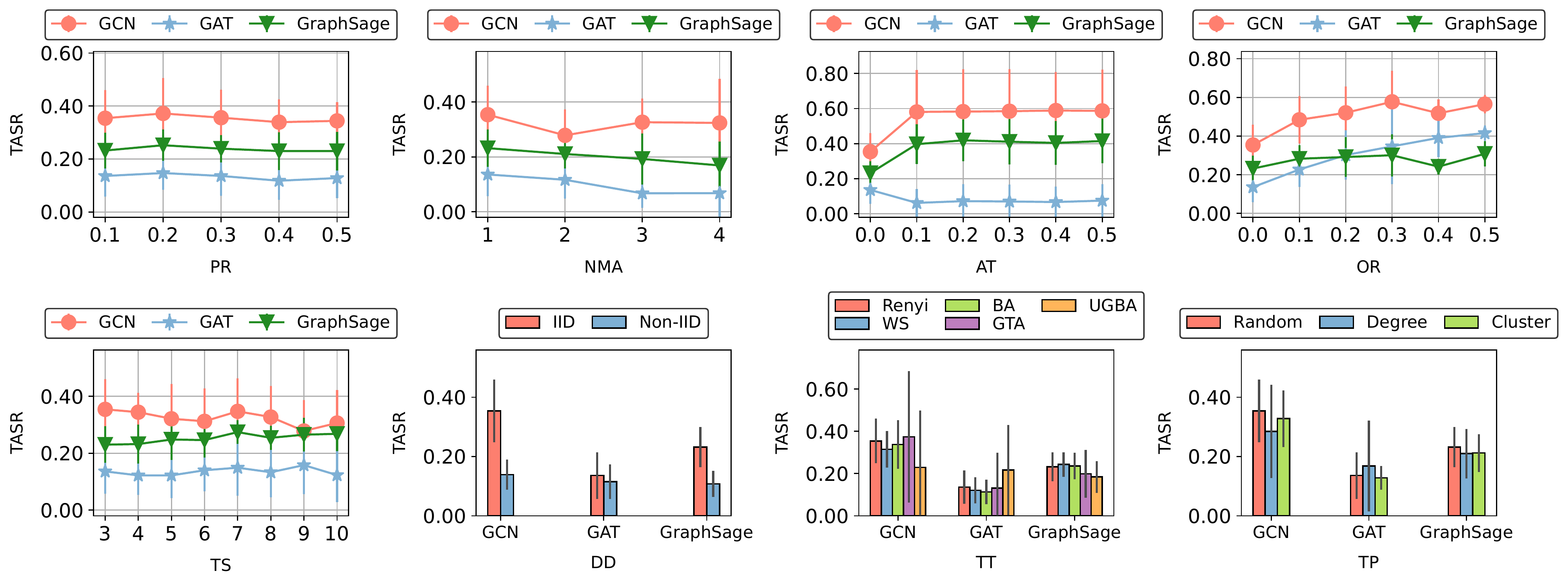}
%\caption{fig2}
\end{minipage}
}%
\centering
\caption{Graph backdoor attacks  on Cora.}~\label{figs:Appendix-factors-Cora}
\end{figure}

\begin{figure}[!]
\centering
\subfigure[ACC]{
\begin{minipage}[t]{1.0\linewidth}
\centering
\includegraphics[width=5.0in]{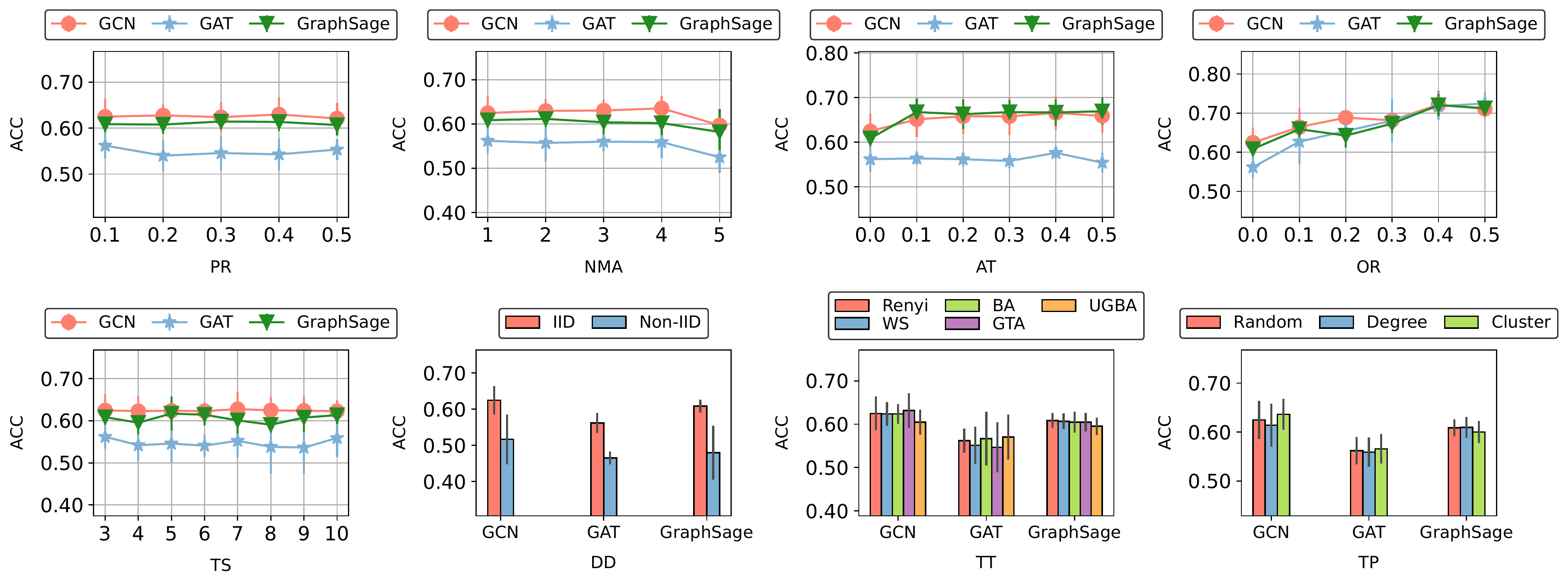}
%\caption{fig1}
\end{minipage}%
}%
\\
\subfigure[ASR]{
\begin{minipage}[t]{1.0\linewidth}
\centering
\includegraphics[width=5.0in]{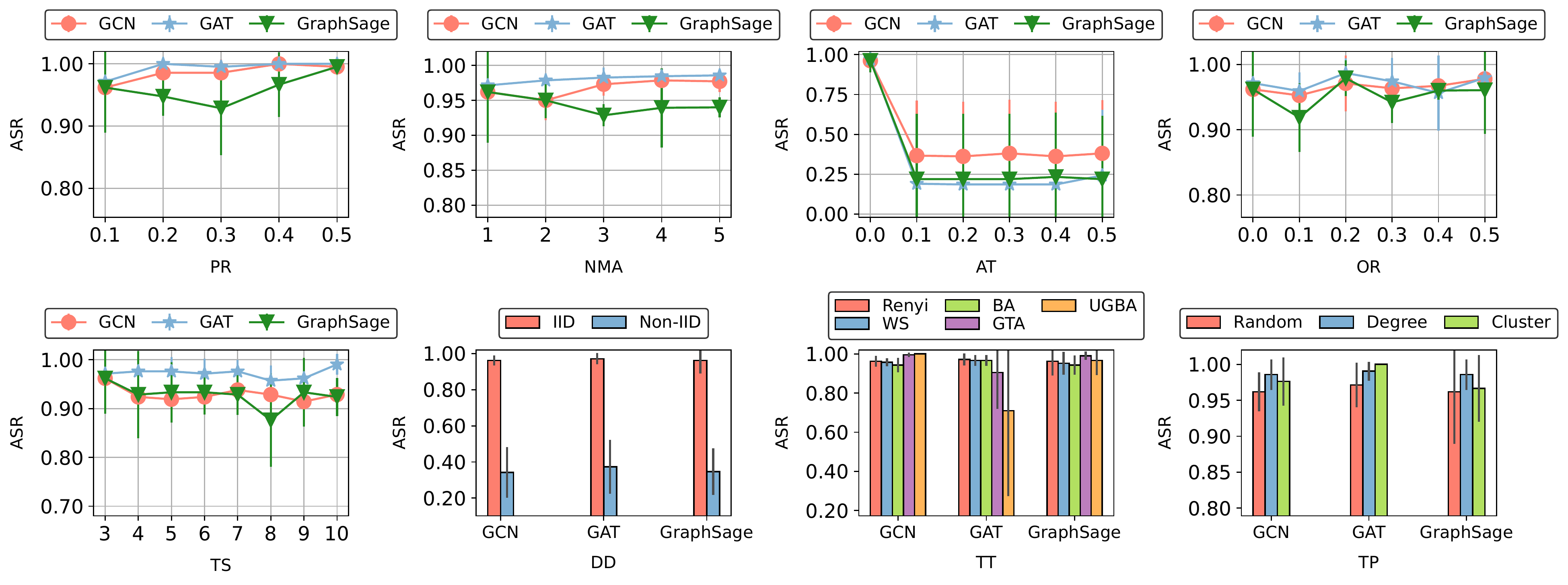}
%\caption{fig2}
\end{minipage}%
}%
\\
\subfigure[TASR]{
\begin{minipage}[t]{1.0\linewidth}
\centering
\includegraphics[width=5.0in]{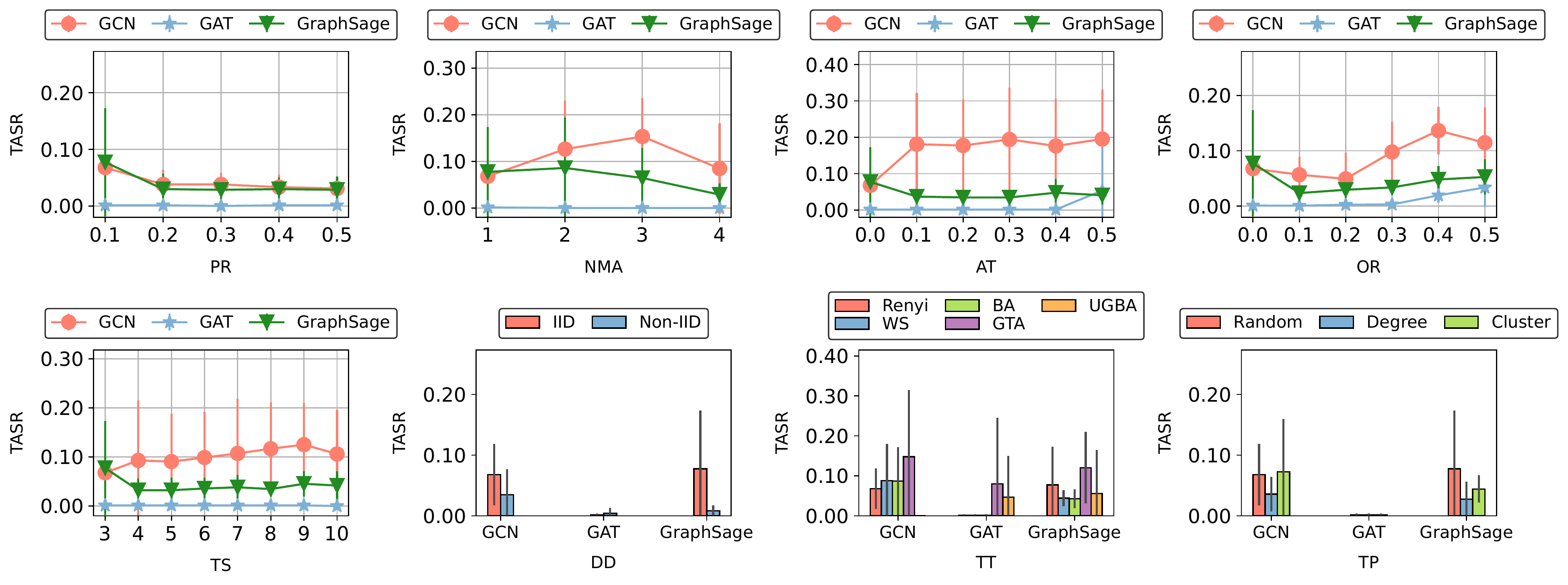}
%\caption{fig2}
\end{minipage}
}%
\centering
\caption{Graph backdoor attack on CiteSeer.}~\label{figs:Appendix-factors-Citeseer}
\end{figure}

\begin{figure}[!]
\centering
\subfigure[ACC]{
\begin{minipage}[t]{1.0\linewidth}
\centering
\includegraphics[width=5.0in]{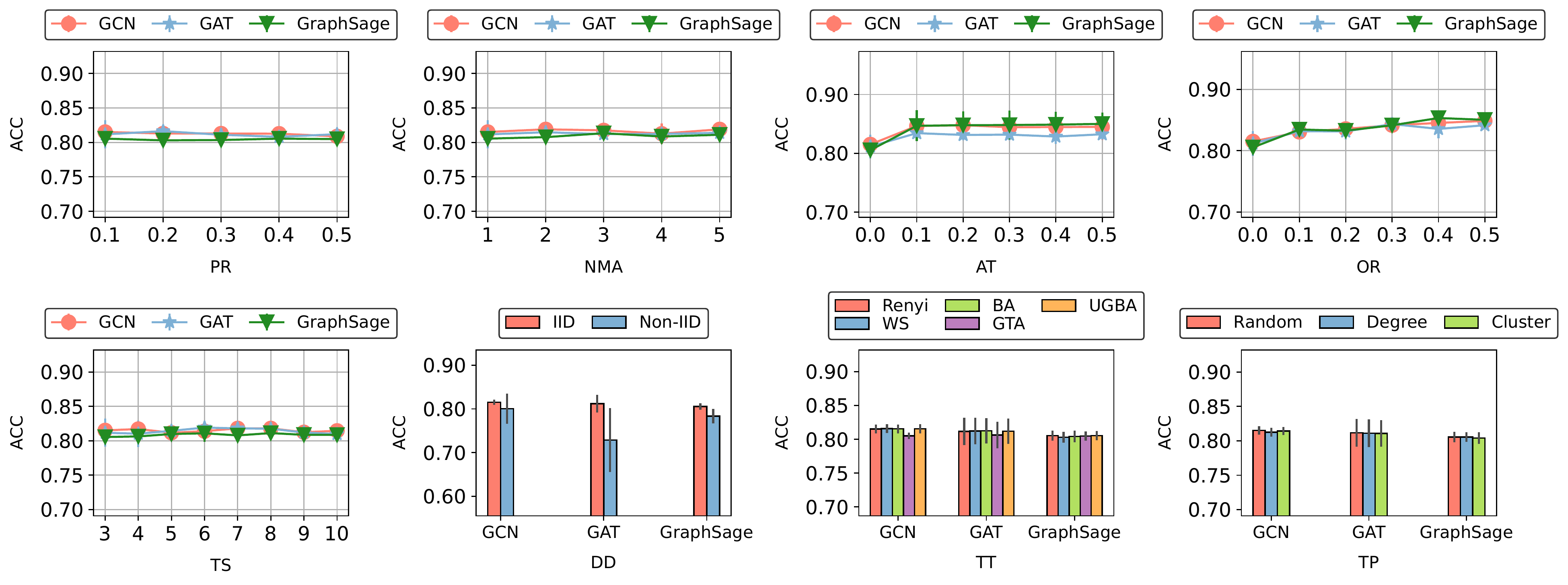}
%\caption{fig1}
\end{minipage}%
}%
\\
\subfigure[ASR]{
\begin{minipage}[t]{1.0\linewidth}
\centering
\includegraphics[width=5.0in]{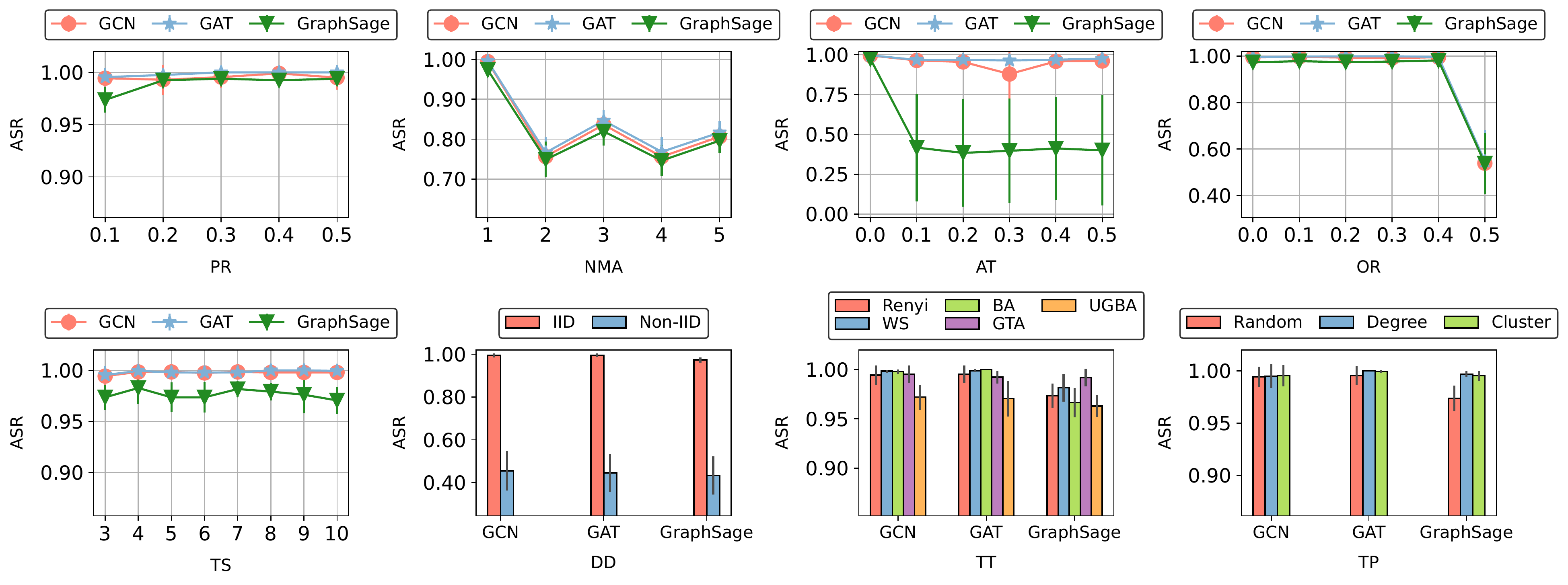}
%\caption{fig2}
\end{minipage}%
}%
\\
\subfigure[TASR]{
\begin{minipage}[t]{1.0\linewidth}
\centering
\includegraphics[width=5.0in]{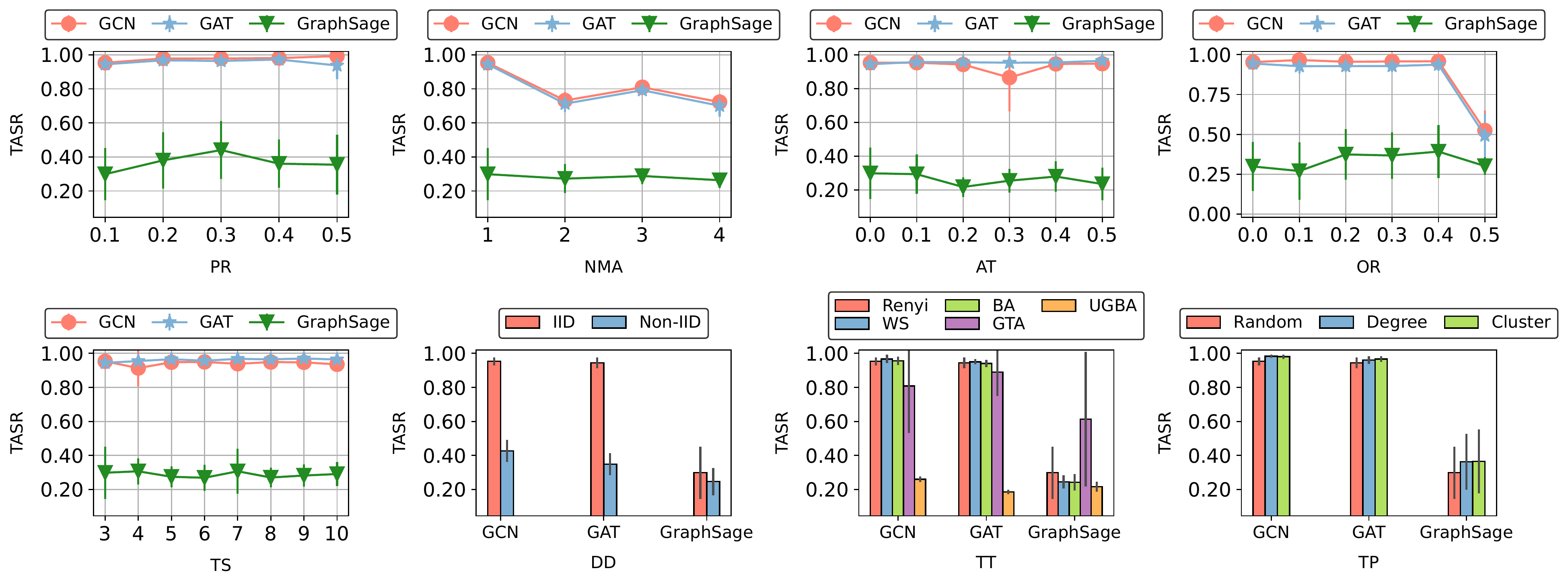}
%\caption{fig2}
\end{minipage}
}%
\centering
\caption{Graph backdoor attack on Pubmed.}~\label{figs:Appendix-factors-Pubmed}
\end{figure}

\begin{figure}[!]
\centering
\subfigure[ACC]{
\begin{minipage}[t]{1.0\linewidth}
\centering
\includegraphics[width=5.0in]{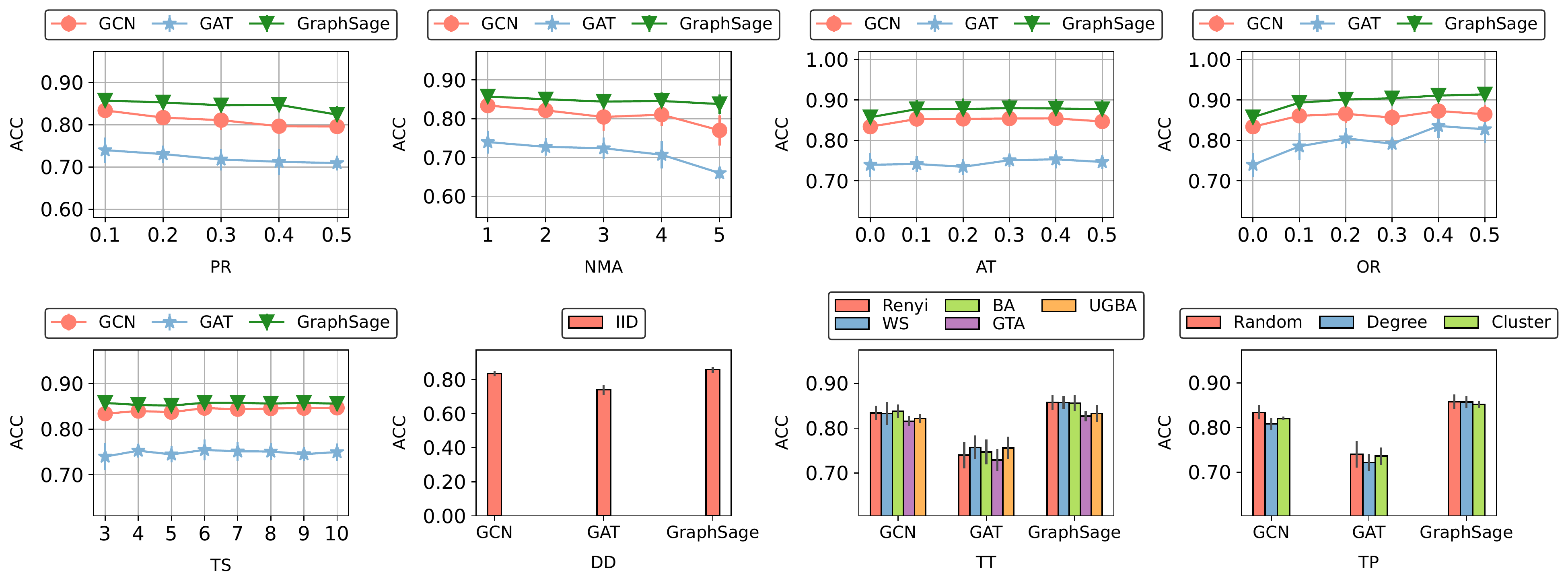}
%\caption{fig1}
\end{minipage}%
}%
\\
\subfigure[ASR]{
\begin{minipage}[t]{1.0\linewidth}
\centering
\includegraphics[width=5.0in]{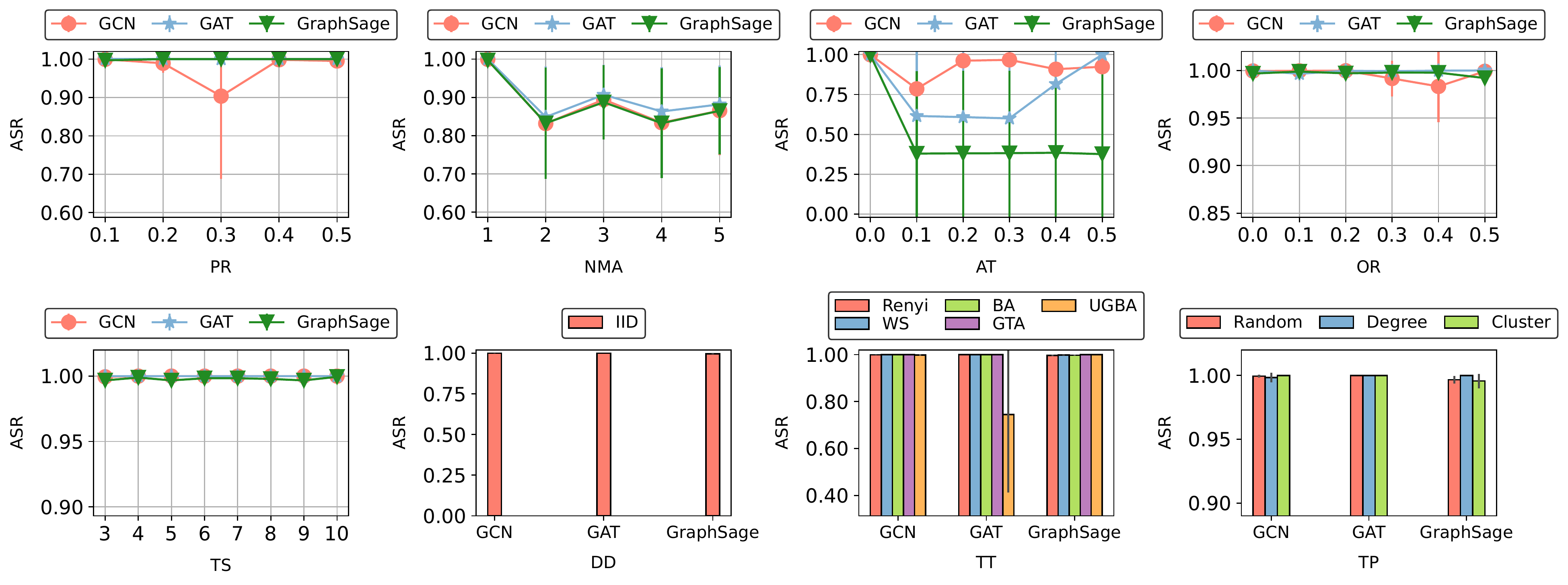}
%\caption{fig2}
\end{minipage}%
}%
\\
\subfigure[TASR]{
\begin{minipage}[t]{1.0\linewidth}
\centering
\includegraphics[width=5.0in]{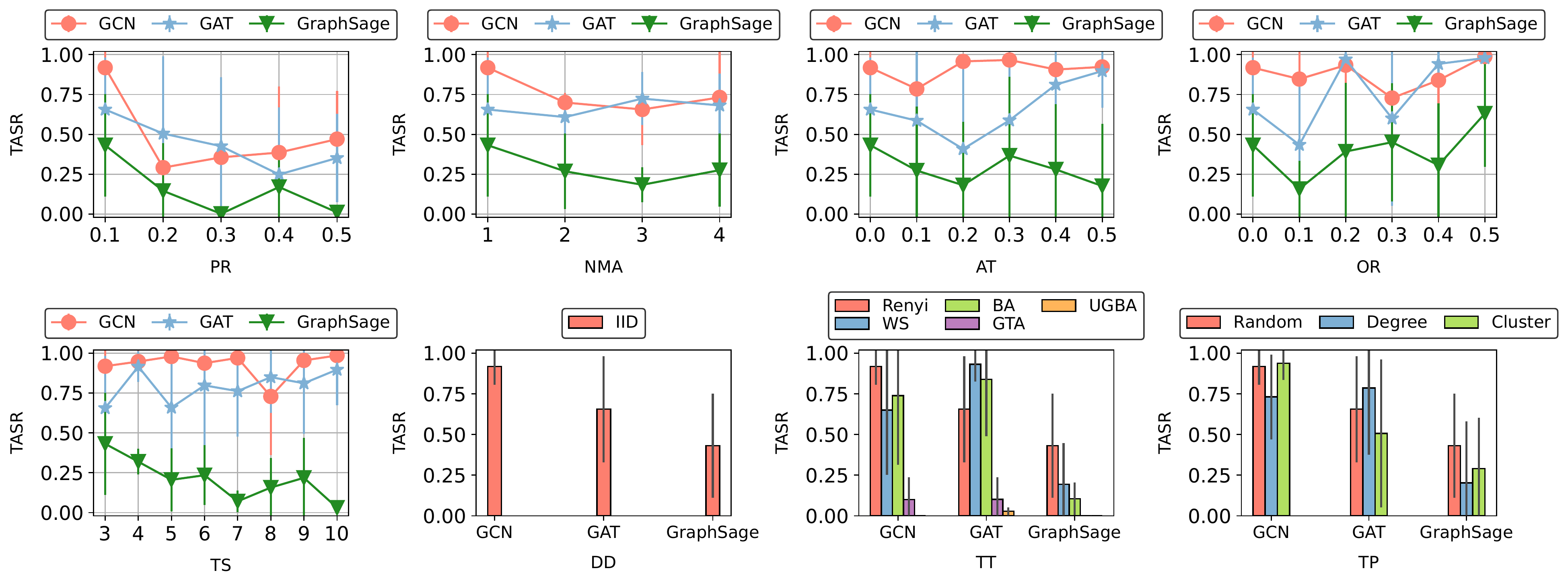}
%\caption{fig2}
\end{minipage}
}%
\centering
\caption{Graph backdoor attack on CS.}~\label{figs:Appendix-factors-CS}
\end{figure}

\begin{figure}[!]
\centering
\subfigure[ACC]{
\begin{minipage}[t]{1.0\linewidth}
\centering
\includegraphics[width=5.0in]{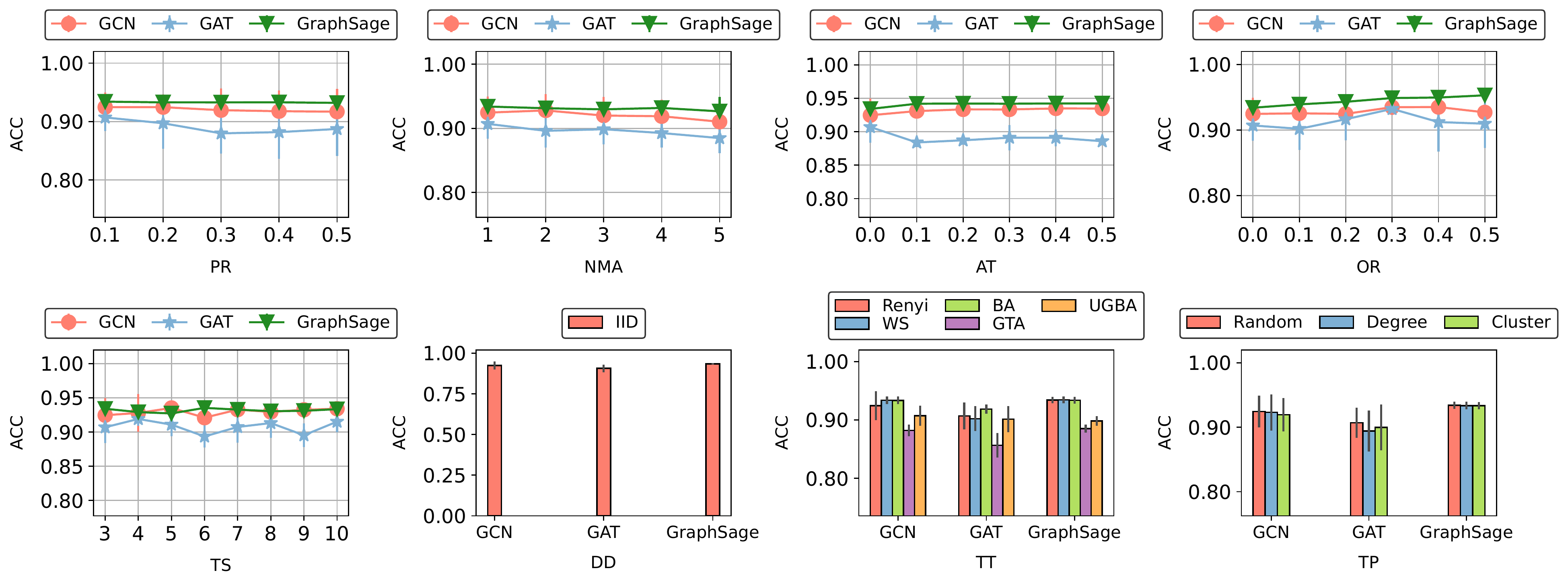}
%\caption{fig1}
\end{minipage}%
}%
\\
\subfigure[ASR]{
\begin{minipage}[t]{1.0\linewidth}
\centering
\includegraphics[width=5.0in]{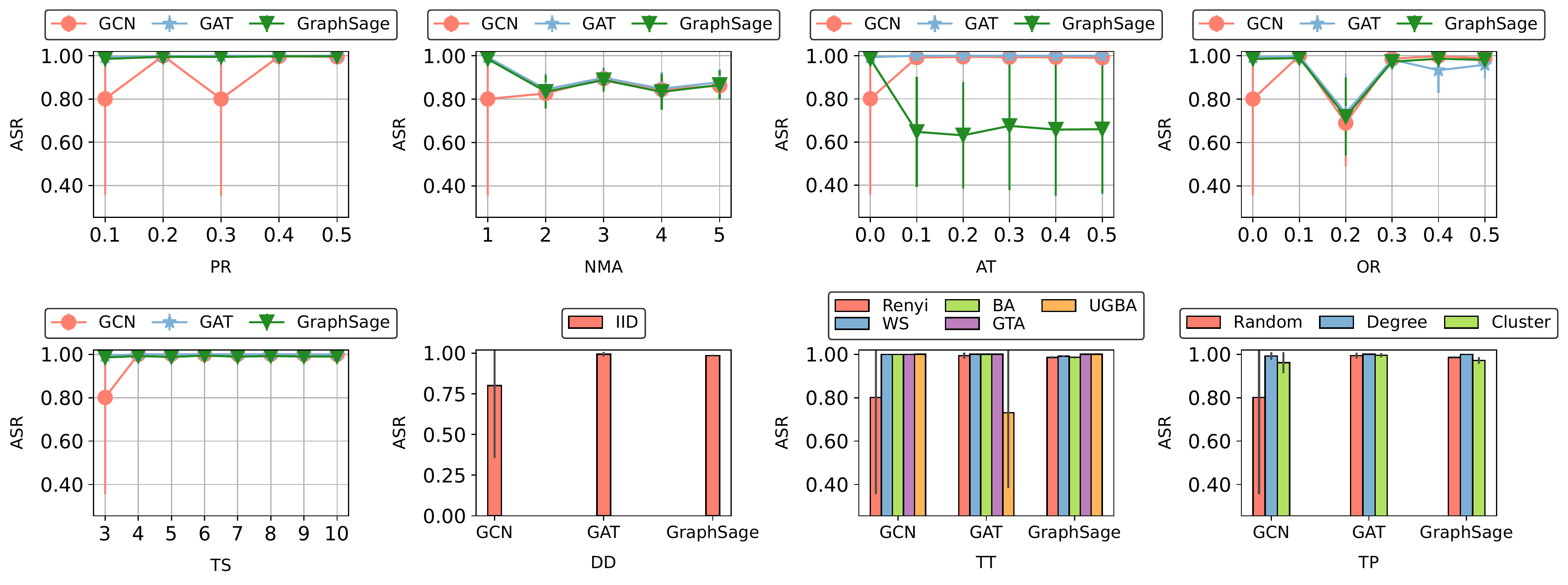}
%\caption{fig2}
\end{minipage}%
}%
\\
\subfigure[TASR]{
\begin{minipage}[t]{1.0\linewidth}
\centering
\includegraphics[width=5.0in]{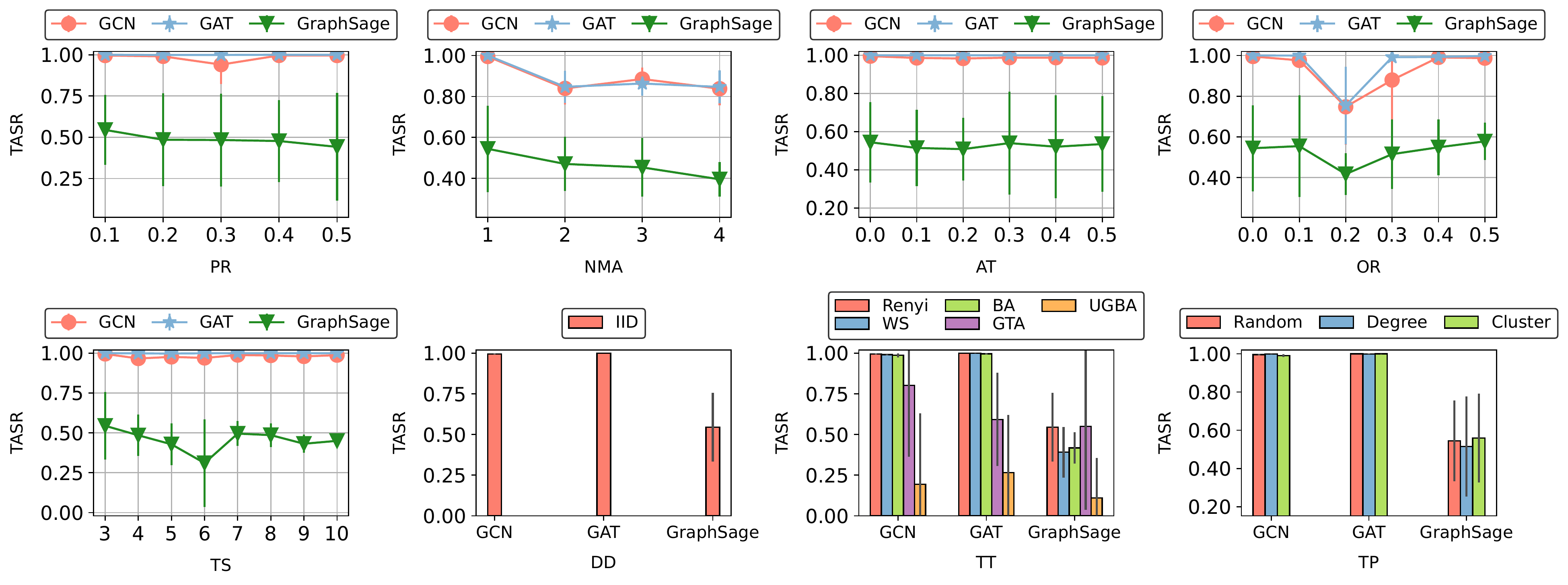}
%\caption{fig2}
\end{minipage}
}%
\centering
\caption{Graph backdoor attack on Physics.}~\label{figs:Appendix-factors-Physics}
\end{figure}

\begin{figure}[!]
\centering
\subfigure[ACC]{
\begin{minipage}[t]{1.0\linewidth}
\centering
\includegraphics[width=5.0in]{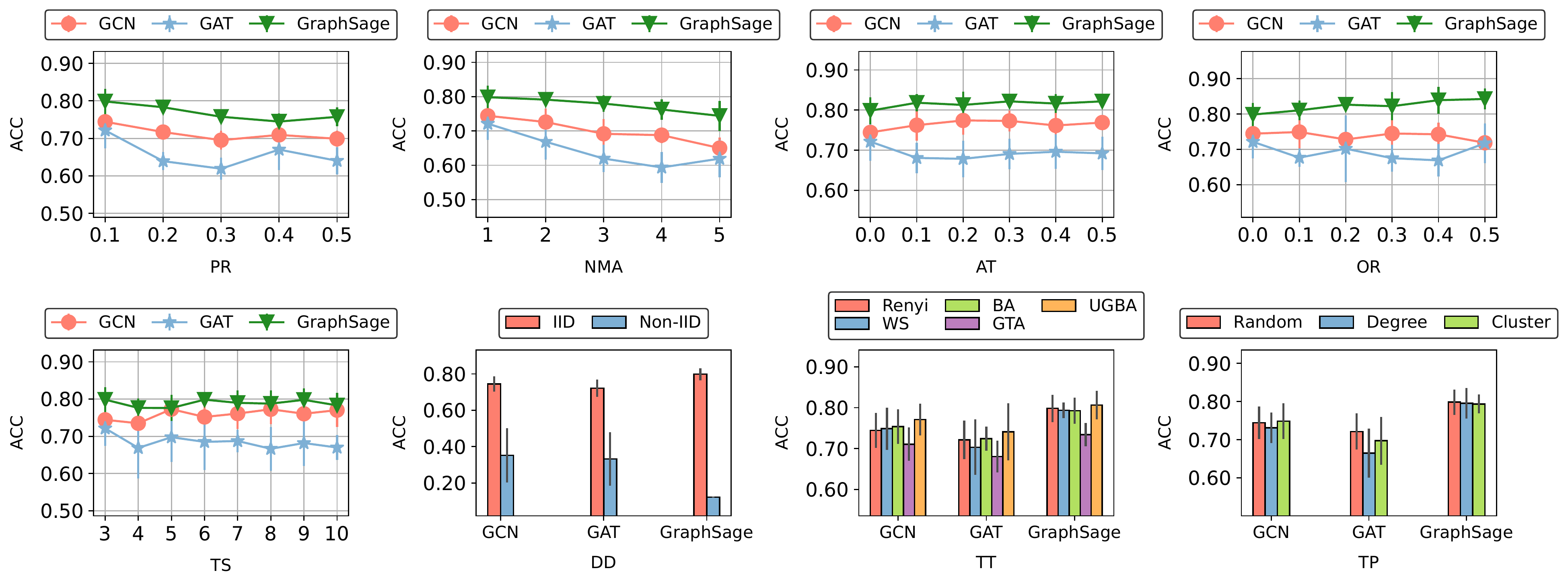}
%\caption{fig1}
\end{minipage}%
}%
\\
\subfigure[ASR]{
\begin{minipage}[t]{1.0\linewidth}
\centering
\includegraphics[width=5.0in]{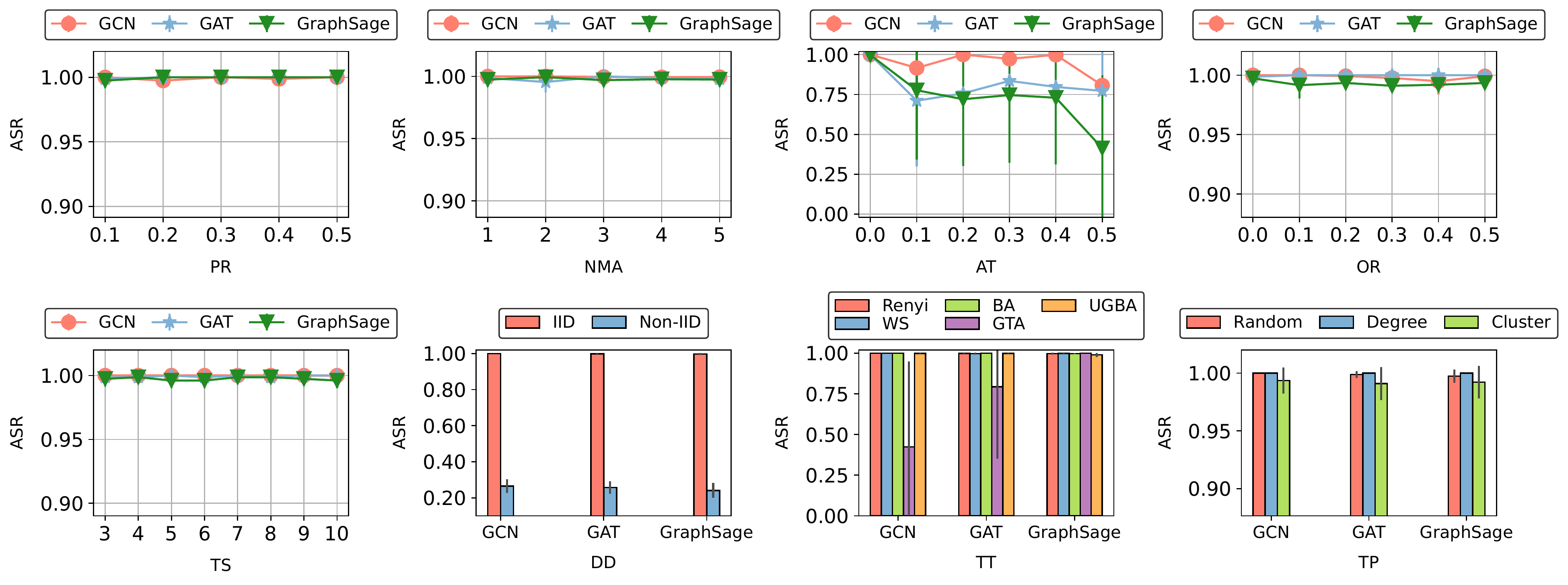}
%\caption{fig2}
\end{minipage}%
}%
\\
\subfigure[TASR]{
\begin{minipage}[t]{1.0\linewidth}
\centering
\includegraphics[width=5.0in]{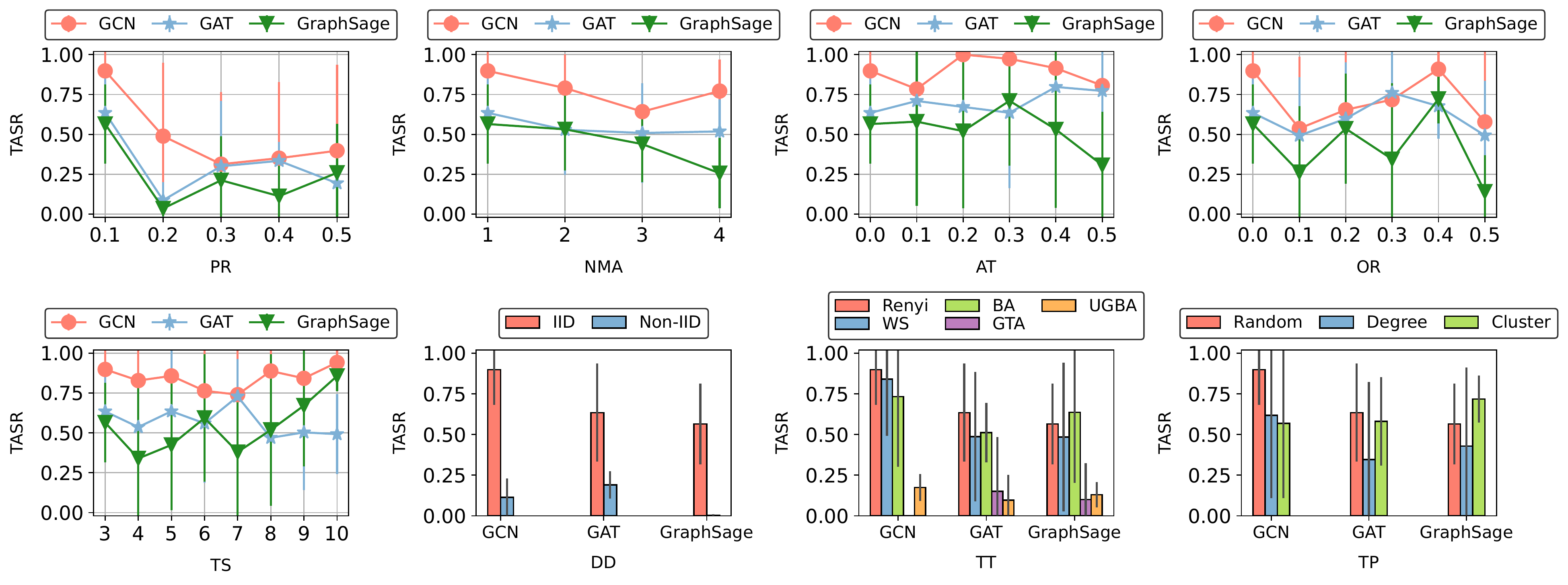}
%\caption{fig2}
\end{minipage}
}%
\centering
\caption{Graph backdoor attack on Photo.}~\label{figs:Appendix-factors-Photo}
\end{figure}

\begin{figure}[!]
\centering
\subfigure[ACC]{
\begin{minipage}[t]{1.0\linewidth}
\centering
\includegraphics[width=5.0in]{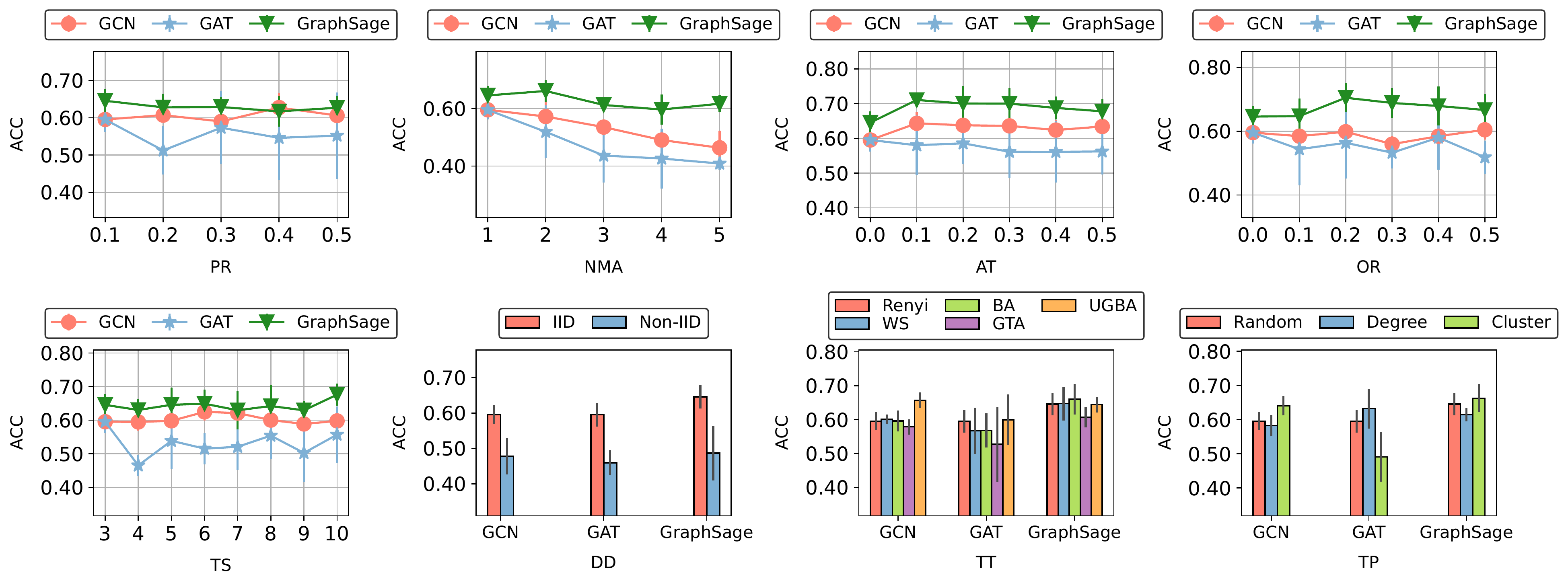}
%\caption{fig1}
\end{minipage}%
}%
\\
\subfigure[ASR]{
\begin{minipage}[t]{1.0\linewidth}
\centering
\includegraphics[width=5.0in]{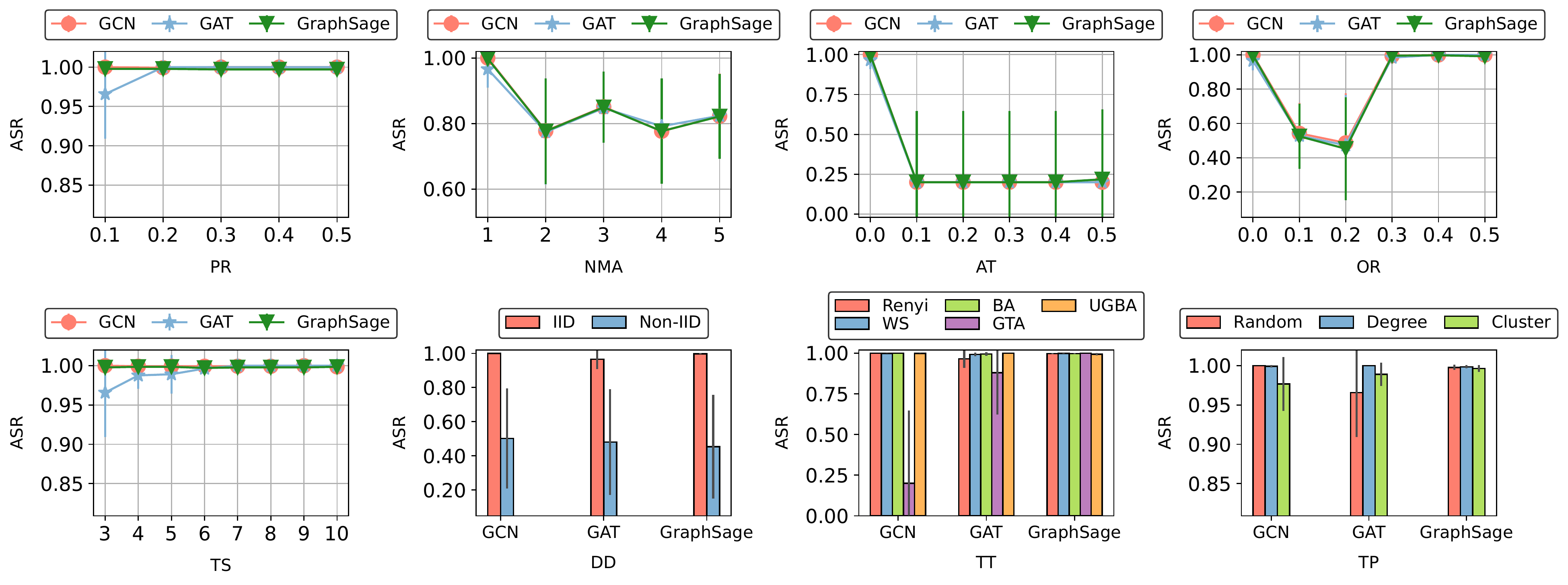}
%\caption{fig2}
\end{minipage}%
}%
\\
\subfigure[TASR]{
\begin{minipage}[t]{1.0\linewidth}
\centering
\includegraphics[width=5.0in]{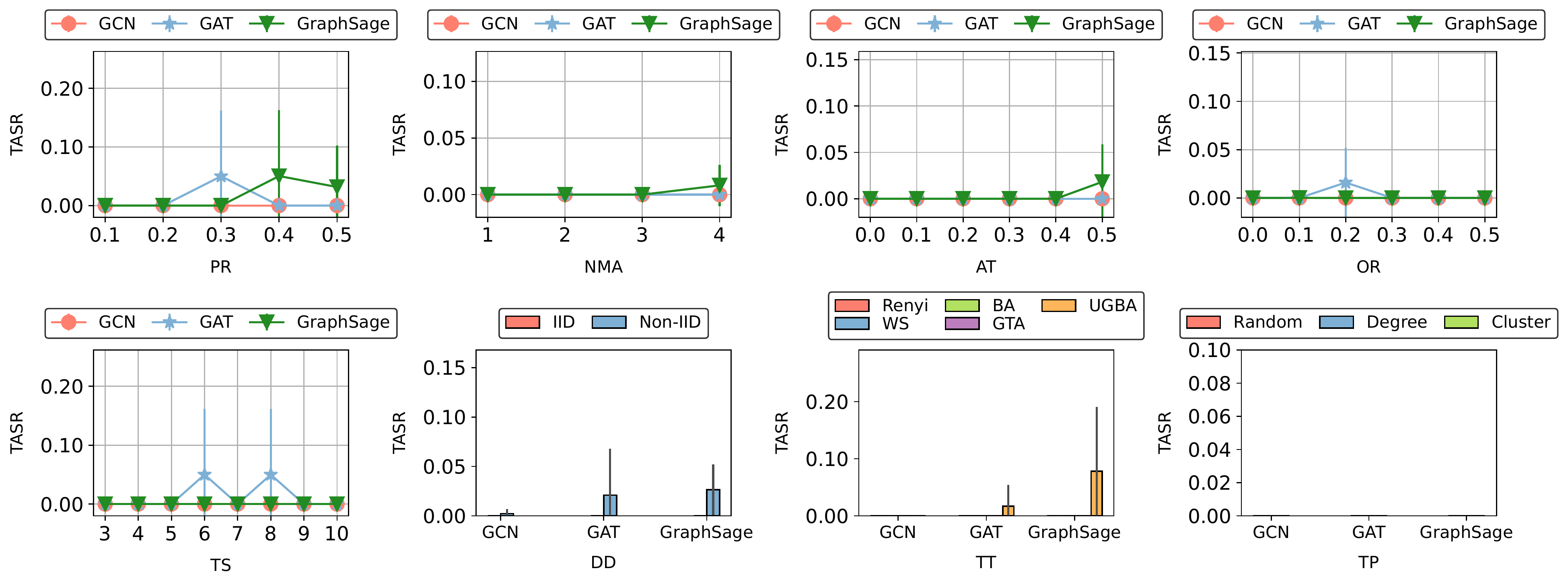}
%\caption{fig2}
\end{minipage}
}%
\centering
\caption{Graph backdoor attack on Computers.}~\label{figs:Appendix-factors-Computers}
\end{figure}

\begin{figure}[!]
\centering
\subfigure[ACC]{
\begin{minipage}[t]{1.0\linewidth}
\centering
\includegraphics[width=5.0in]{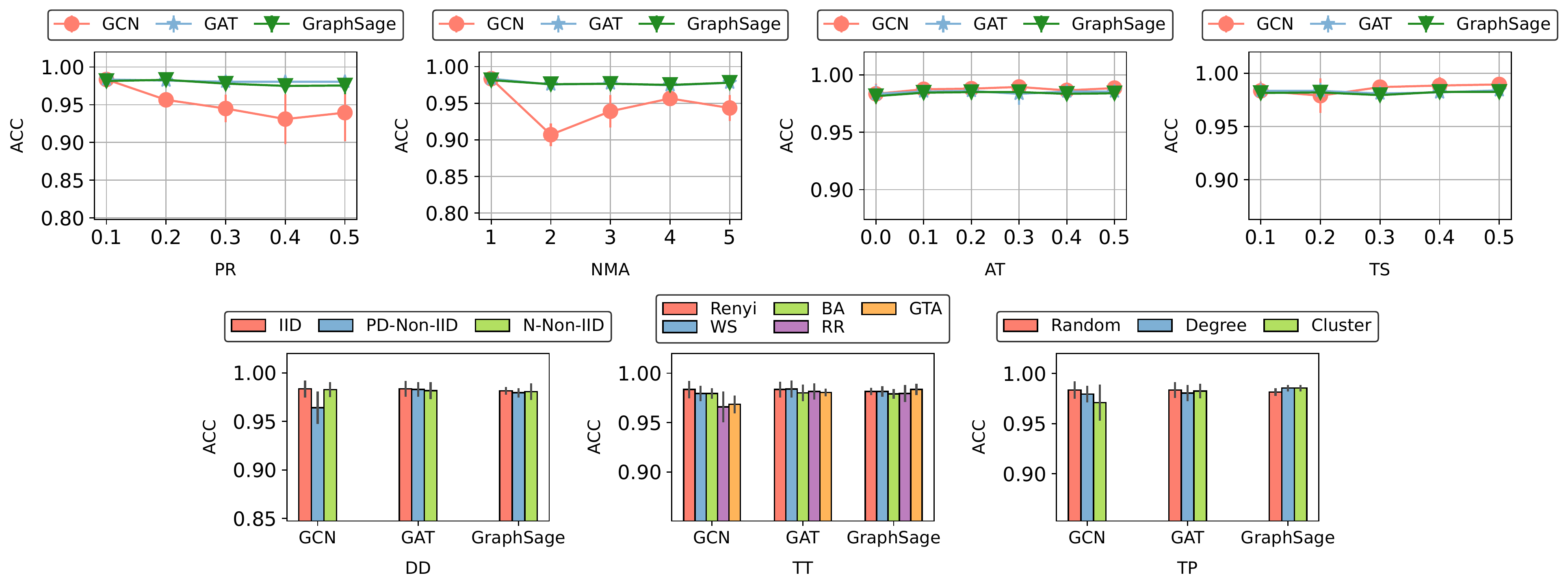}
%\caption{fig1}
\end{minipage}%
}%
\\
\subfigure[ASR]{
\begin{minipage}[t]{1.0\linewidth}
\centering
\includegraphics[width=5.0in]{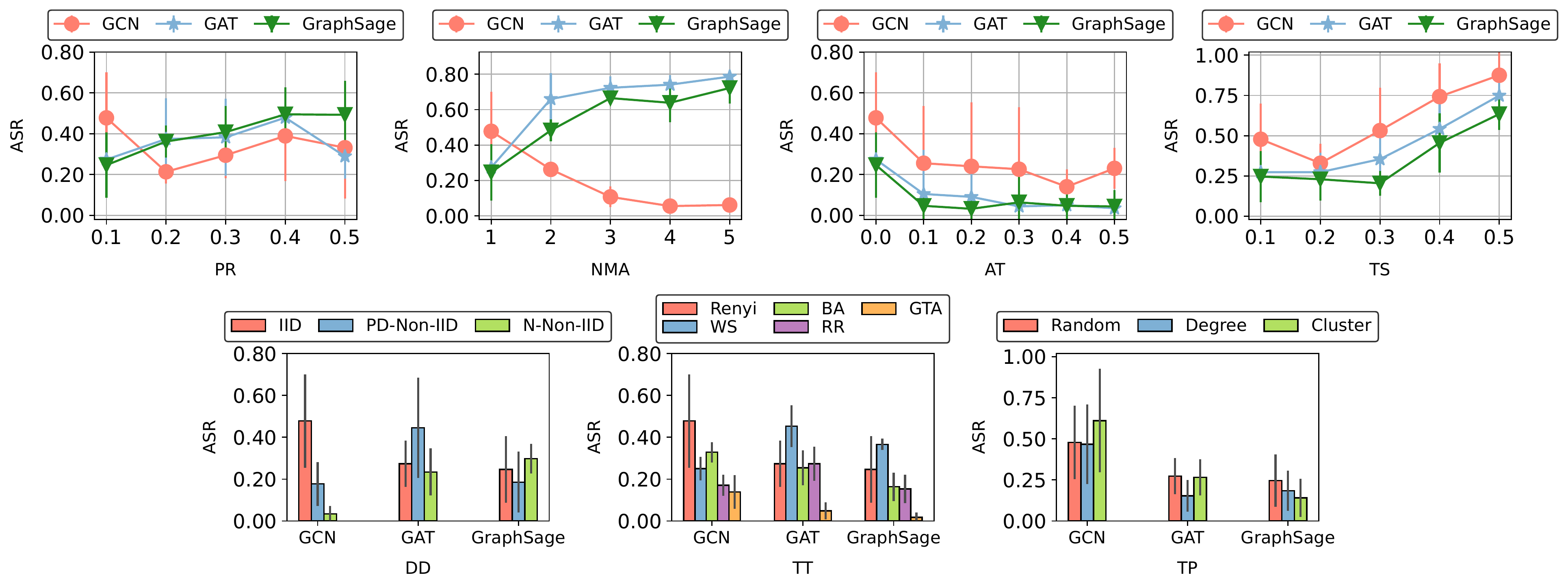}
%\caption{fig2}
\end{minipage}%
}%
\\
\subfigure[TASR]{
\begin{minipage}[t]{1.0\linewidth}
\centering
\includegraphics[width=5.0in]{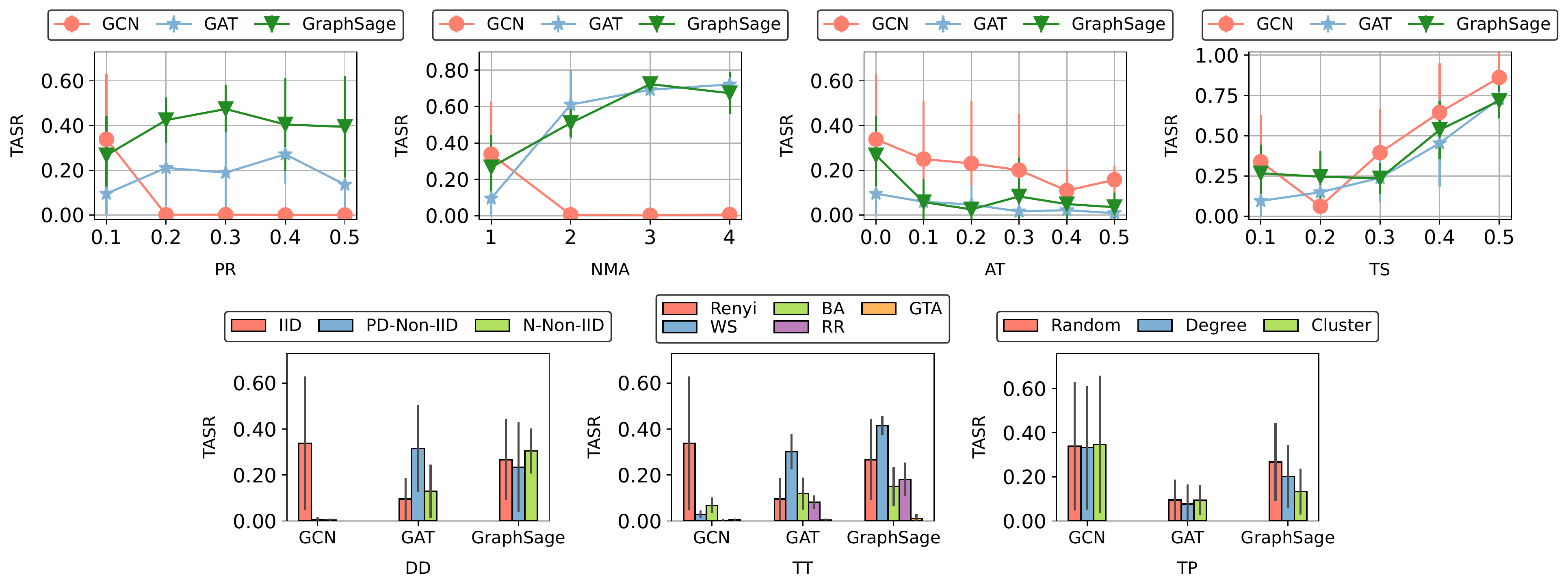}
%\caption{fig2}
\end{minipage}
}%
\centering
\caption{Graph backdoor attack on AIDS.}~\label{figs:Appendix-factors-AIDS}
\end{figure}

\begin{figure}[!]
\centering
\subfigure[ACC]{
\begin{minipage}[t]{1.0\linewidth}
\centering
\includegraphics[width=5.0in]{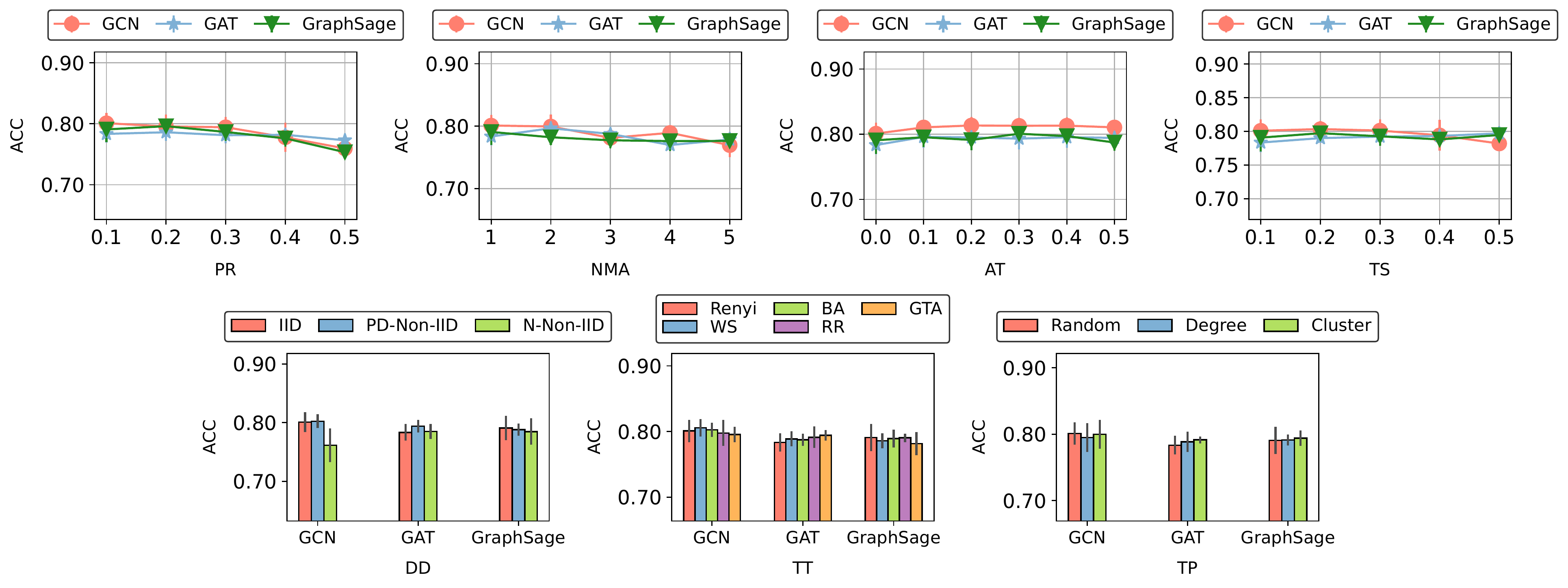}
%\caption{fig1}
\end{minipage}%
}%
\\
\subfigure[ASR]{
\begin{minipage}[t]{1.0\linewidth}
\centering
\includegraphics[width=5.0in]{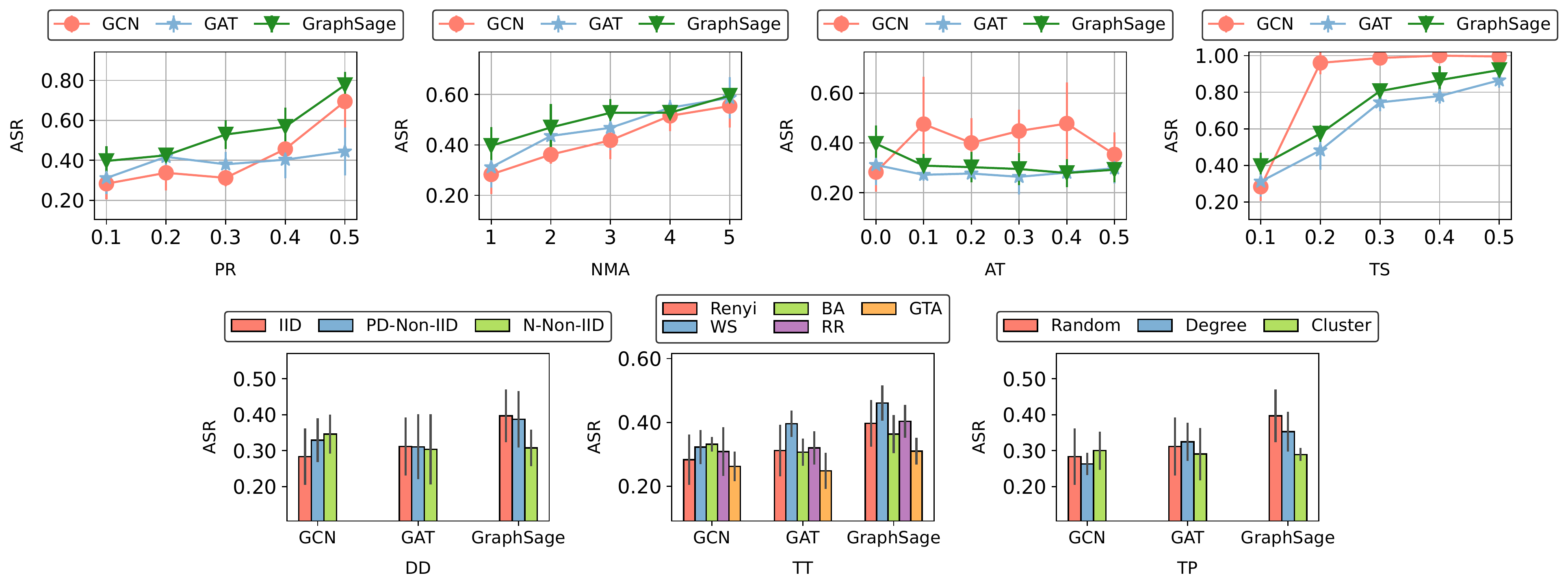}
%\caption{fig2}
\end{minipage}%
}%
\\
\subfigure[TASR]{
\begin{minipage}[t]{1.0\linewidth}
\centering
\includegraphics[width=5.0in]{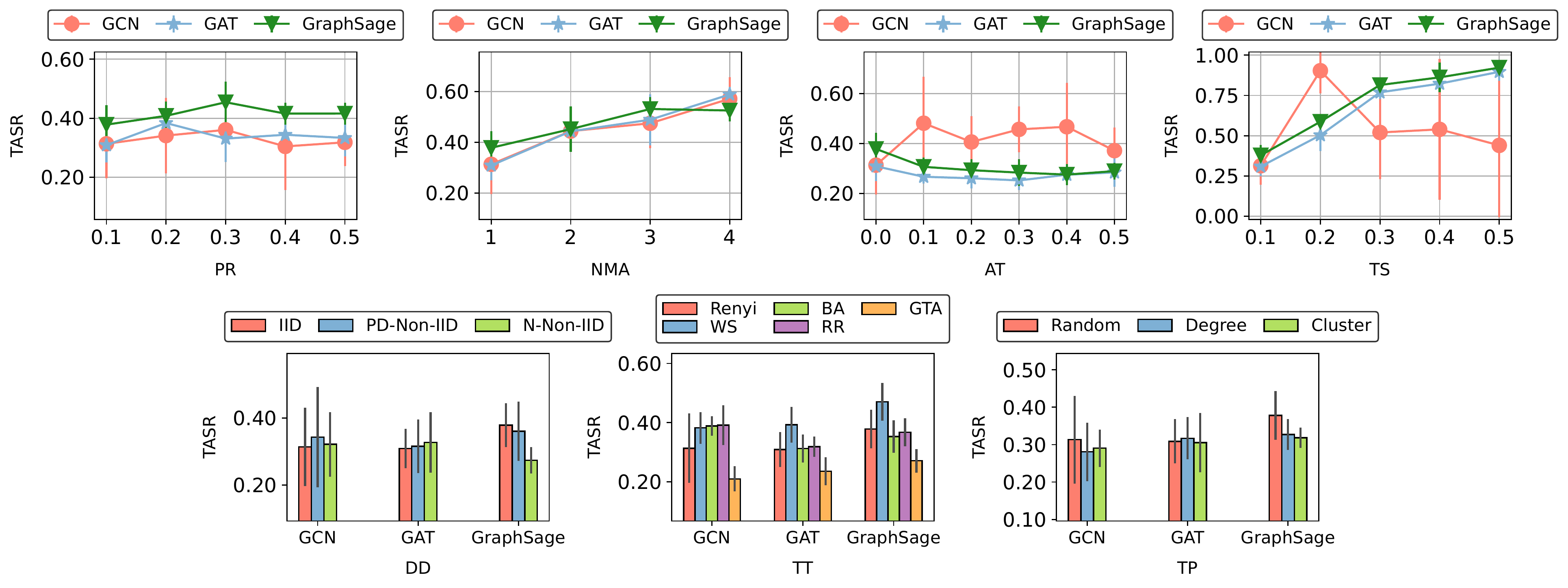}
%\caption{fig2}
\end{minipage}
}%
\centering
\caption{Graph backdoor attack on NCI1.}~\label{figs:Appendix-factors-NCI1}
\end{figure}

\begin{figure}[!]
\centering
\subfigure[ACC]{
\begin{minipage}[t]{1.0\linewidth}
\centering
\includegraphics[width=5.0in]{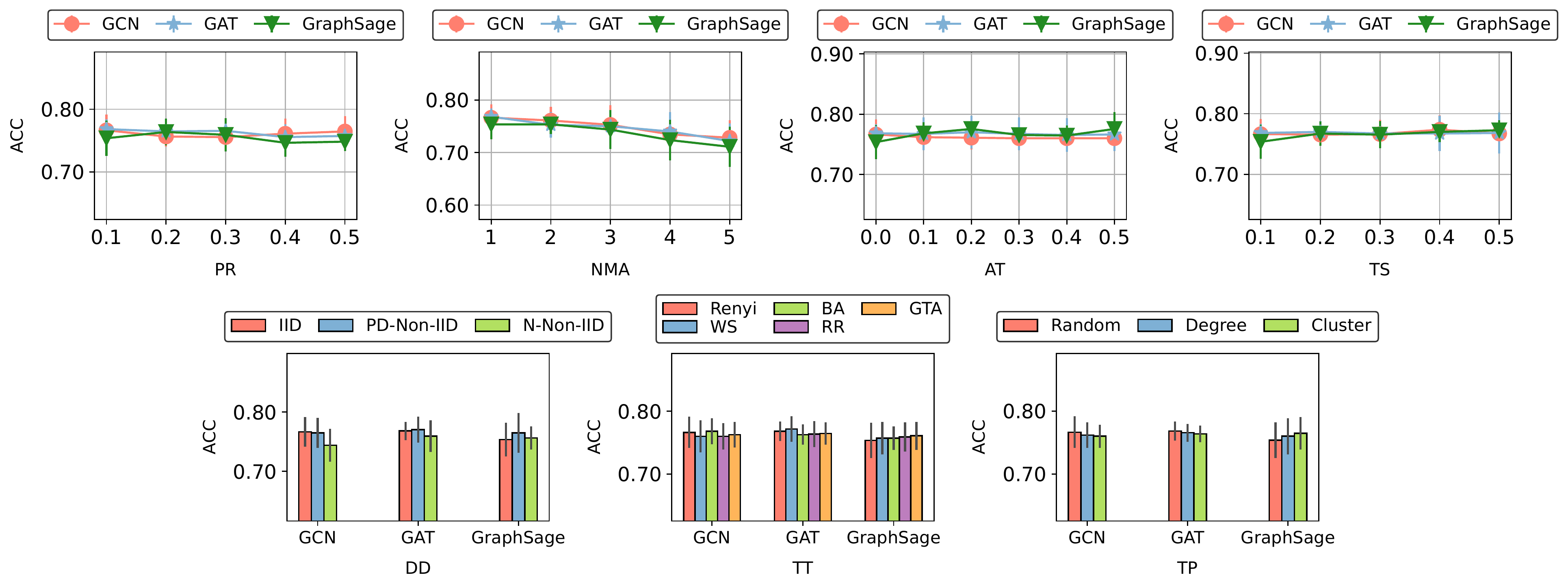}
%\caption{fig1}
\end{minipage}%
}%
\\
\subfigure[ASR]{
\begin{minipage}[t]{1.0\linewidth}
\centering
\includegraphics[width=5.0in]{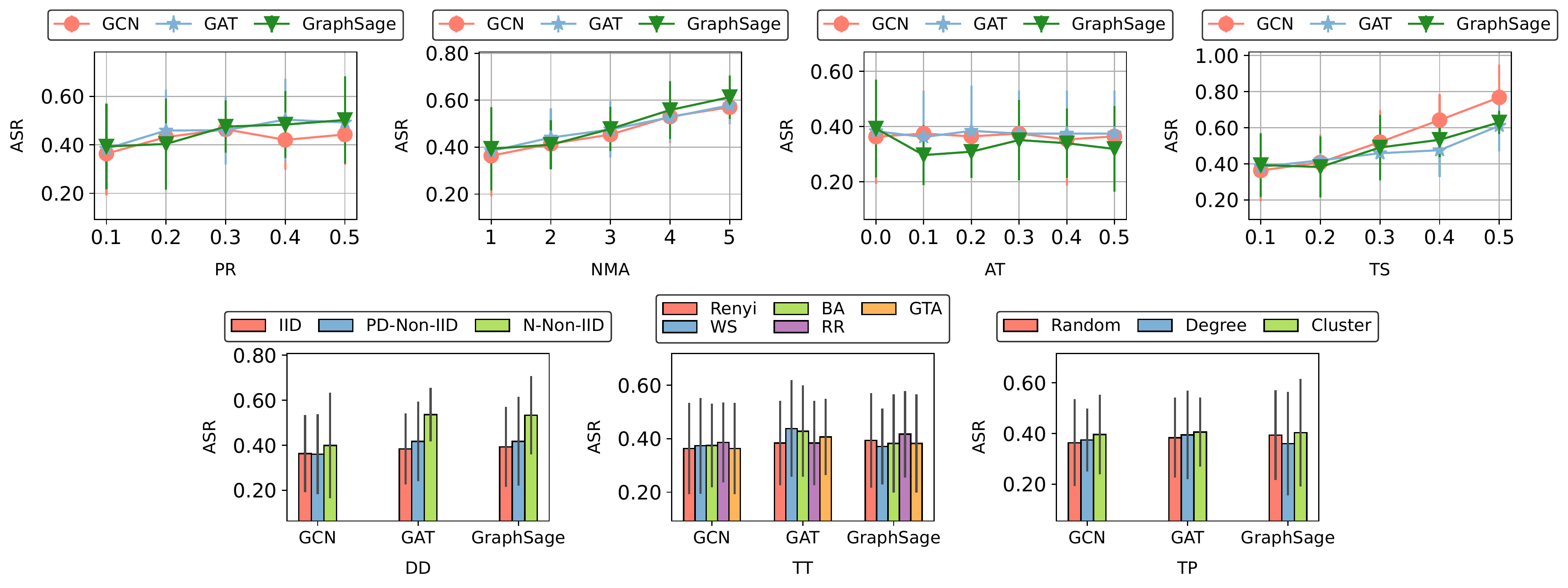}
%\caption{fig2}
\end{minipage}%
}%
\\
\subfigure[TASR]{
\begin{minipage}[t]{1.0\linewidth}
\centering
\includegraphics[width=5.0in]{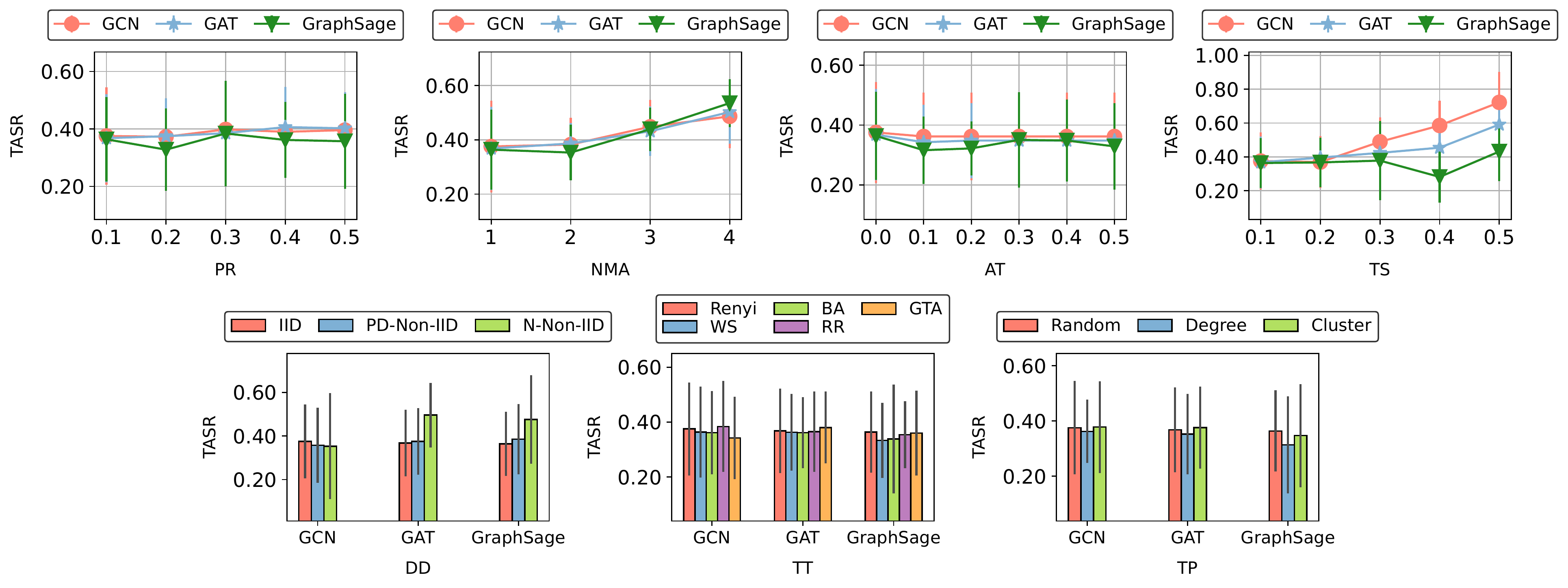}
%\caption{fig2}
\end{minipage}
}%
\centering
\caption{Graph backdoor attack on PROTEINS-full.}~\label{figs:Appendix-factors-PROTEINS-full}
\end{figure}

\begin{figure}[!]
\centering
\subfigure[ACC]{
\begin{minipage}[t]{1.0\linewidth}
\centering
\includegraphics[width=5.0in]{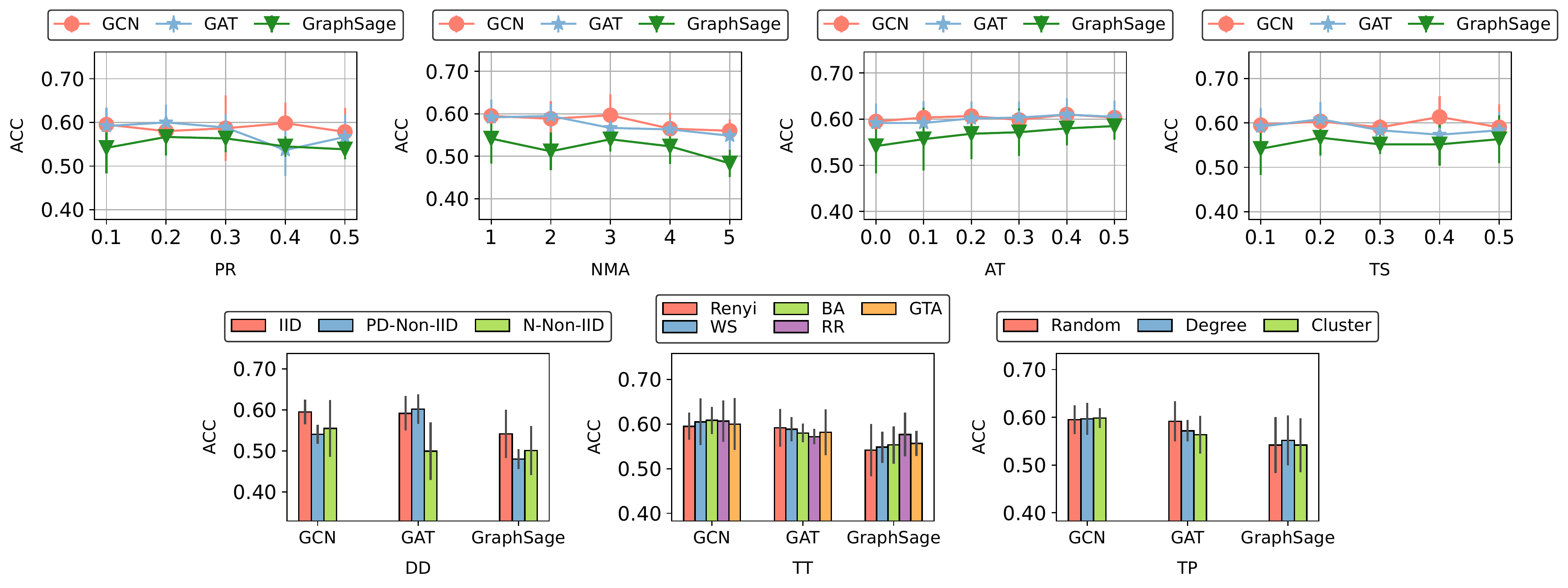}
%\caption{fig1}
\end{minipage}%
}%
\\
\subfigure[ASR]{
\begin{minipage}[t]{1.0\linewidth}
\centering
\includegraphics[width=5.0in]{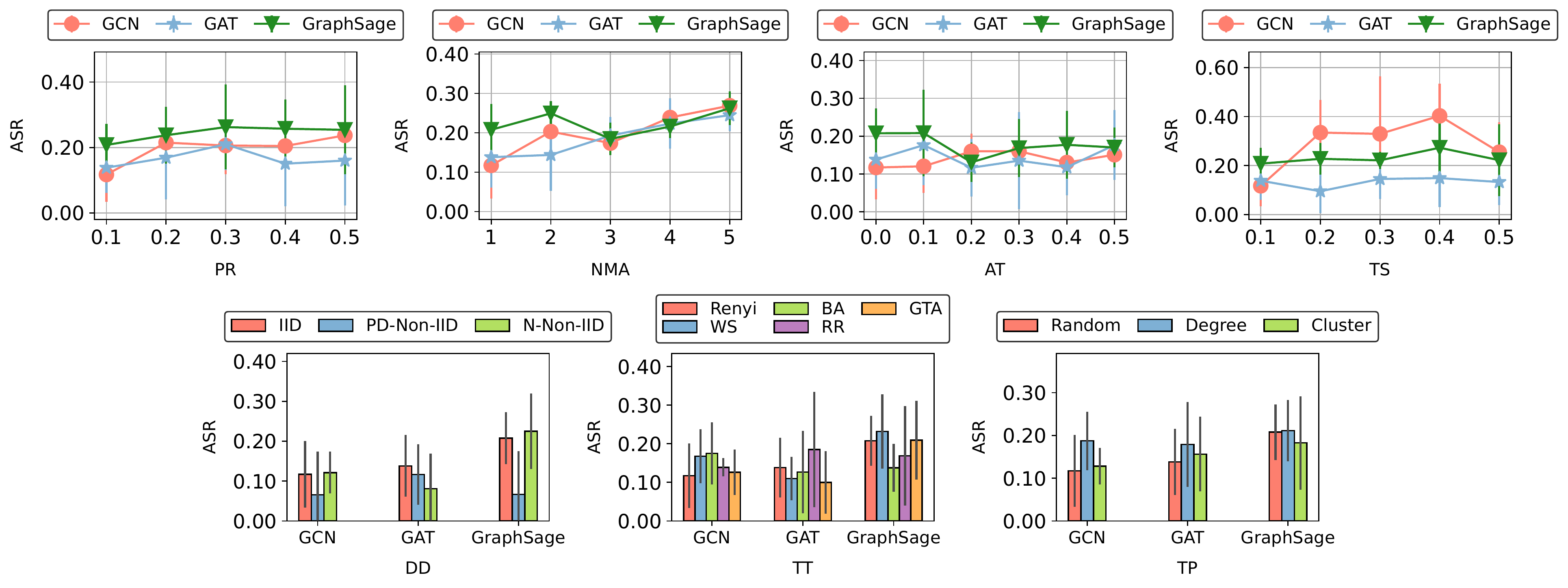}
%\caption{fig2}
\end{minipage}%
}%
\\
\subfigure[TASR]{
\begin{minipage}[t]{1.0\linewidth}
\centering
\includegraphics[width=5.0in]{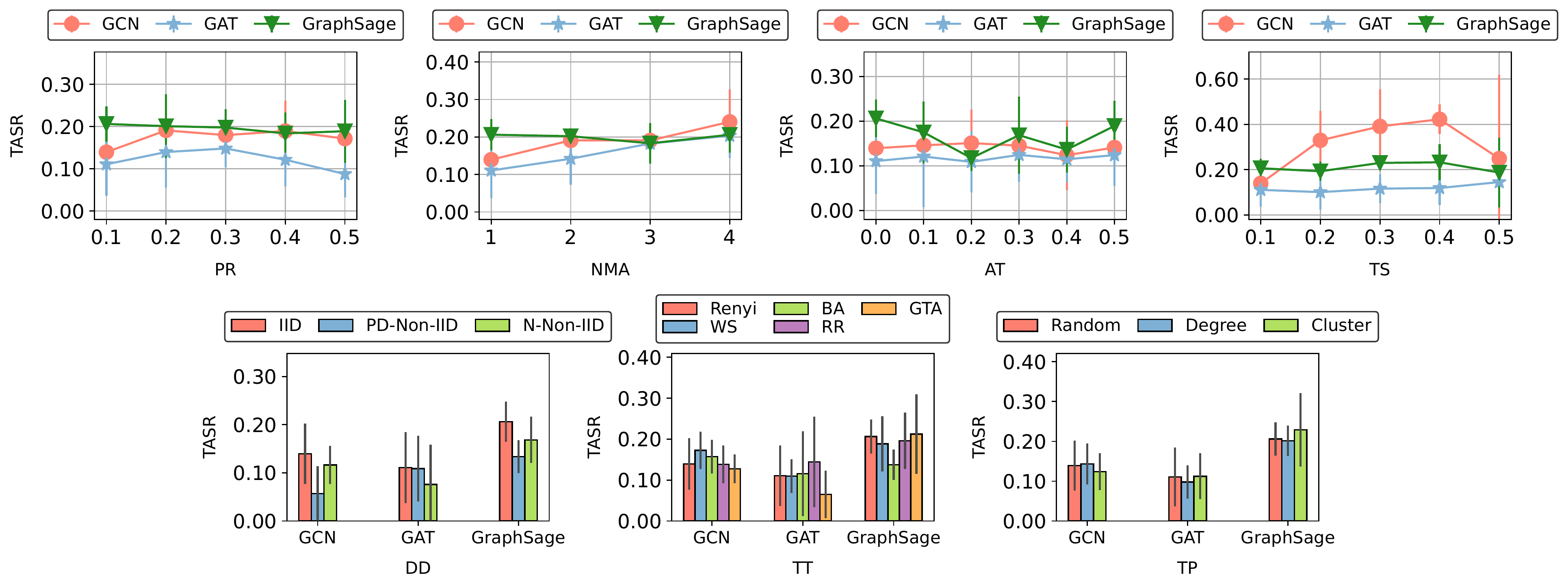}
%\caption{fig2}
\end{minipage}
}%
\centering
\caption{Graph backdoor attack on ENZYMES.}~\label{figs:Appendix-factors-ENZYMES}
\end{figure}

\begin{figure}[!]
\centering
\subfigure[ACC]{
\begin{minipage}[t]{1.0\linewidth}
\centering
\includegraphics[width=5.0in]{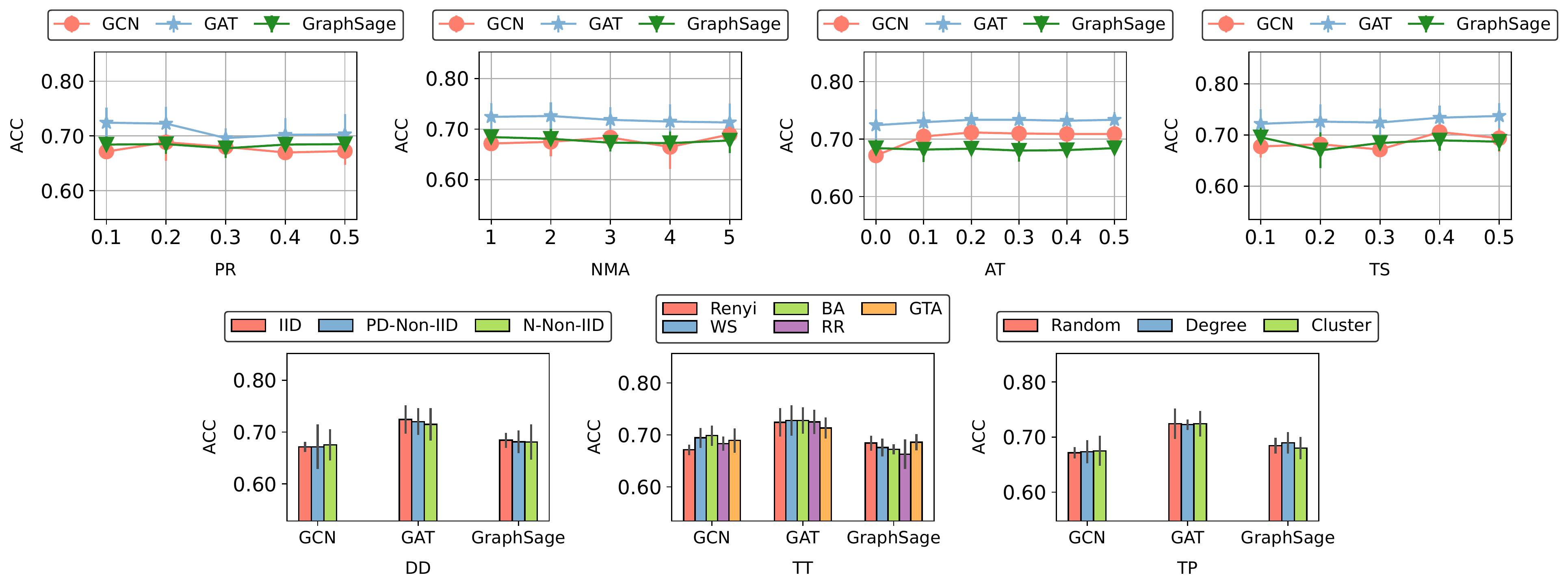}
%\caption{fig1}
\end{minipage}%
}%
\\
\subfigure[ASR]{
\begin{minipage}[t]{1.0\linewidth}
\centering
\includegraphics[width=5.0in]{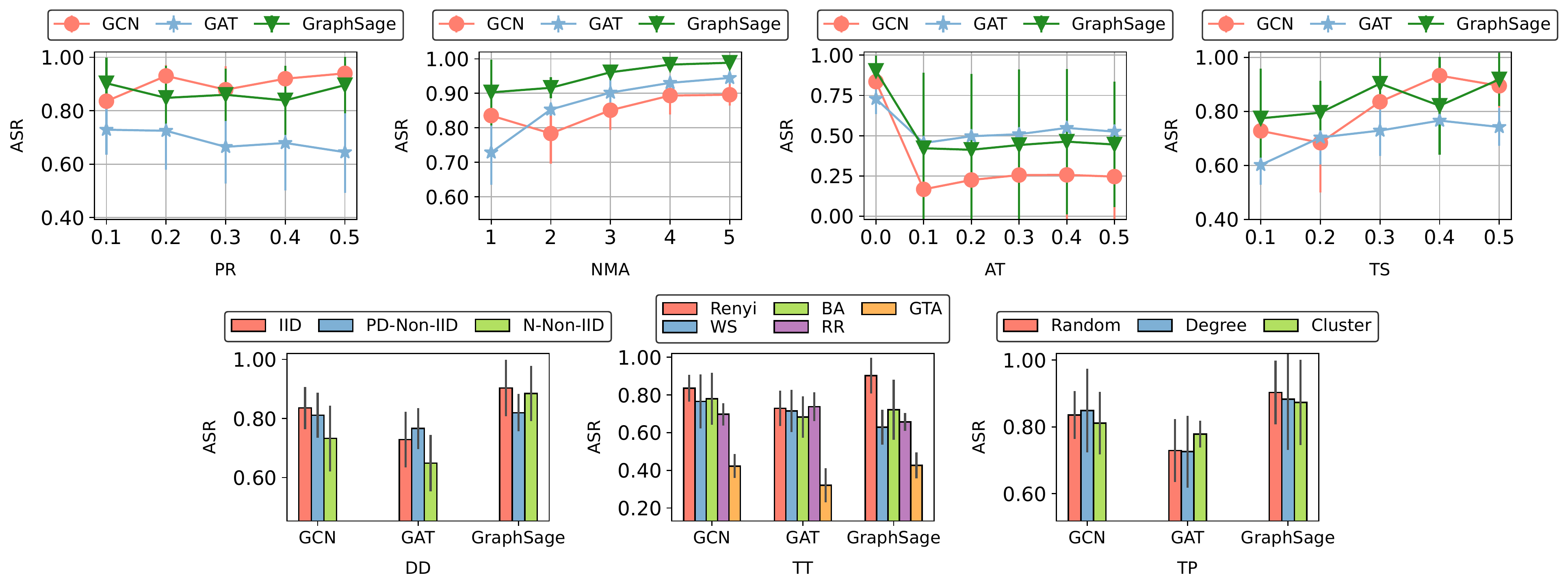}
%\caption{fig2}
\end{minipage}%
}%
\\
\subfigure[TASR]{
\begin{minipage}[t]{1.0\linewidth}
\centering
\includegraphics[width=5.0in]{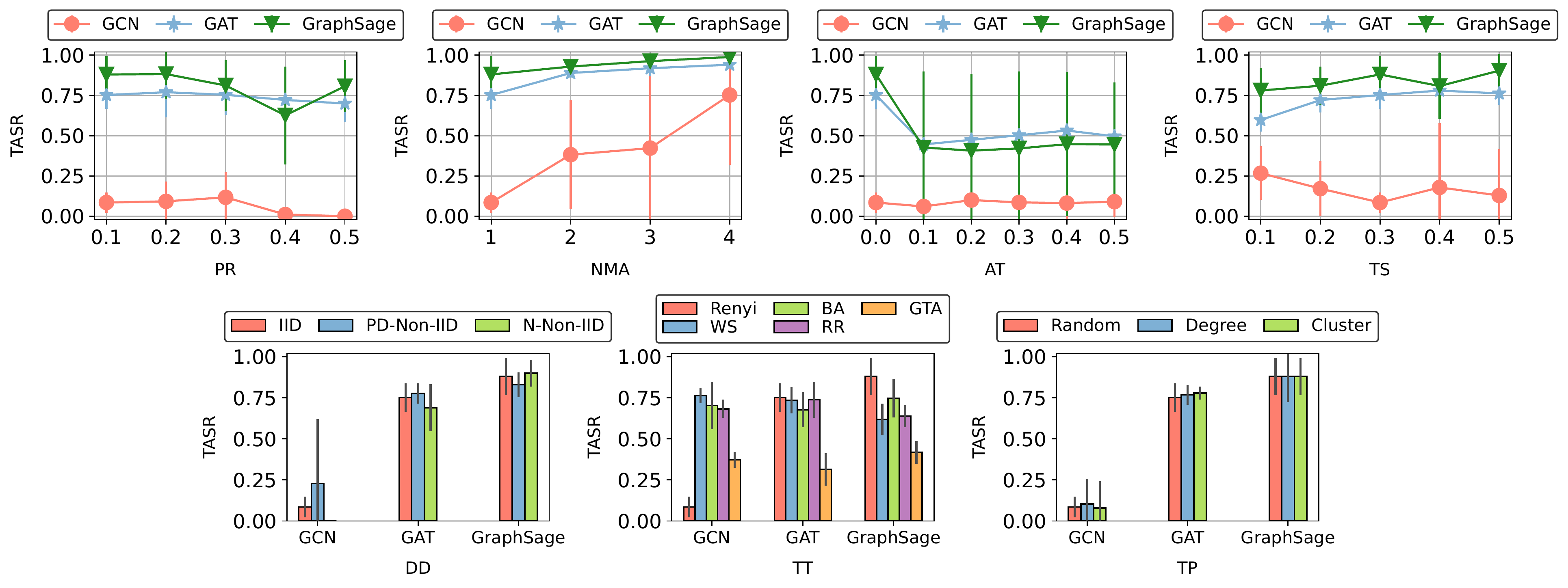}
%\caption{fig2}
\end{minipage}
}%
\centering
\caption{Graph backdoor attack on DD.}~\label{figs:Appendix-factors-DD}
\end{figure}

\begin{figure}[!]
\centering
\subfigure[ACC]{
\begin{minipage}[t]{1.0\linewidth}
\centering
\includegraphics[width=5.0in]{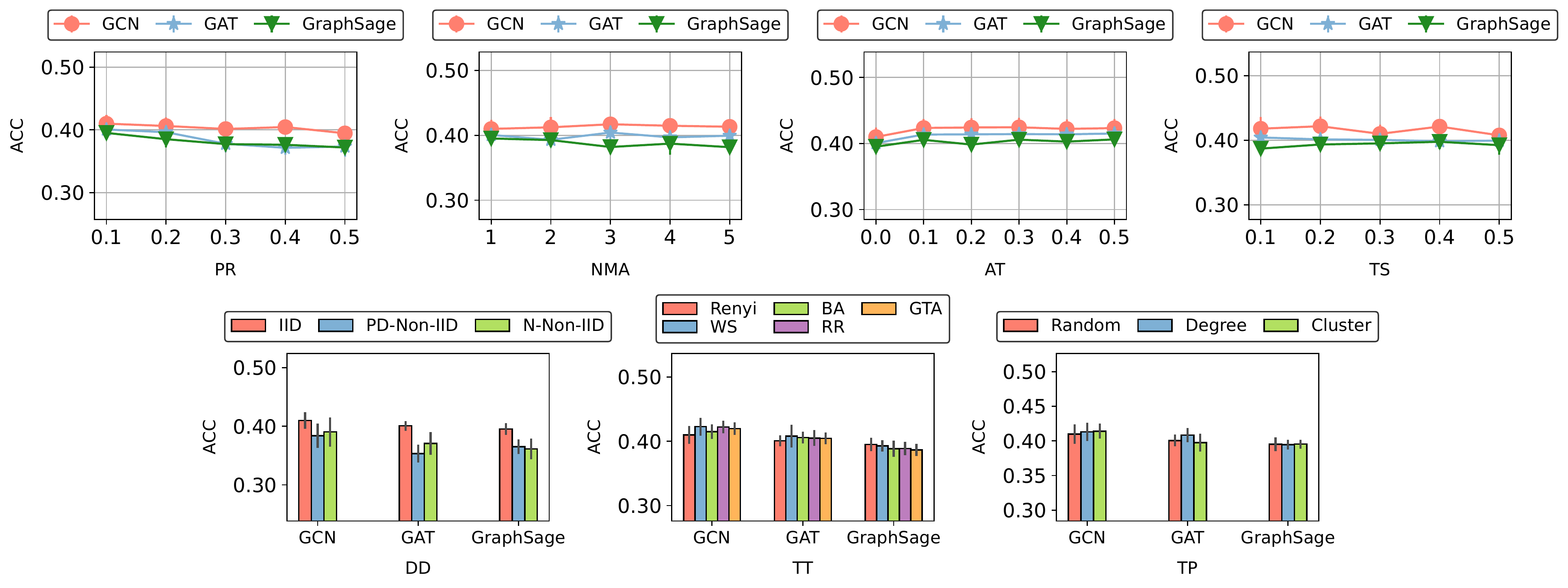}
%\caption{fig1}
\end{minipage}%
}%
\\
\subfigure[ASR]{
\begin{minipage}[t]{1.0\linewidth}
\centering
\includegraphics[width=5.0in]{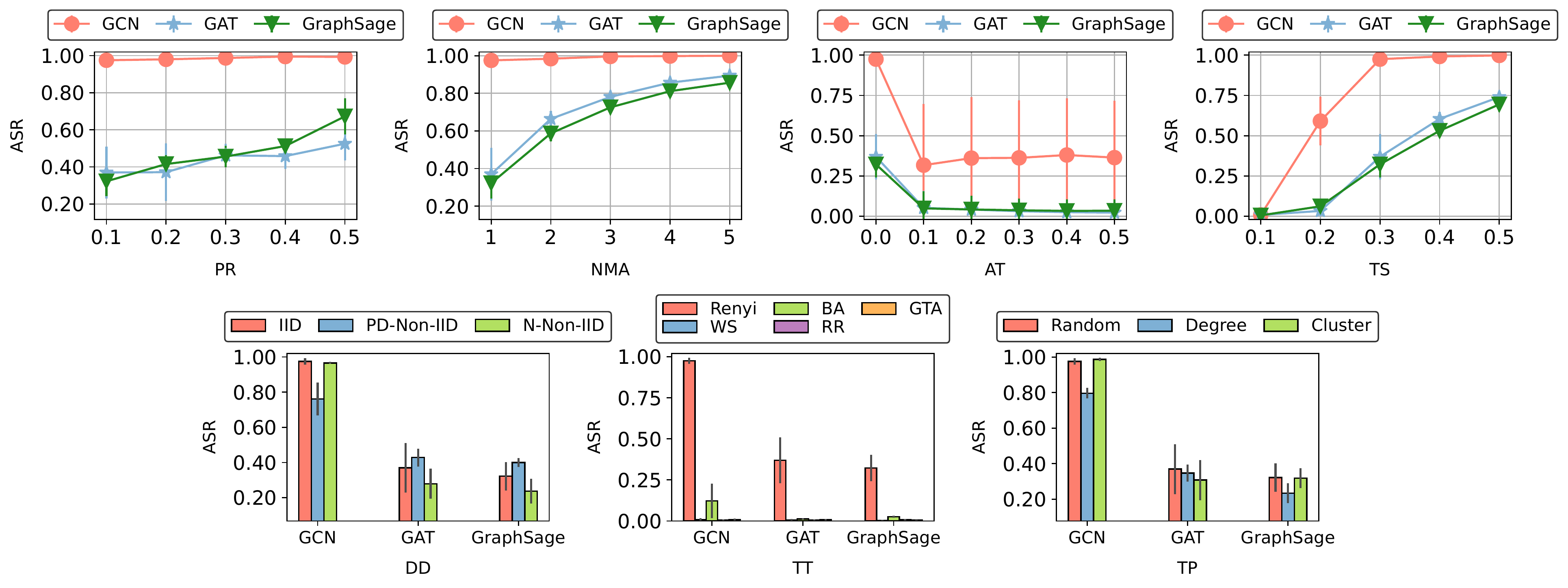}
%\caption{fig2}
\end{minipage}%
}%
\\
\subfigure[TASR]{
\begin{minipage}[t]{1.0\linewidth}
\centering
\includegraphics[width=5.0in]{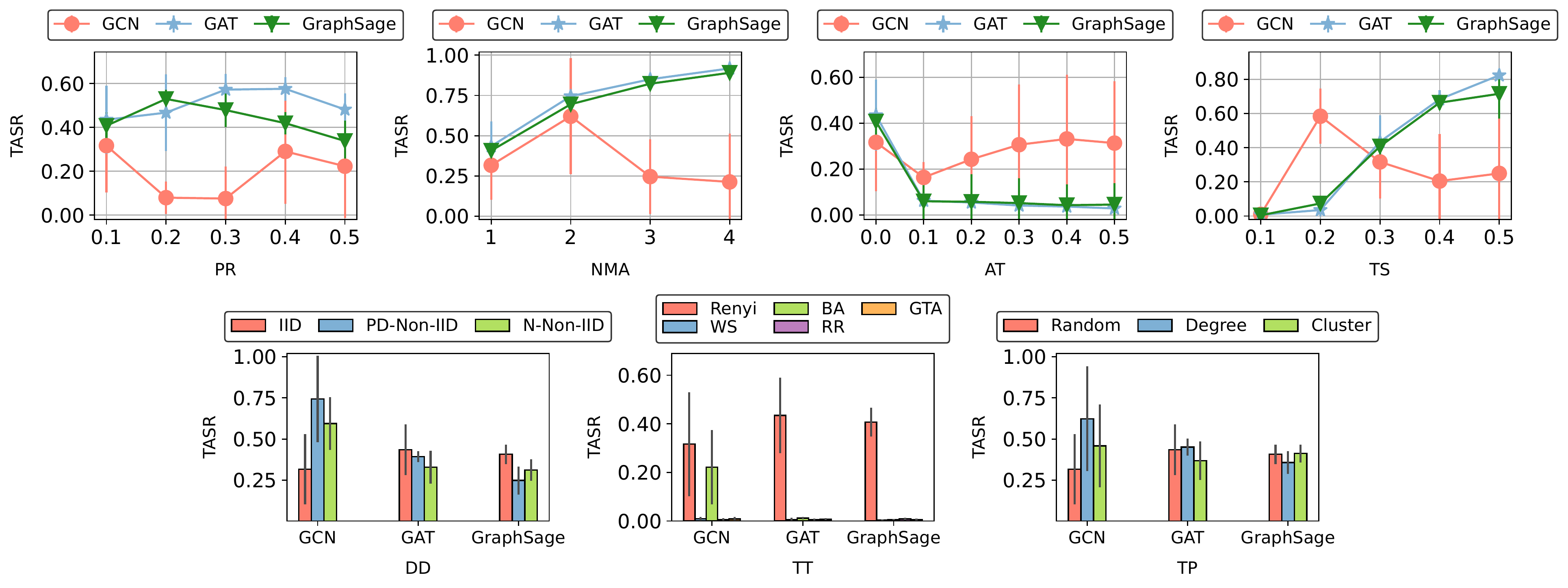}
%\caption{fig2}
\end{minipage}
}%
\centering
\caption{Graph backdoor attack on COLORS-3.}~\label{figs:Appendix-factors-COLORS-3}
\end{figure}

The experiments investigating critical factors are depicted in Figures~\ref{figs:Appendix-factors-Cora}-\ref{figs:Appendix-factors-COLORS-3}. Several key observations can be made: (1) In both node and graph-level tasks, an increase in PR is associated with a rise in ASR across the majority of datasets.
(2) In node-level tasks, ASR decreases as NMA increases, whereas in graph-level tasks, ASR increases with the growth of NMA.
(3) Backdoor attacks are effective in any training round; however, ASR tends to decrease with increasing AT across most datasets.
(4) ACC increases with rising OR for the majority of datasets, while ASR observations are not consistent. For instance, ASR exhibits a trough in the Computers dataset when OR is 0.2 and remains relatively stable in the CS dataset.
(5) TS has minimal influence on node-level tasks, as a sizable ASR is already present when TS is at its minimum value of 3. However, ASR progressively increases with TS in graph-level tasks.
(6) In node-level tasks, IID exhibits a higher ASR than Non-IID.
(7) Adaptive triggers yield higher ASR in node-level tasks compared to universal triggers, whereas in graph-level tasks, universal triggers outperform adaptive triggers for most tasks.
(8) The importance-based position achieves higher ASR than random positions across most datasets. 

% \textbf{Cora.} The investigation of critical factors of graph backdoor attack on FedGNN on Cora shown in Figure~\ref{figs:Appendix-factors-Cora}.

% The experimental results details of each datasets can be seen  as follows.
% \textbf{CiteSeer.} The investigation of critical factors of graph backdoor attack on FedGNN on Dataset CiteSeer shown in Figure~\ref{figs:Appendix-factors-Citeseer}.

% \textbf{Pubmed.} The investigation of critical factors of graph backdoor attack on FedGNN on Pubmed shown in Figure~\ref{figs:Appendix-factors-Citeseer}.

% \textbf{CS.} The investigation of critical factors of graph backdoor attack on FedGNN on CS shown in Figure~\ref{figs:Appendix-factors-CS}.

% \textbf{Physics.} The investigation of critical factors of graph backdoor attack on FedGNN on Physics shown in Figure~\ref{figs:Appendix-factors-Physics}.

% \textbf{Photo.} The investigation of critical factors of graph backdoor attack on FedGNN on Photo shown in Figure~\ref{figs:Appendix-factors-Photo}.

% \textbf{Computers.} The investigation of critical factors of graph backdoor attack on FedGNN on Computers shown in Figure~\ref{figs:Appendix-factors-Computers}.

% \textbf{AIDS.} The investigation of critical factors of graph backdoor attack on FedGNN on AIDS shown in Figure~\ref{figs:Appendix-factors-AIDS}.

% \textbf{NCI1.} The investigation of critical factors of graph backdoor attack on FedGNN on NCI1 shown in Figure~\ref{figs:Appendix-factors-NCI1}.

% \textbf{PROTEINS-full.} The investigation of critical factors of graph backdoor attack on FedGNN on PROTEINS-full shown in Figure~\ref{figs:Appendix-factors-PROTEINS-full}.

% \textbf{ENZYMES.} The investigation of critical factors of graph backdoor attack on FedGNN on ENZYMES shown in Figure~\ref{figs:Appendix-factors-ENZYMES}.

% \textbf{DD.} The investigation of critical factors of graph backdoor attack on FedGNN on DD shown in Figure~\ref{figs:Appendix-factors-DD}.

% \textbf{COLORS-3.} The investigation of critical factors of graph backdoor attack on FedGNN on COLORS-3 shown in Figure~\ref{figs:Appendix-factors-COLORS-3}.

\end{document}